\documentclass[10pt,twocolumn,letterpaper]{article}

\usepackage{cvpr}              
\usepackage{multirow}
\usepackage[accsupp]{axessibility}
\definecolor{cvprblue}{rgb}{0.21,0.49,0.74}
\usepackage[pagebackref,breaklinks,colorlinks,allcolors=cvprblue]{hyperref}

\def\paperID{*****} 
\def\confName{CVPR}
\def\confYear{2026}

\title{RetinexDualV2: Physically-Grounded Dual Retinex for Generalized UHD Image Restoration}

\author{
    Mohab Kishawy$^\dagger$ \quad Jun Chen \\
    {\tt\small \{kishawym, chenjun\}@mcmaster.ca} \\
    McMaster University \\
    $^\dagger$Corresponding Author
}

\begin{document}

{
\twocolumn[{
\renewcommand\twocolumn[1][]{#1}
\maketitle
\vspace{-6mm}
\begin{center}

\setlength{\abovecaptionskip}{1.5mm}
\setlength{\parskip}{0mm}
\setlength{\baselineskip}{0mm}
\begin{minipage}[c]{1\textwidth}
    \centering
    \setlength{\tabcolsep}{1pt}
    \begin{tabular}{c cc@{\hspace{3pt}} cc@{\hspace{3pt}} cc}
    & Input & Output & Input & Output & Input & Output \\
    \raisebox{0.7cm}{\rotatebox{90}{\scriptsize LLIE}} &
    \includegraphics[width=0.155\textwidth]{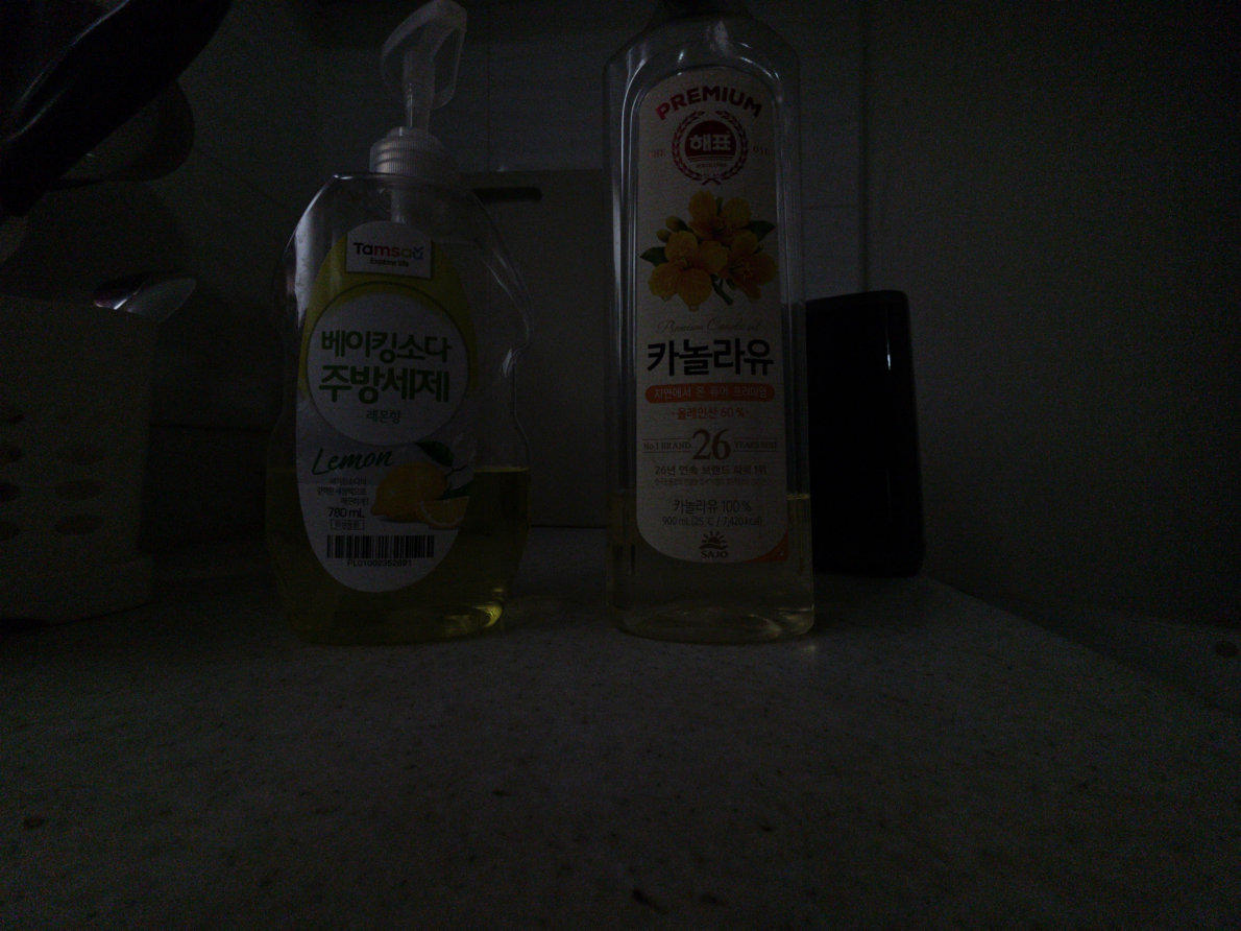} &
    \includegraphics[width=0.155\textwidth]{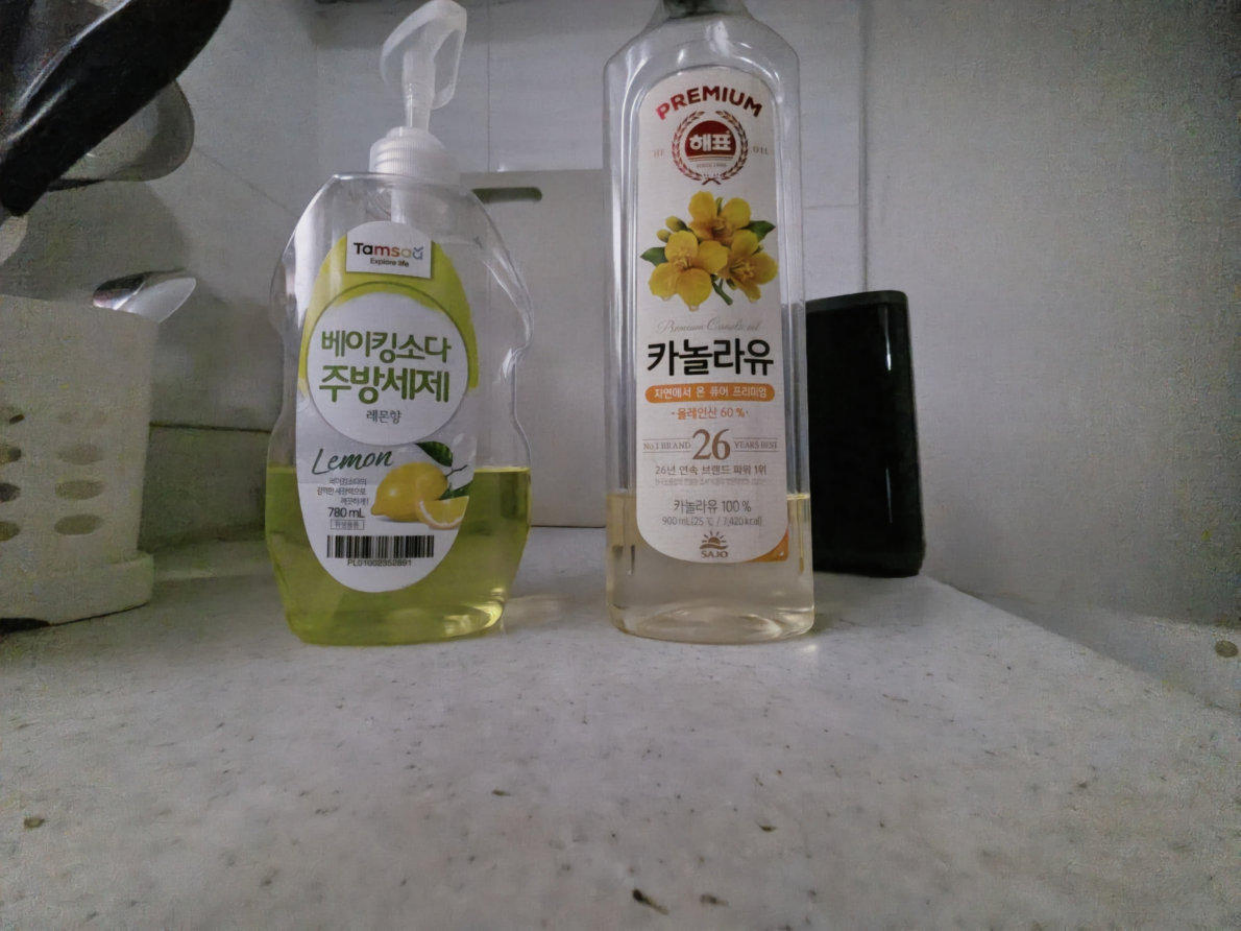} &
    \includegraphics[width=0.155\textwidth]{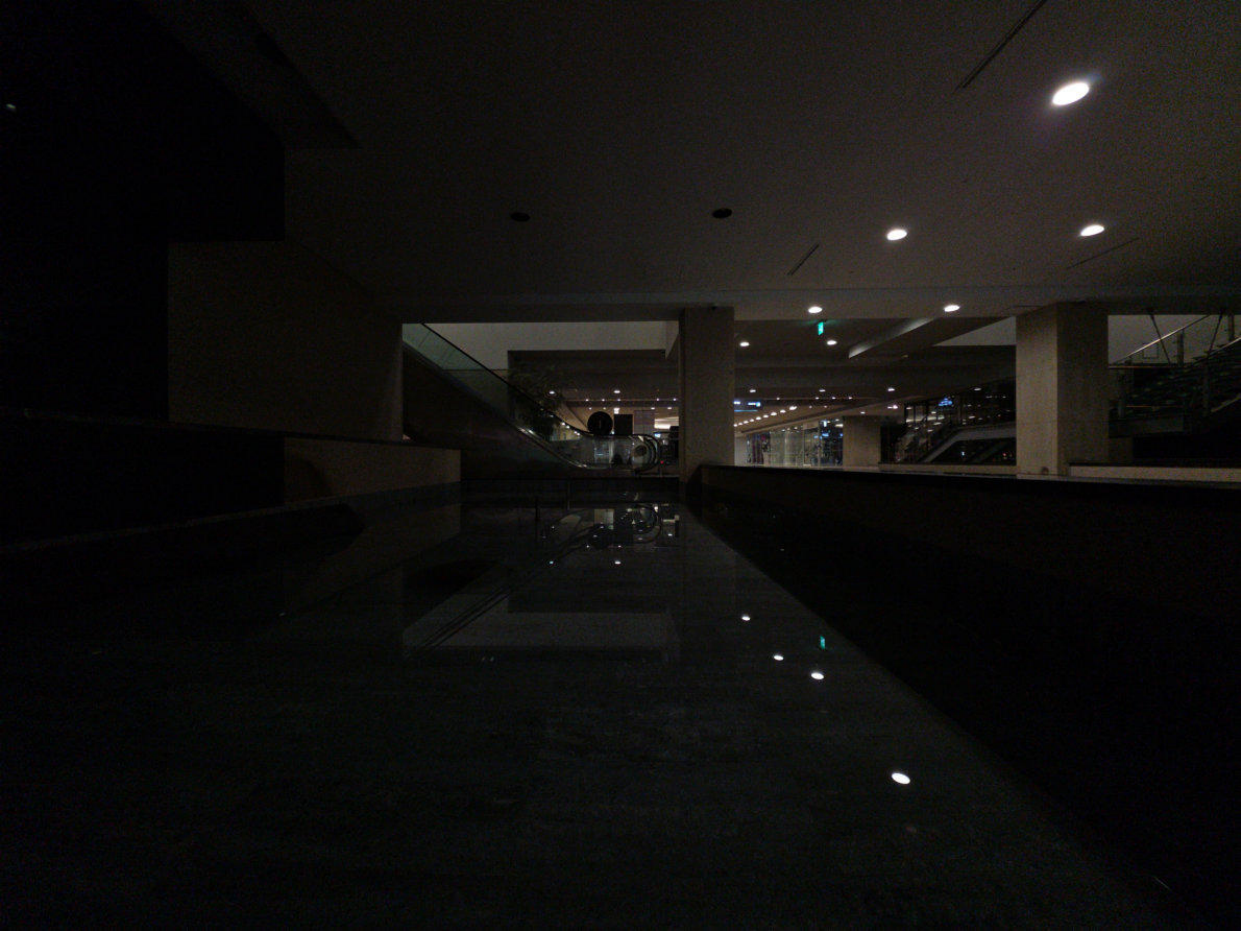} &
    \includegraphics[width=0.155\textwidth]{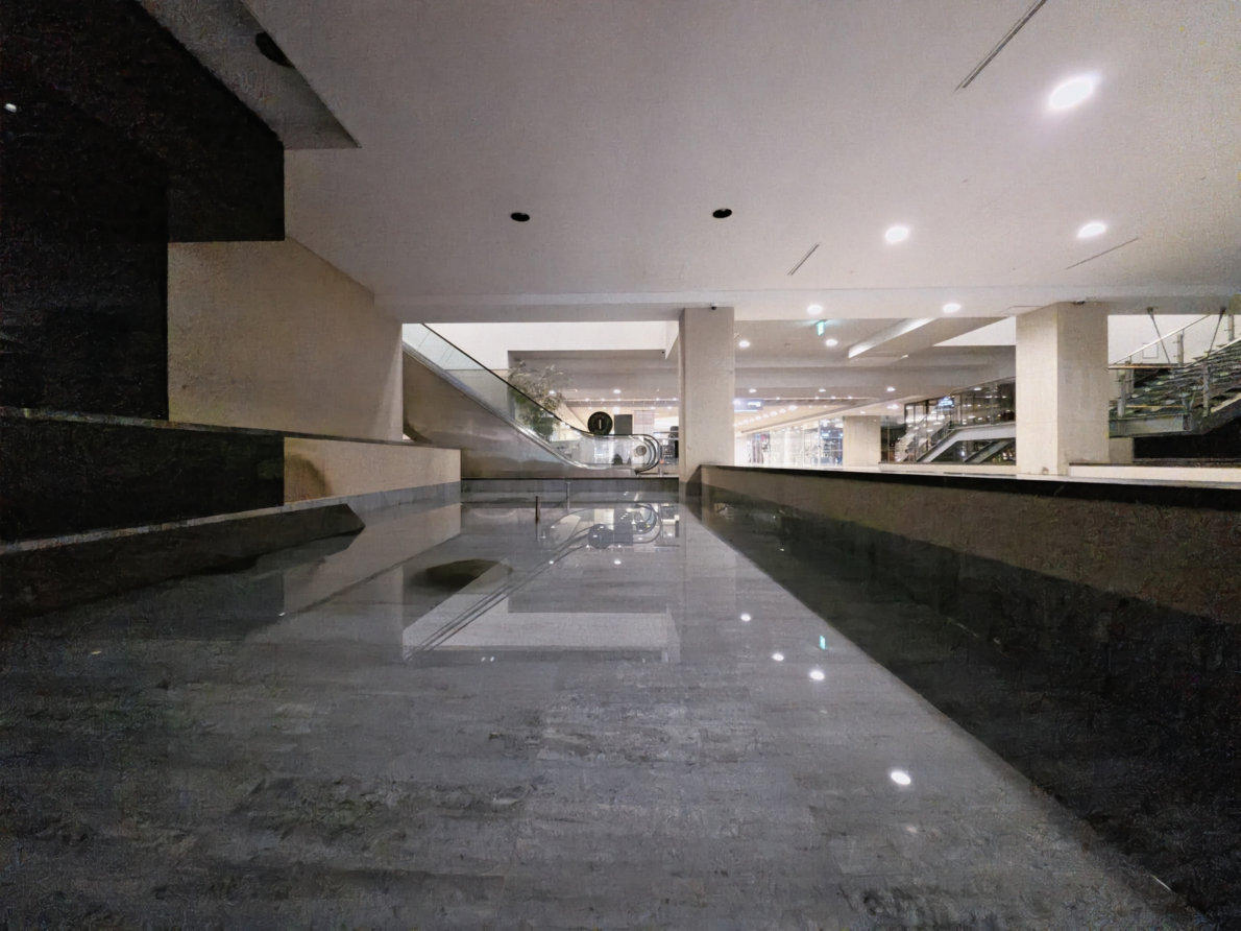} &
    \includegraphics[width=0.155\textwidth]{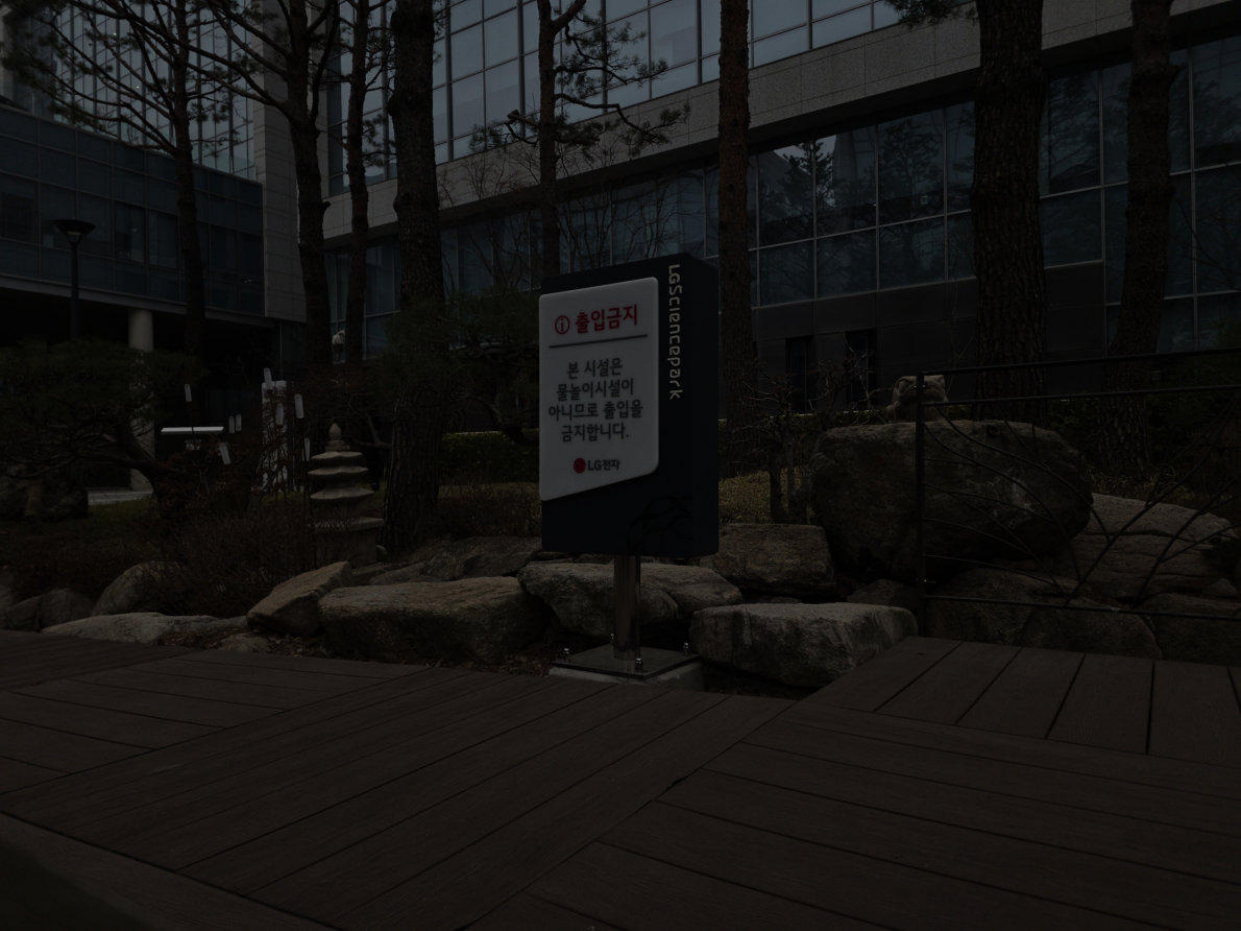} &
    \includegraphics[width=0.155\textwidth]{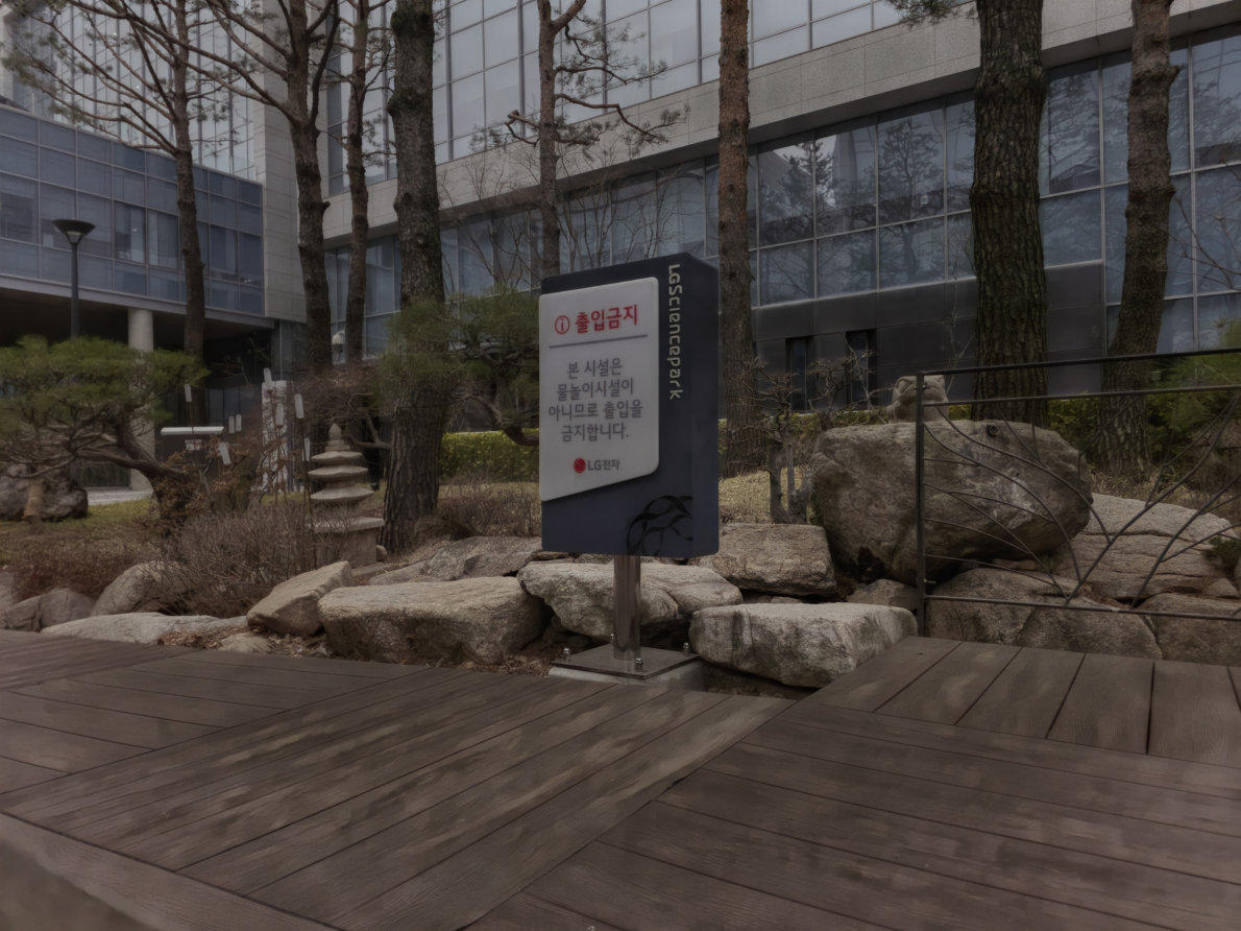} \\
    \raisebox{0.4cm}{\rotatebox{90}{\scriptsize Dehazing}} &
    \includegraphics[width=0.155\textwidth]{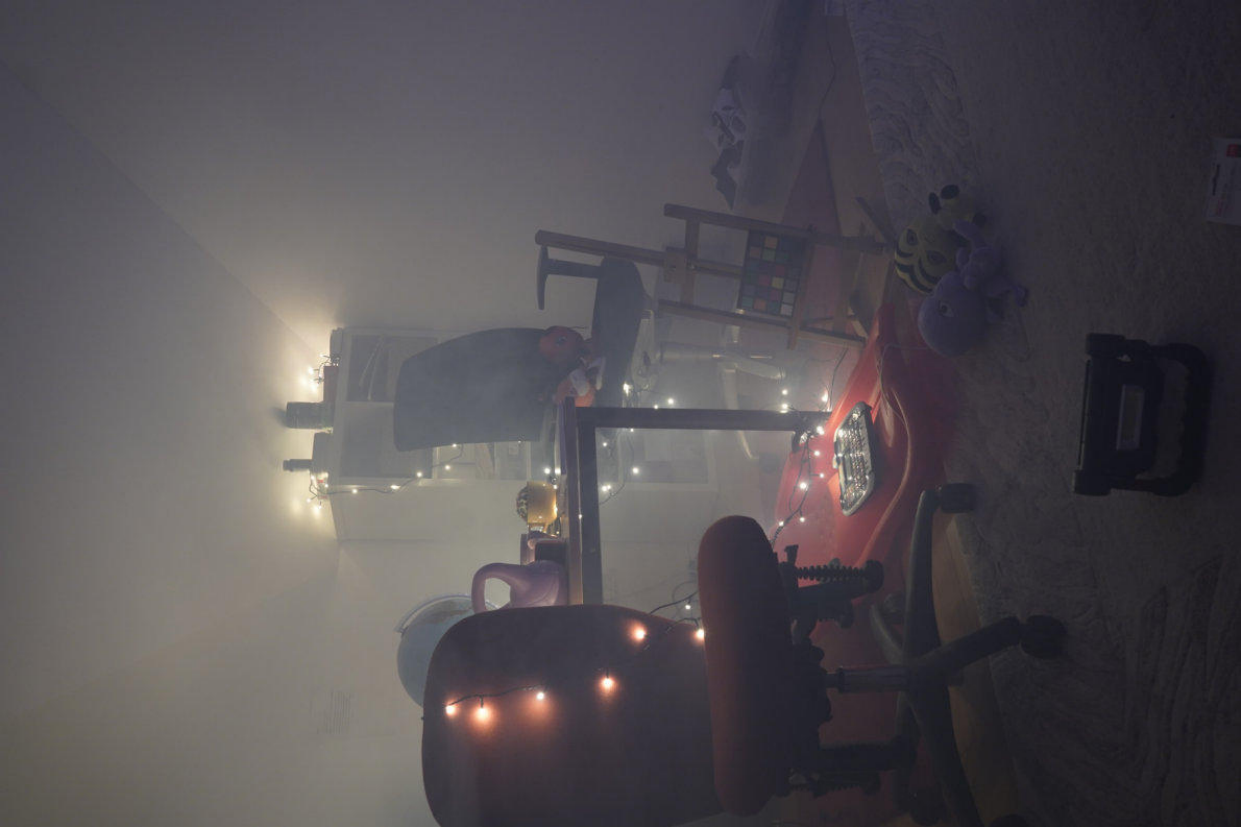} &
    \includegraphics[width=0.155\textwidth]{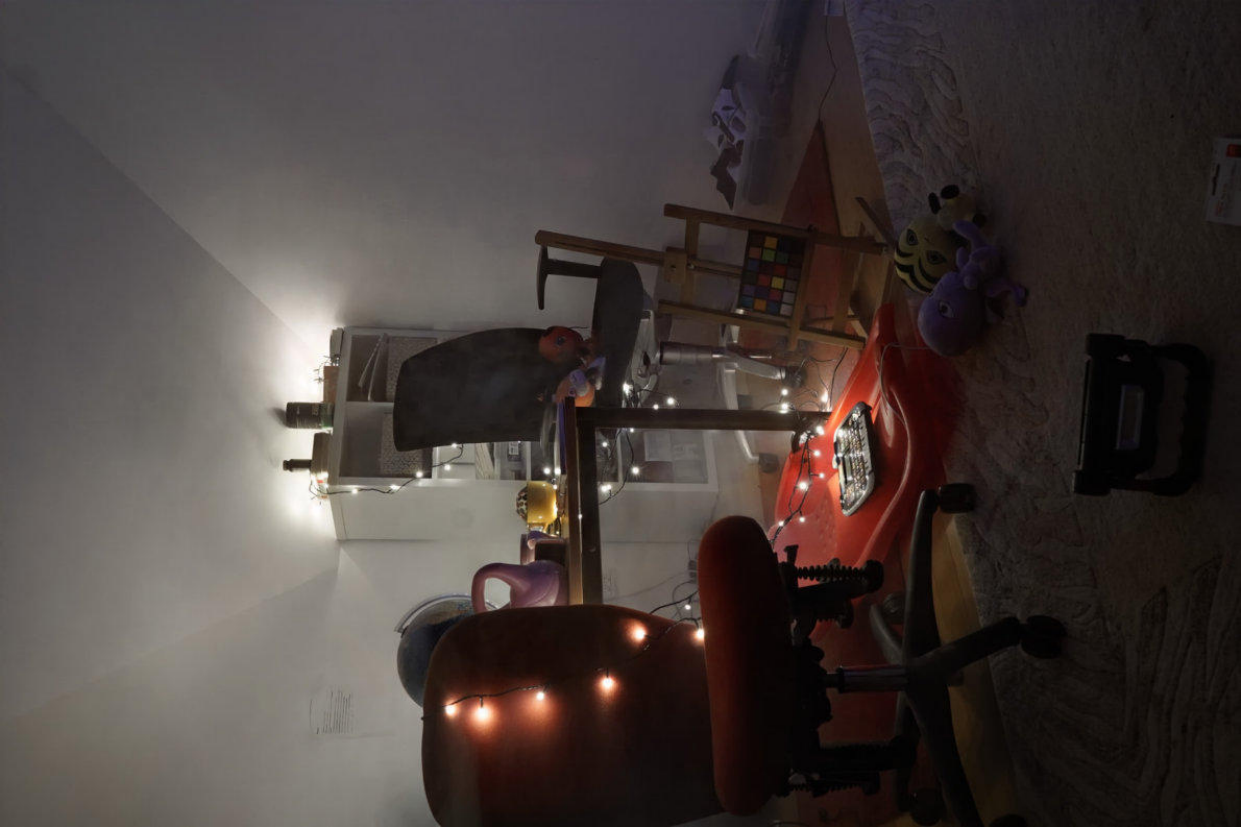} &
    \includegraphics[width=0.155\textwidth]{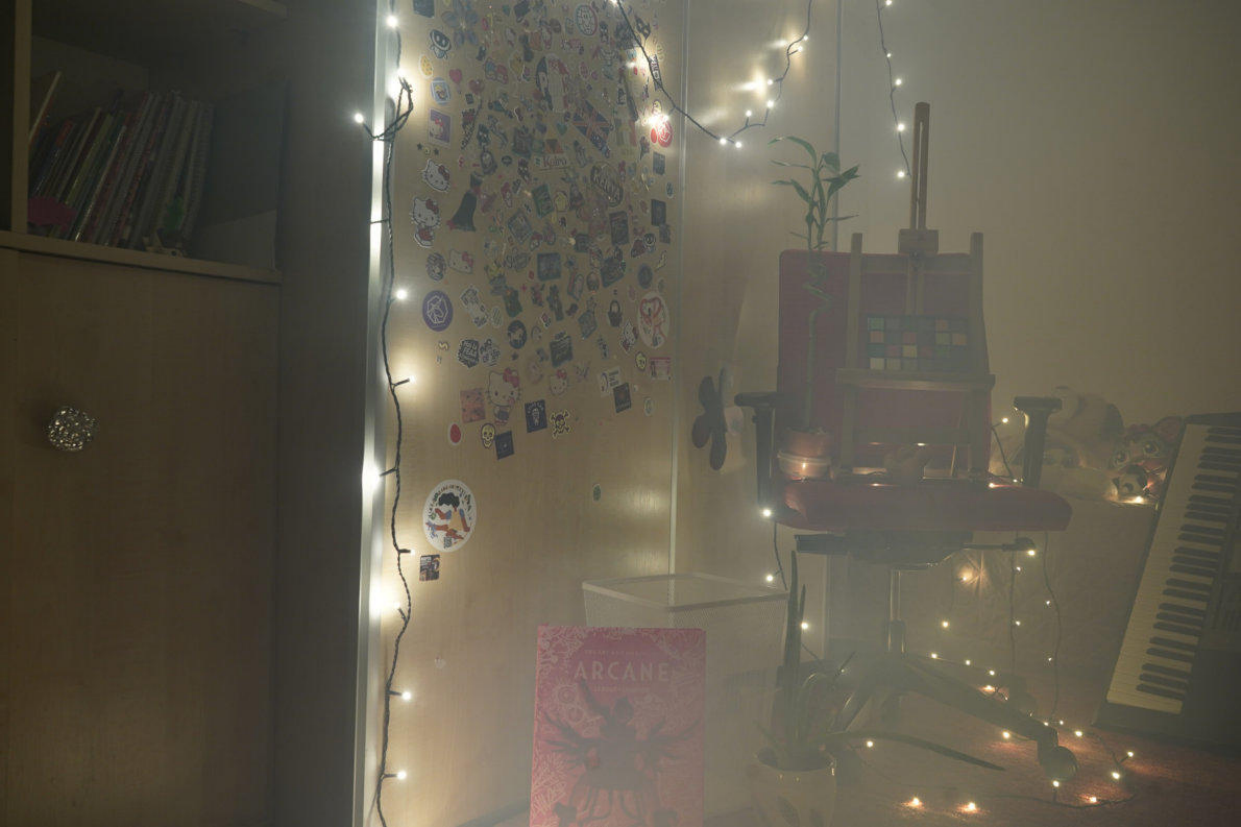} &
    \includegraphics[width=0.155\textwidth]{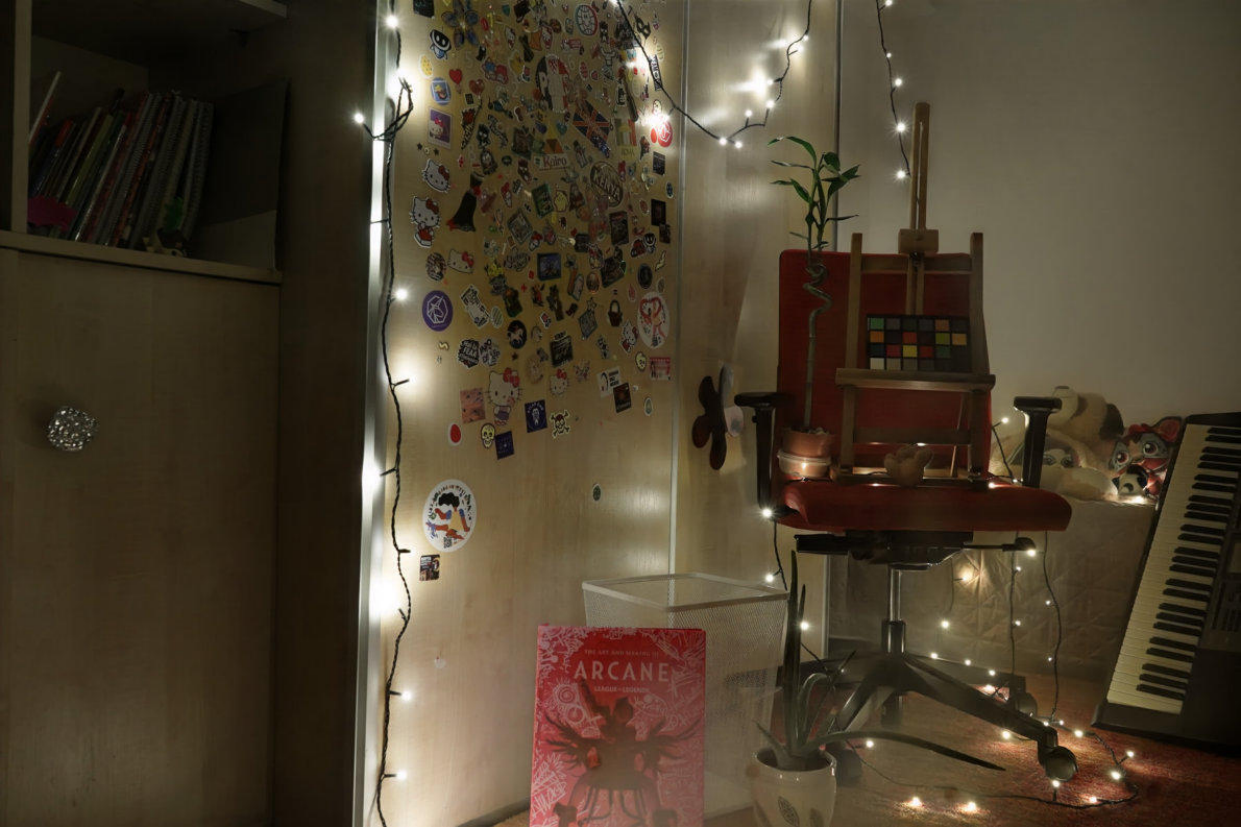} &
    \includegraphics[width=0.155\textwidth]{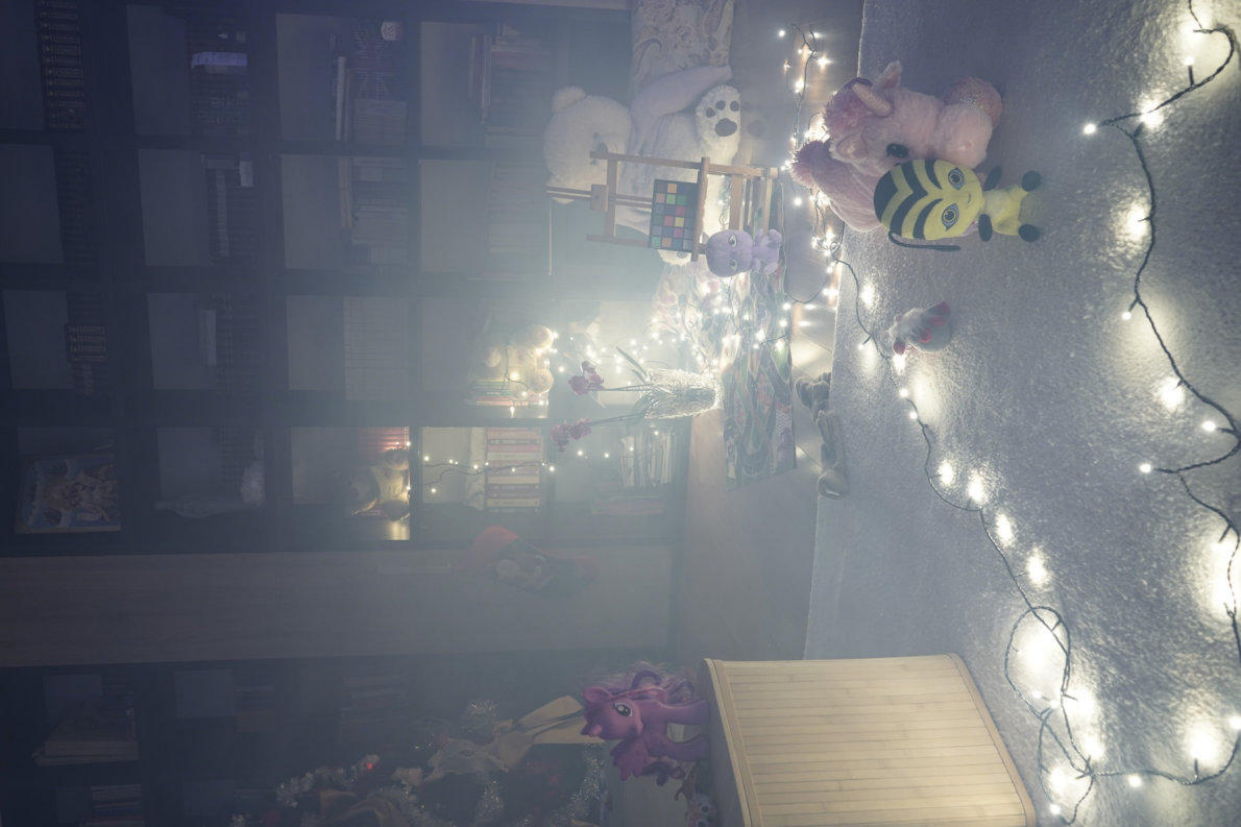} &
    \includegraphics[width=0.155\textwidth]{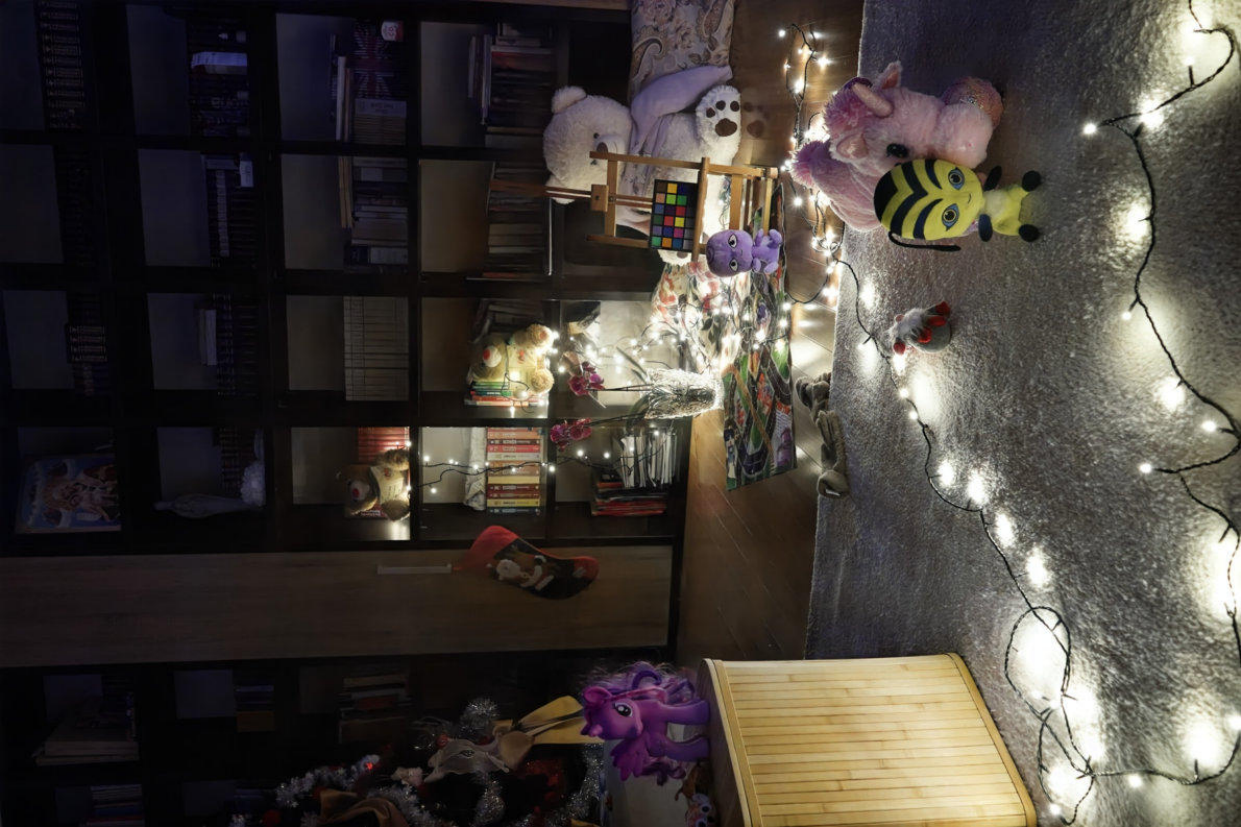} \\
    \raisebox{0.4cm}{\rotatebox{90}{\scriptsize Deraining}} &
    \includegraphics[width=0.155\textwidth]{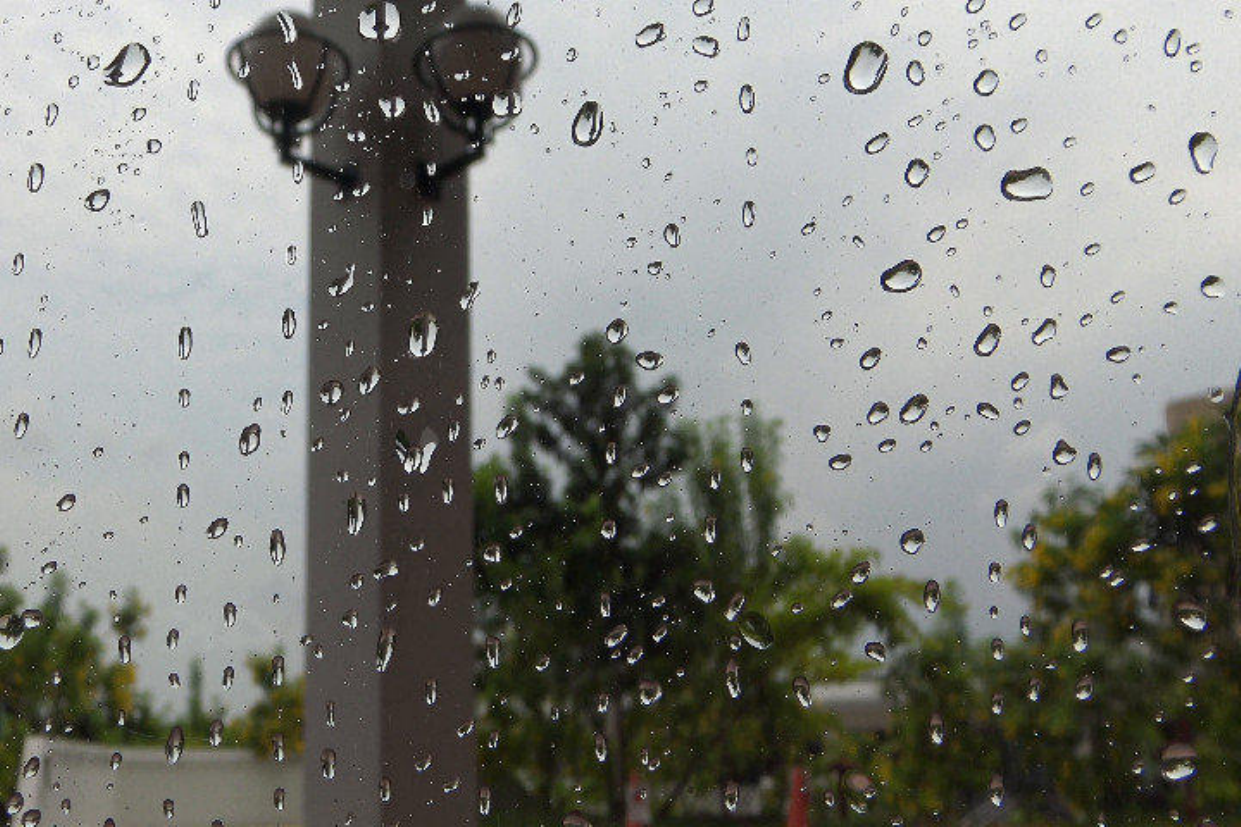} &
    \includegraphics[width=0.155\textwidth]{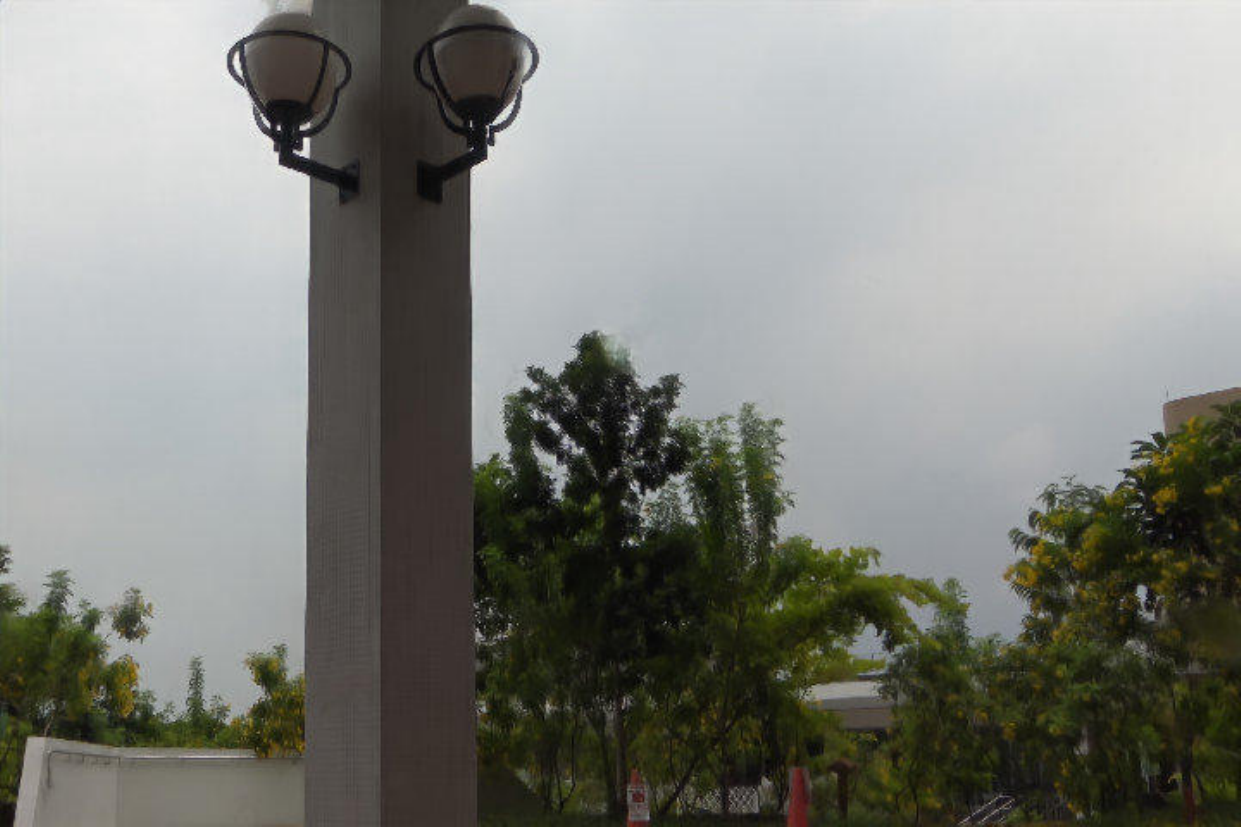} &
    \includegraphics[width=0.155\textwidth]{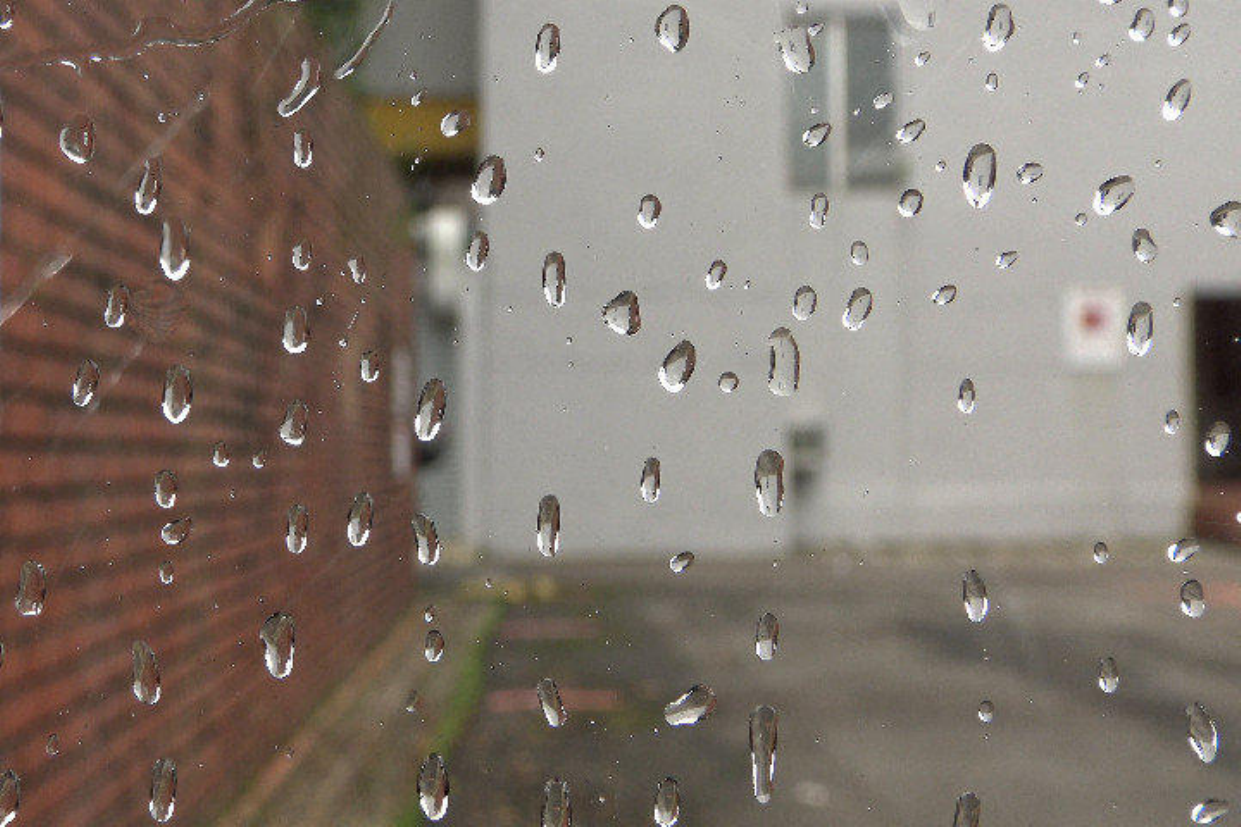} &
    \includegraphics[width=0.155\textwidth]{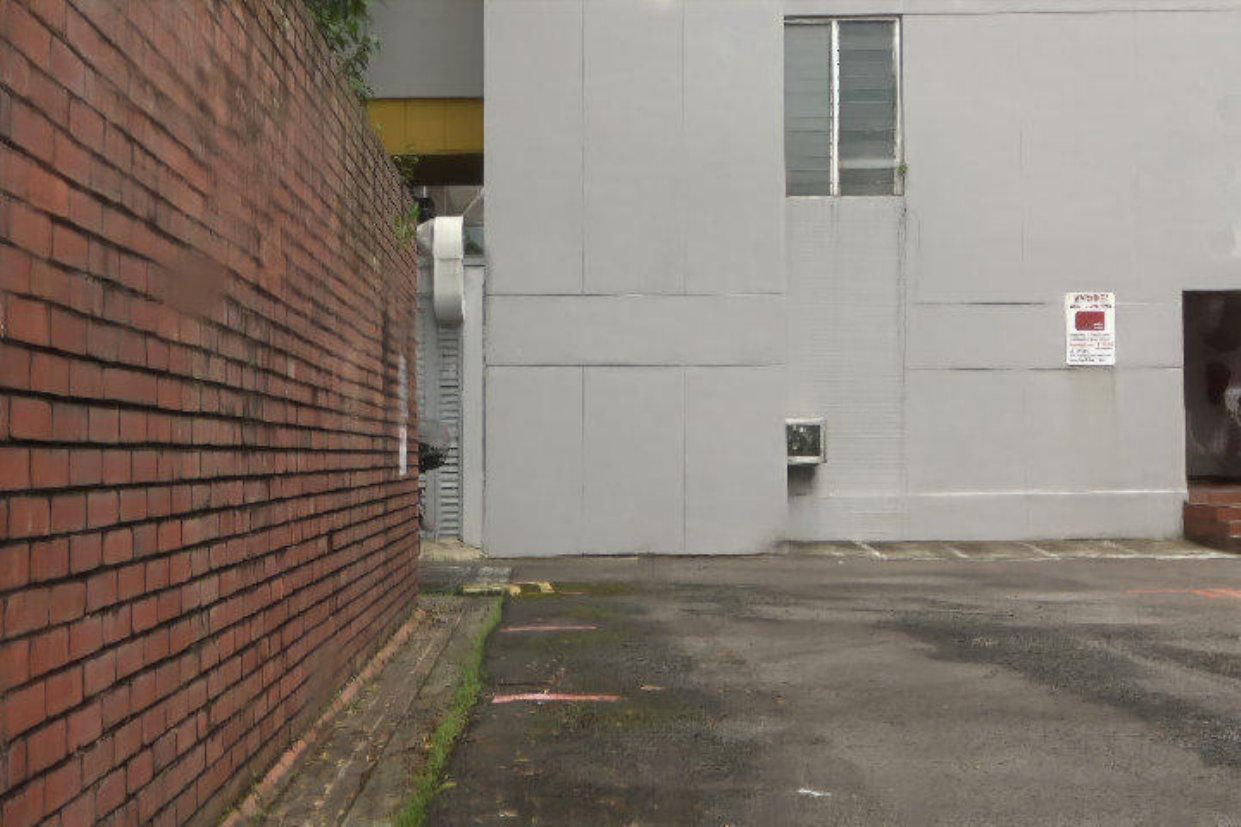} &
    \includegraphics[width=0.155\textwidth]{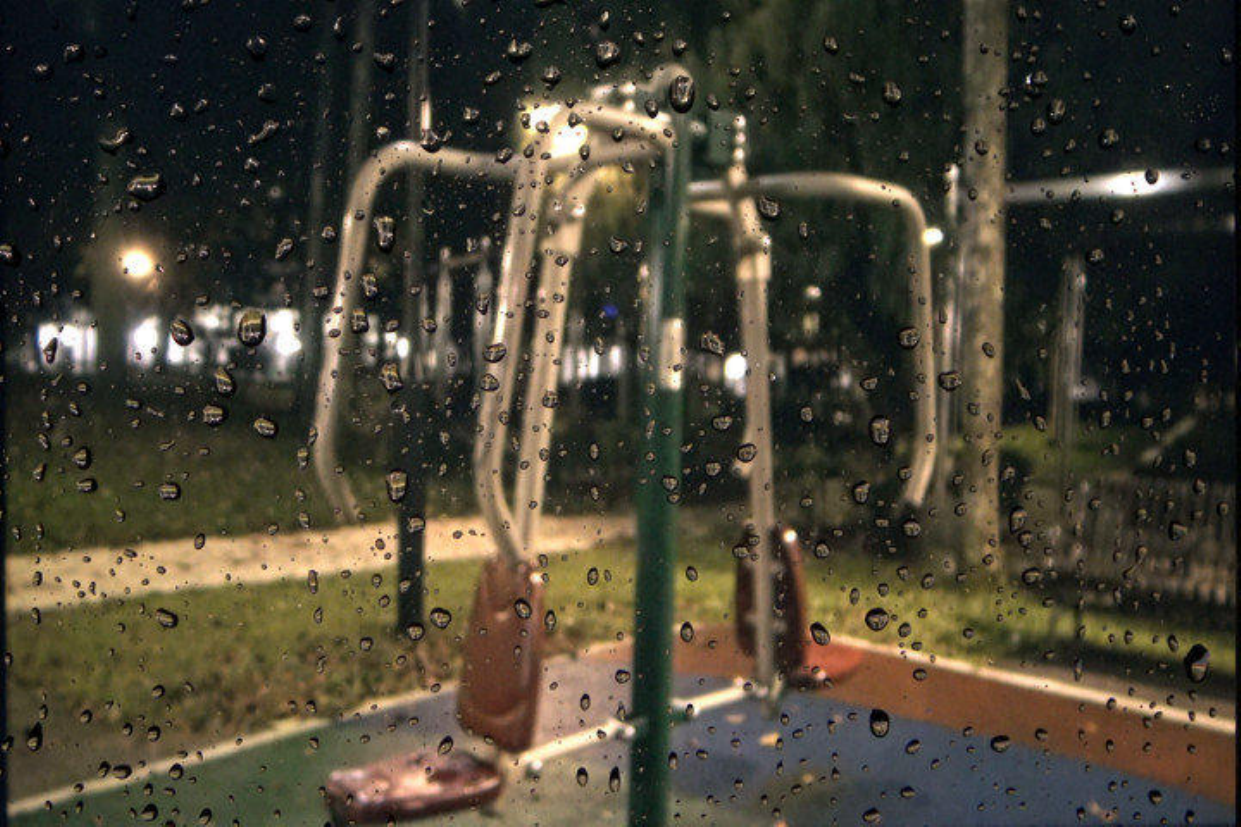} &
    \includegraphics[width=0.155\textwidth]{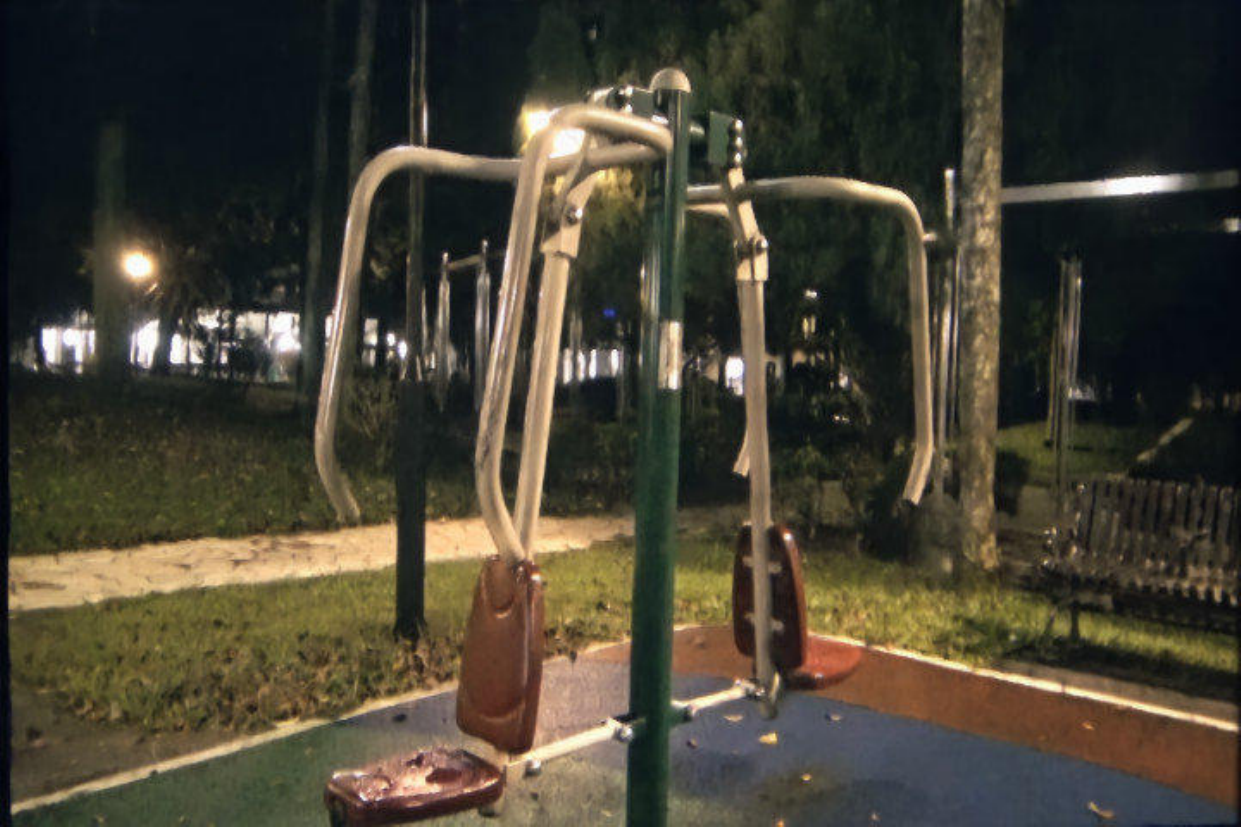} \\
    \raisebox{0.3cm}{\rotatebox{90}{\scriptsize Deshadowing}} &
    \includegraphics[width=0.155\textwidth]{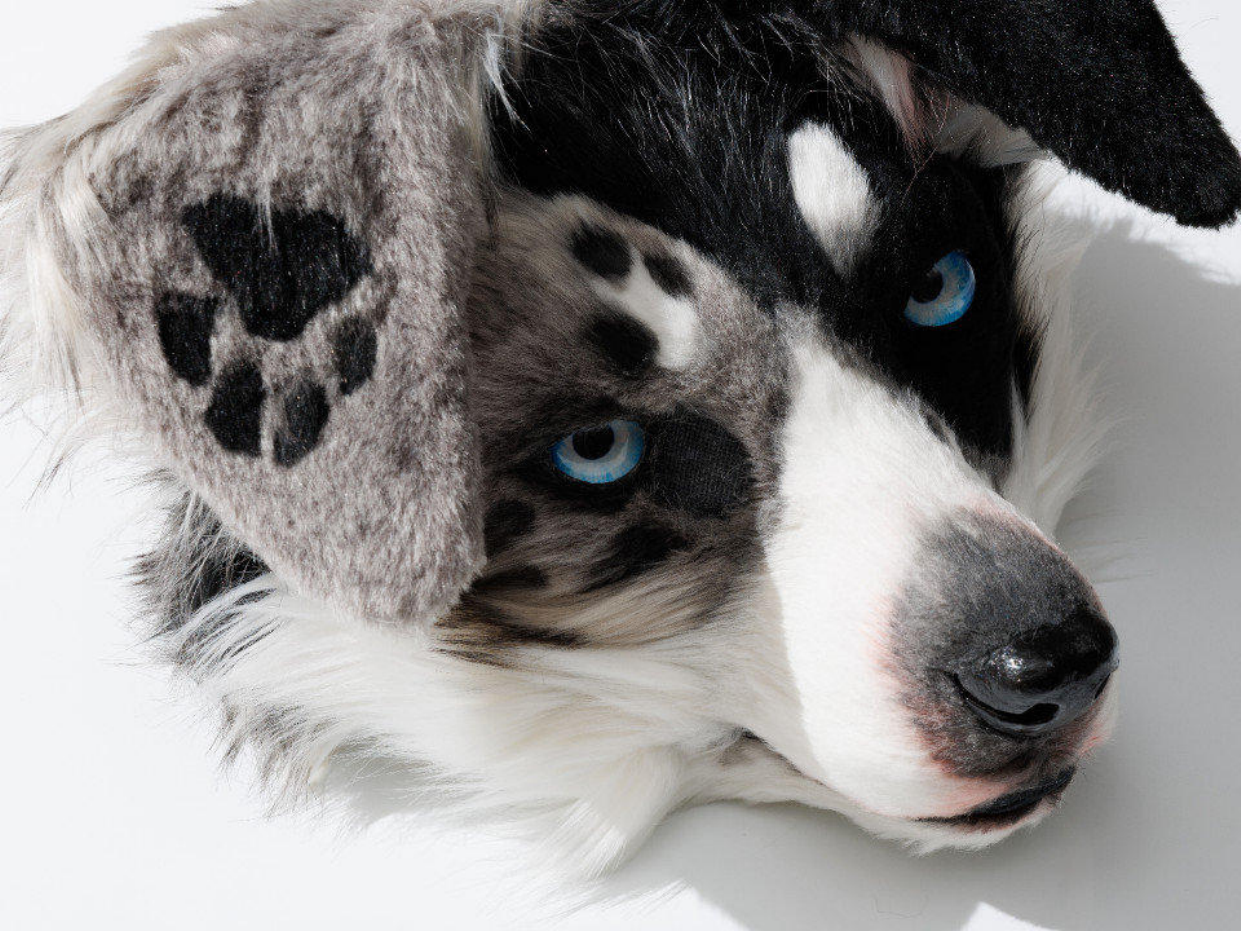} &
    \includegraphics[width=0.155\textwidth]{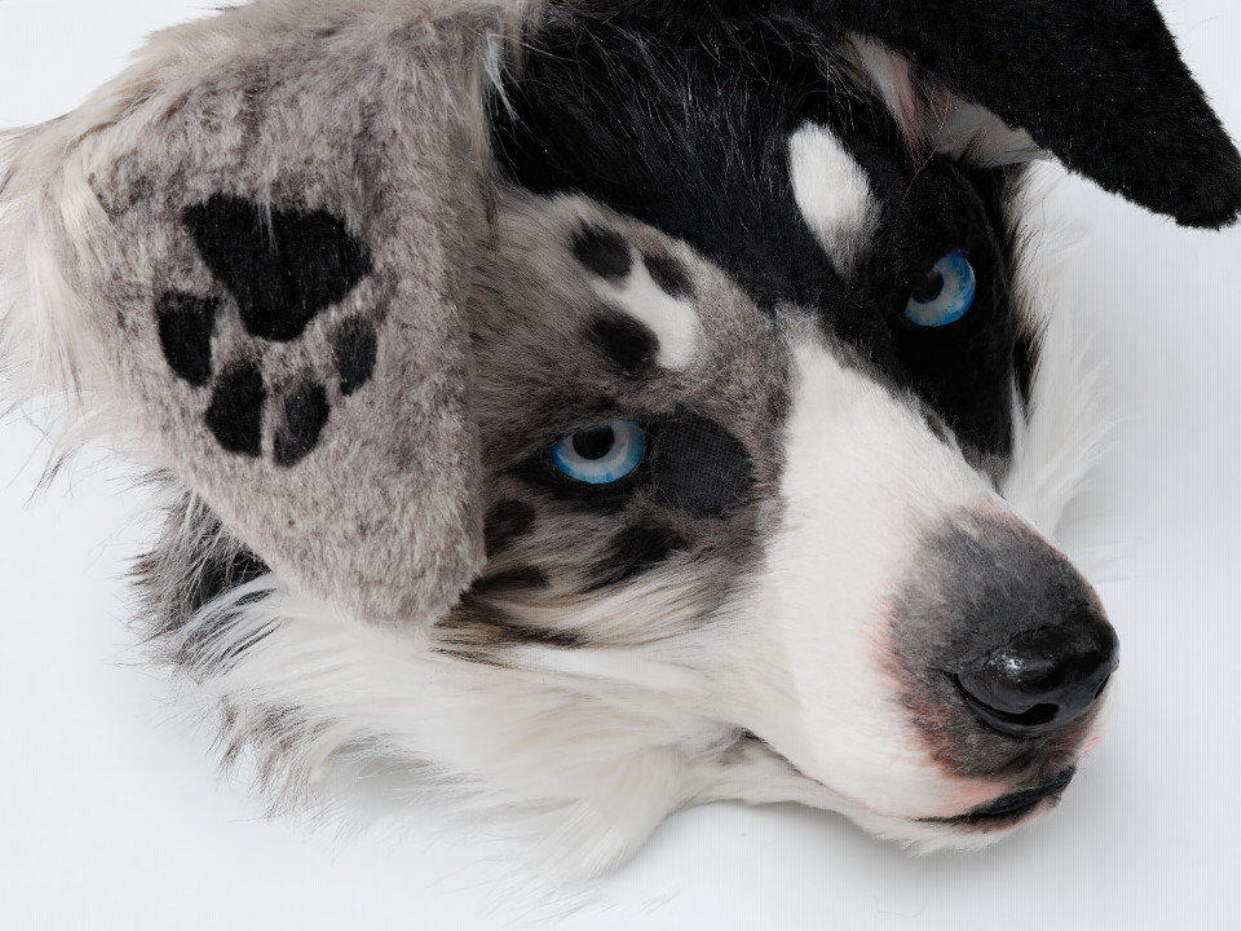} &
    \includegraphics[width=0.155\textwidth]{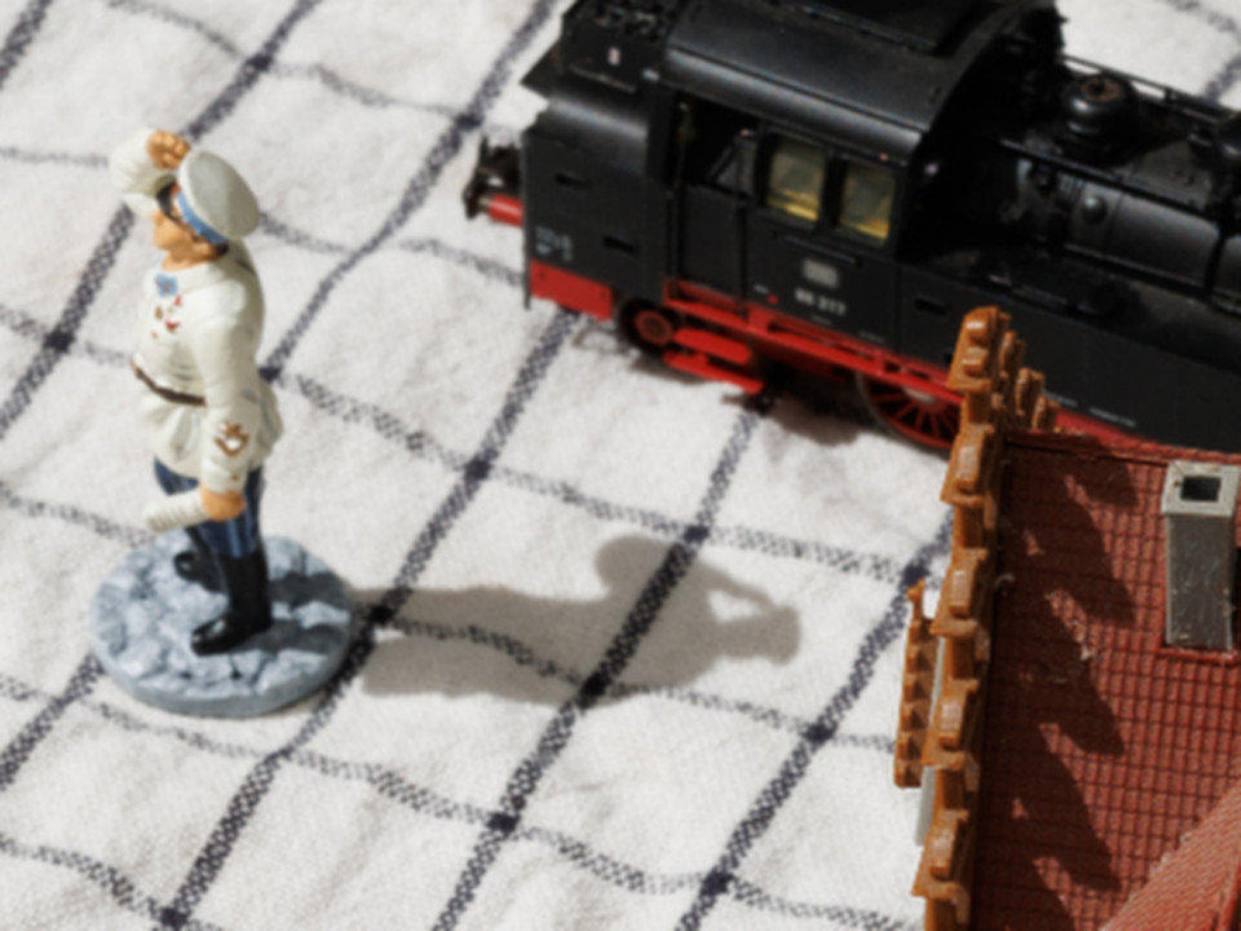} &
    \includegraphics[width=0.155\textwidth]{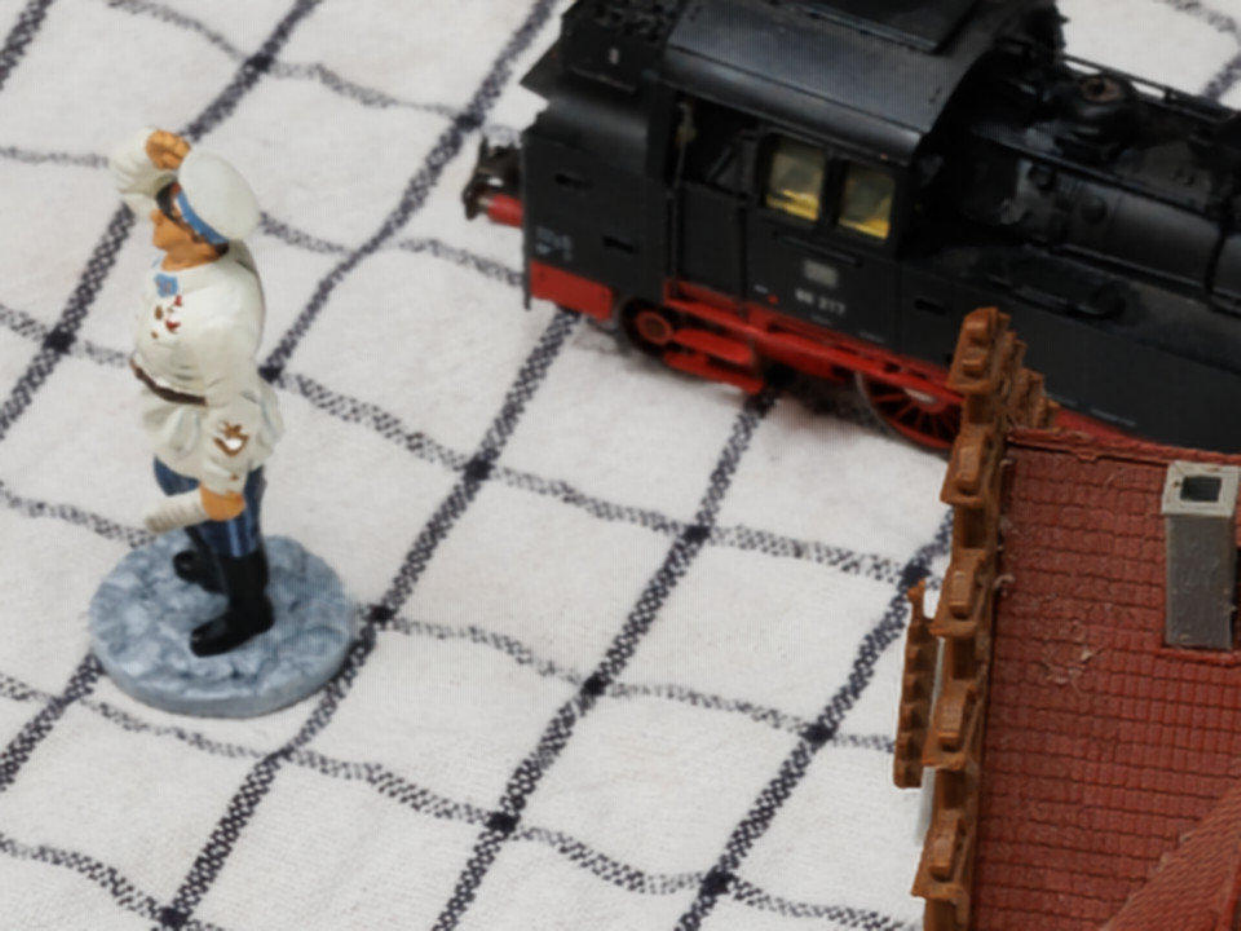} &
    \includegraphics[width=0.155\textwidth]{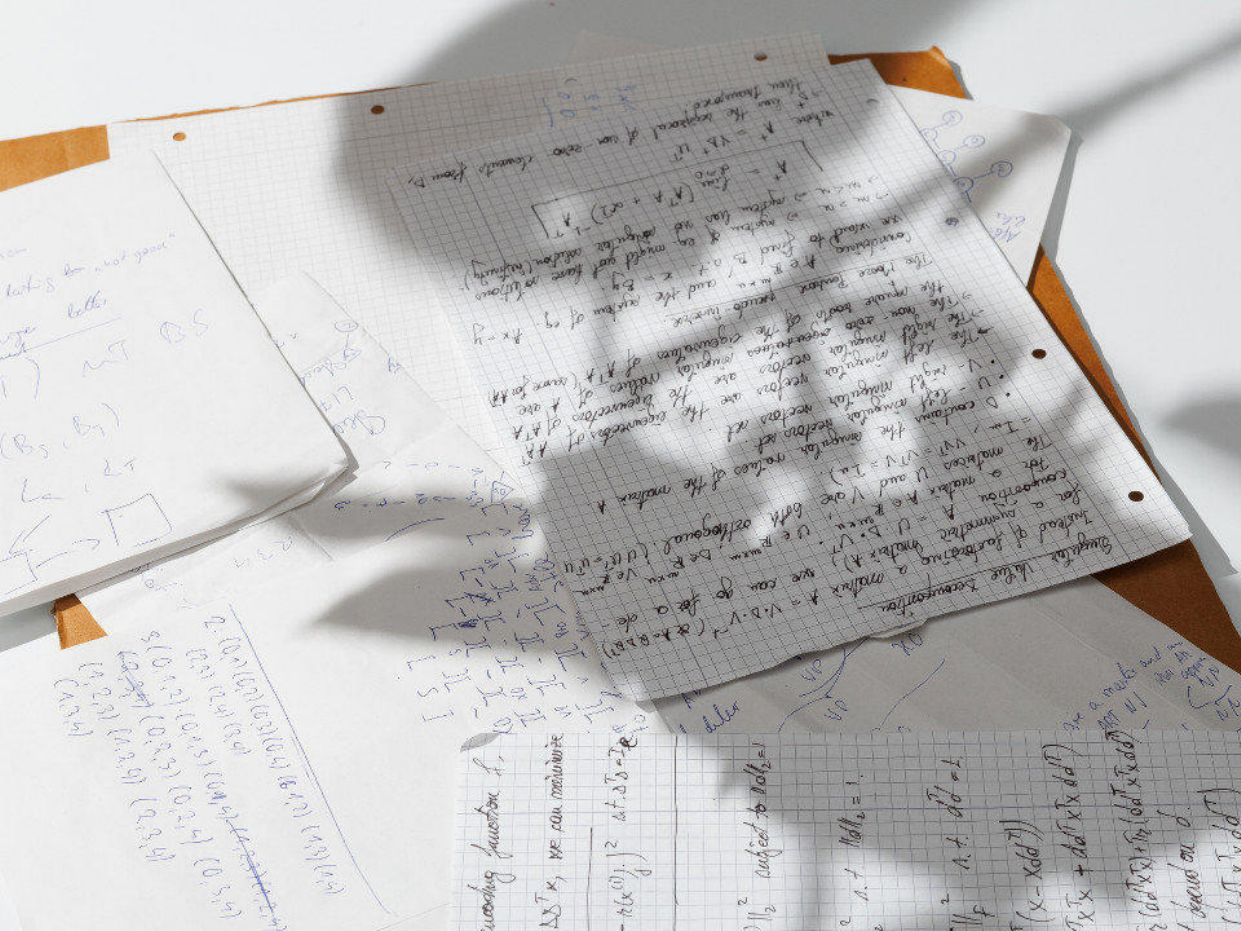} &
    \includegraphics[width=0.155\textwidth]{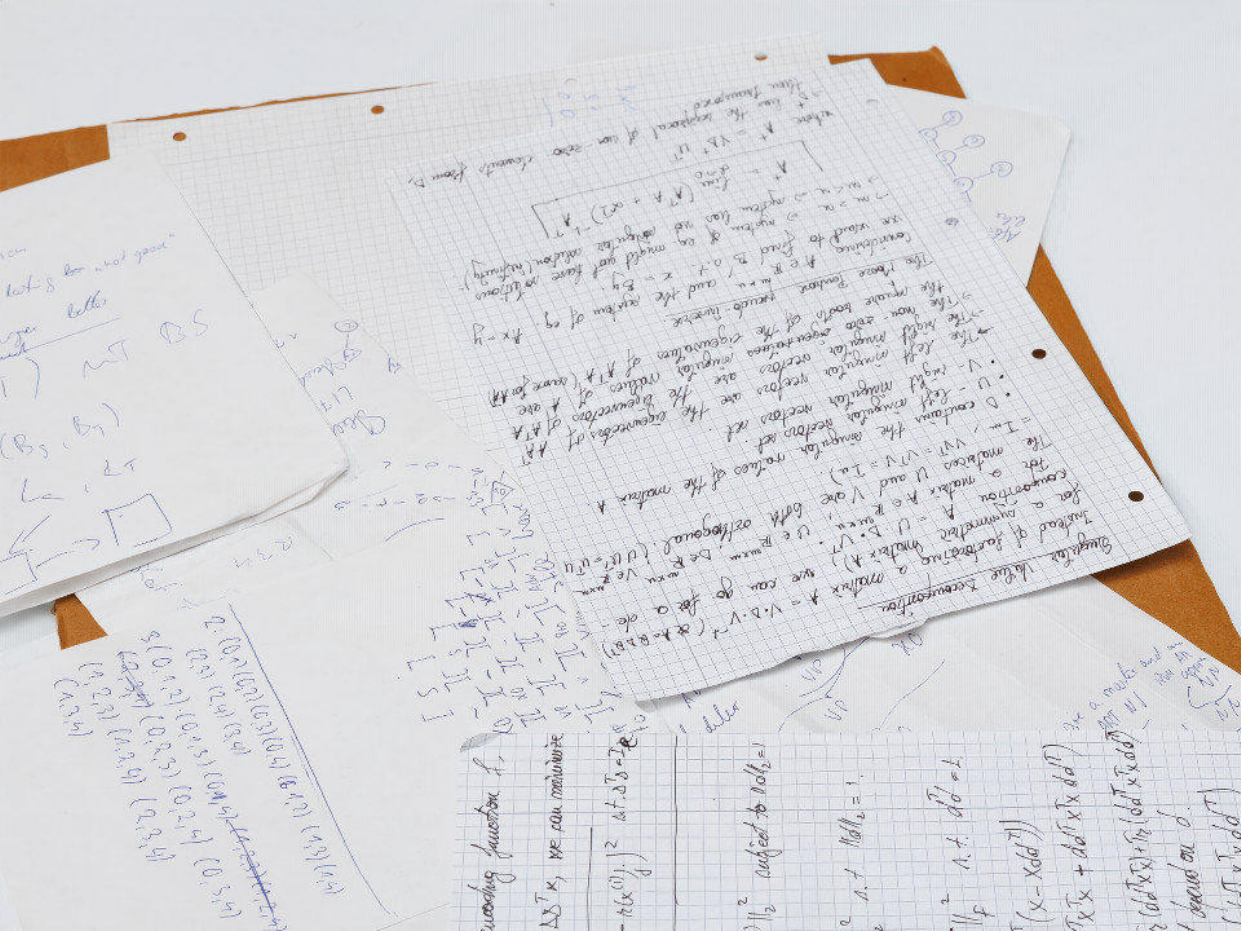} \\
    \end{tabular}
    \setlength{\parskip}{0mm}
\end{minipage}
\vspace{-3mm}
\captionof{figure}{Visual results of our proposed method on multiple image restoration tasks for NTIRE 2026 test sets, demonstrating its generalizability across low-light enhancement, dehazing, deraining, and shadow removal.}
\vspace{3mm}
\label{fig:teaser}
\end{center}}]
}

\begin{abstract}
We propose RetinexDualV2, a unified, physically grounded dual-branch framework for diverse Ultra-High-Definition (UHD) image restoration. Unlike generic models, our method employs a Task-Specific Physical Grounding Module (TS-PGM) to extract degradation-aware priors (e.g., rain masks and dark channels). These explicitly guide a Retinex decomposition network via a novel Physical-Conditioned Multi-head Self-Attention (PC-MSA) mechanism, enabling robust reflection and illumination correction. This physical conditioning allows a single architecture to handle various complex degradations seamlessly, without task-specific structural modifications. RetinexDualV2 demonstrates exceptional generalizability, securing 4\textsuperscript{th} place in the NTIRE 2026 Day and Night Raindrop Removal Challenge and 5\textsuperscript{th} place in the Joint Noise Low-light Enhancement (JNLLIE) Challenge. Extensive experiments confirm the state-of-the-art performance and efficiency of our physically motivated approach. 
Code is available at \url{https://github.com/ErrorLogic1211/RetinexDual/tree/master/RetinexDualV2}
\end{abstract}    
\section{Introduction}
\label{sec:intro}

Although imaging sensors and displays have advanced significantly to deliver highly detailed and sophisticated Ultra-High-Definition (UHD) images, restoring captured information distorted or degraded by light exposure, haze, shadow, or weather conditions has grown increasingly difficult. This causes a bottleneck for more advanced tasks, such as object detection, semantic segmentation, and visual reasoning. Accordingly, image restoration (IR) became more crucial not only for the plausibility of the image but also for the improvement of computer vision in general.

Numerous learning-based methods have emerged to tackle image restoration challenges \cite{lv2024fourier,9710307,cai2023retinexformer}, substantially improving the removal of various degradation artifacts. Nevertheless, these techniques struggle to process UHD images effectively, as the enormous data volume and resolution demand excessive computational resources. This constraint has become a major obstacle in achieving high-quality UHD image restoration.

To advance UHD image restoration (IR), several dedicated research efforts have been developed \cite{wang2024uhdformer,Li2023ICLR,yu2022towards,Zhao_2025_CVPR}.
Some approaches rely on a Downsampling-Enhancement-Upsampling paradigm \cite{wang2024uhdformer,Li2023ICLR}, which lowers computational demands by processing features in a reduced-resolution space before reconstructing full resolution. However, this strategy inevitably discards important high-frequency details because the resampling steps (downsampling and upsampling) are decoupled from the core enhancement process \cite{Yu_2024_CVPR}.
Other methods instead emphasize frequency-domain processing, decomposing the image across multiple frequency bands to exploit global contextual information and simplify the modeling of certain degradation patterns that become more apparent in the frequency domain \cite{zou2024wavemamba, Zhao_2025_CVPR}. While effective in many scenarios, this frequency-based strategy lacks universal applicability; in cases dominated by localized degradations, the spatial domain often provides a more direct and accurate depiction of the correspondence between degraded and clean image regions \cite{kishawy2025retinexdualretinexbaseddualnature}.

Alternatively, Retinexformer \cite{cai2023retinexformer} introduced a Retinex-based model for low-light enhancement, inspiring several follow-up works \cite{10.1007/978-981-96-6596-9_30,LIU20251969}. Although these methods achieved top performance in low-light restoration, they generally treat reflectance and illumination as homogeneous components and do not extend Retinex theory into a more general framework for broader image restoration tasks beyond illumination/color correction. RetinexDual \cite{kishawy2025retinexdualretinexbaseddualnature} addressed this problem by proposing an approach that recognizes the dual nature of retinex decomposition. However, it faces challenges in certain complex scenarios as there is no physical-grounding for the learning process.

Another promising direction lies in incorporating physics-informed strategies that embed domain-specific physical models or priors directly into the deep learning framework, such as the dark channel prior in dehazing~\cite{10550844}, the illumination invariant prior in low-light enhancement~\cite{quadprior}, or the region of interest in degradation mechanisms~\cite{10656337}. This approach leverages established knowledge of image formation processes to regularize training, guide feature learning, and constrain outputs toward physically plausible solutions.

Based on this, we propose RetinexDualV2, a physically grounded evolution of RetinexDual that preserves its effective dual-branch handling of reflectance and illumination while infusing task-specific physical priors to guide restoration. 
Specifically, the degraded input is fed into two parallel extractors: a Retinex decomposer and a task-specific physical grounding module. The resulting reflectance and illumination components are then processed through Physical-Grounded Scale Attentive Mamba (PG-SAMBA) for spatially adaptive, detail-preserving reflectance correction and Physical-Grounded Frequency Illumination Adaptor (PG-FIA) for physically consistent illumination adjustment. Within each layer, the physical features are incorporated using Physical-Conditioned Multi-head Self-Attention (PC-MSA).
The main contributions are summarized as follows:
\begin{itemize}
    \item We introduce \textbf{RetinexDualV2}, a physically grounded extension of RetinexDual that assesses the restoration of retinex components inside the specialized sub-networks. 
    \item We implement a task-specific physical grounding module that extracts physical characteristics from the input as dark channel prior in dehazing, rain drop ROI in deraining, and illumination-invariant edge detection in low light, and shadow-invariant prior for shadow removal.
    \item In PC-MSA, we inject the physical map into the restoration process using a modulated self-attention mechanism to regularize outputs and ensure physical plausibility.
    \item Extensive evaluations across multiple NTIRE 2026 challenges demonstrate our model's robust generalization and competitive performance. Specifically, RetinexDualV2 secured the 4\textsuperscript{th} place in the Day and Night Raindrop Removal Challenge and the 5\textsuperscript{th} place in the Joint Noise Low-light Enhancement Challenge, while maintaining an efficient parameter footprint of 4.8M.
\end{itemize}

\section{Related Work}
\label{sec:related}

\textbf{Ultra-High-Definition Image Restoration.} Extensive research addresses UHD restoration \cite{wang2024uhdformer, Zhao_2025_CVPR, zou2024wavemamba, Wang_Zhang_Shen_Luo_Stenger_Lu_2023}. Approaches like LLFormer \cite{Wang_Zhang_Shen_Luo_Stenger_Lu_2023} retain UHD details via multi-scale attention but demand massive computational resources. Alternatively, frequency-domain methods like Wave-Mamba \cite{zou2024wavemamba} and ERR \cite{Zhao_2025_CVPR} mitigate computational bottlenecks by processing distinct spectral bands. However, they lack universal applicability; localized spatial degradations often require direct spatial-domain processing to accurately map corrupted regions to pristine counterparts \cite{kishawy2025retinexdualretinexbaseddualnature}.

\textbf{Retinex Theory in Image Restoration.} Recent Low-Light Image Enhancement (LLIE) increasingly adopts Retinex theory, treating illumination as the primary noise carrier \cite{cai2023retinexformer,10.1007/978-981-96-6596-9_30,LIU20251969}. While Retinexformer \cite{cai2023retinexformer} and its extensions \cite{10.1007/978-981-96-6596-9_30,LIU20251969,guo2025eretinex,NEURIPS2024_30699996} exhibit strong performance, they typically process illumination and reflectance homogeneously and neglect broader image restoration applications. RetinexDual \cite{kishawy2025retinexdualretinexbaseddualnature} introduced a dual-branch architecture tailored to each component, yet it struggles in specialized scenarios, indicating that pure Retinex theory requires explicit physical grounding for reliable universal restoration.

\textbf{Physical Grounding in Image Restoration.} Physical priors provide essential regularization for ill-posed restoration tasks. Methods explicitly modeling degradation mechanism, such as the dark channel prior for atmospheric scattering \cite{10550844} or log-chromaticity for illumination invariance \cite{quadprior}, effectively isolate clean signals. Furthermore, structural priors \cite{dong2025towards} geometrically constrain the restoration space. Integrating these explicit priors ensures strict radiometric consistency and robust edge preservation, guiding models toward physically plausible, resilient recovery where purely data-driven statistical mappings often fail.

\section{Methodology}
\label{sec:method}

\begin{figure*}
\centering
\includegraphics[width=\textwidth]{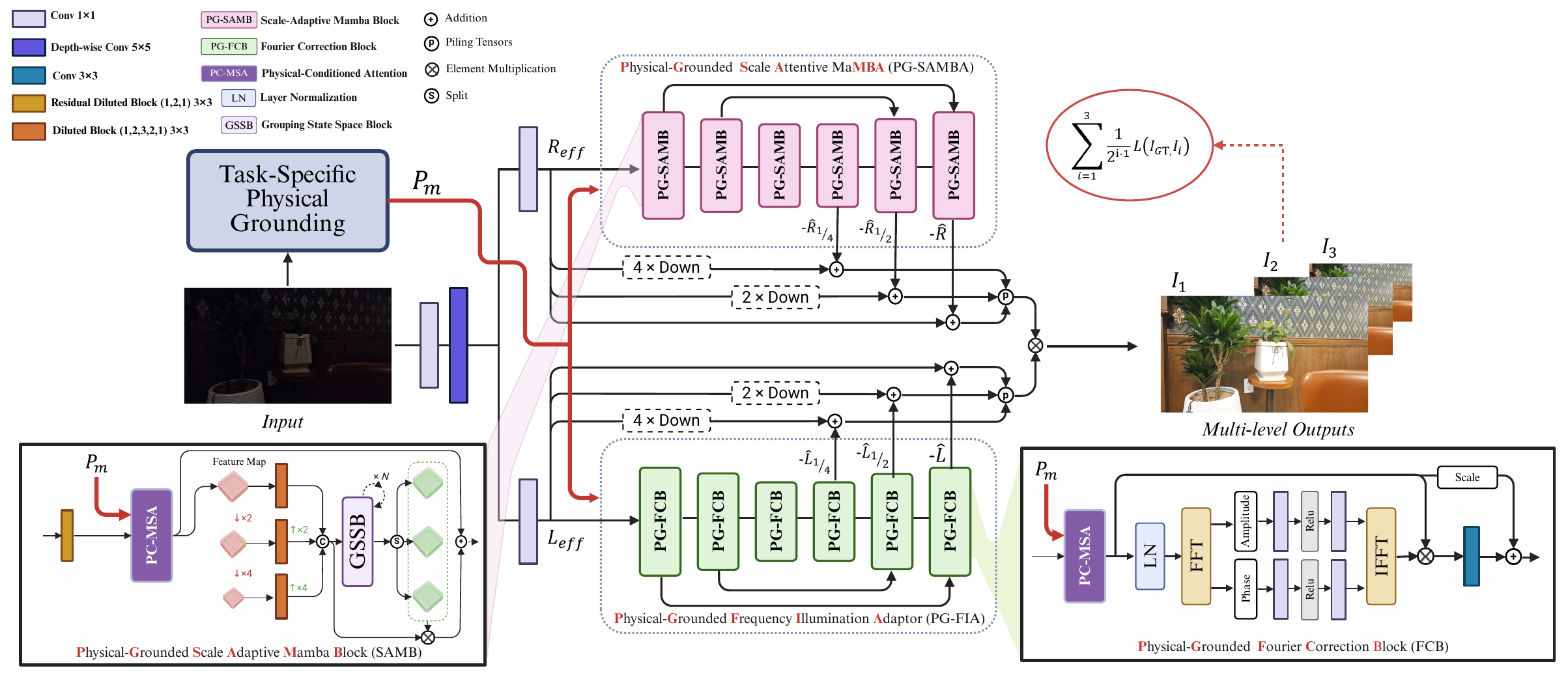} 
\caption{RetinexDualV2 Overview. Based on retinex theory, it decomposes the UHD image into \(R_{eff}\) and \(L_{eff}\) and operates on them using 2 sub-networks: Physical-Grounded Scale Attentive MaMBA (PG-SAMBA), and Physical-Grounded Frequency Illumination Adaptor (PG-FIA), respectively. Both branches receive guidance \(p_m\) from the Task-Specific Physical Grounding Module (TS-PGM). Noting that each output from a different level has a convolution layer before it for processing, which wasn't illustrated for simplicity.}
\label{fig:RD_overview}
\end{figure*}

Our key contribution in this paper is the design of a task-specific physical grounding module (TS-PGM) \ref{subsec:TS-PGM} that can be adjusted to the nature of degradation while integrating this physical prior with the dual branch of retinex decomposition using a Physical-conditioned Multi-head Self-Attention (PC-MSA) in the RetinexDualV2 model \ref{subsec:Retiv2}. The proposed RetinexDualV2 is presented in Fig. \ref{fig:RD_overview}, which contains two sub-networks, PG-SAMBA and PG-FIA, retinex decomposer, and a task-based physical grounding module. In each of the branches, Physical-Grounded Scale Adaptive Mamba Block (PG-SAMB) and Physical-Grounded Fourier Correction Block (PG-FCB) are responsible for restoring the reflectance and illumination components, respectively, while relying on the physical conditioning provided by (TS-PGM). 
\begin{figure}
  \centering
  \small
  \setlength{\tabcolsep}{1pt}
  \newcommand{\priorsImgH}{0.0625\textheight}
  \begin{tabular}{c c@{\hspace{3pt}} c@{\hspace{3pt}} c}
    \includegraphics[height=\priorsImgH,keepaspectratio]{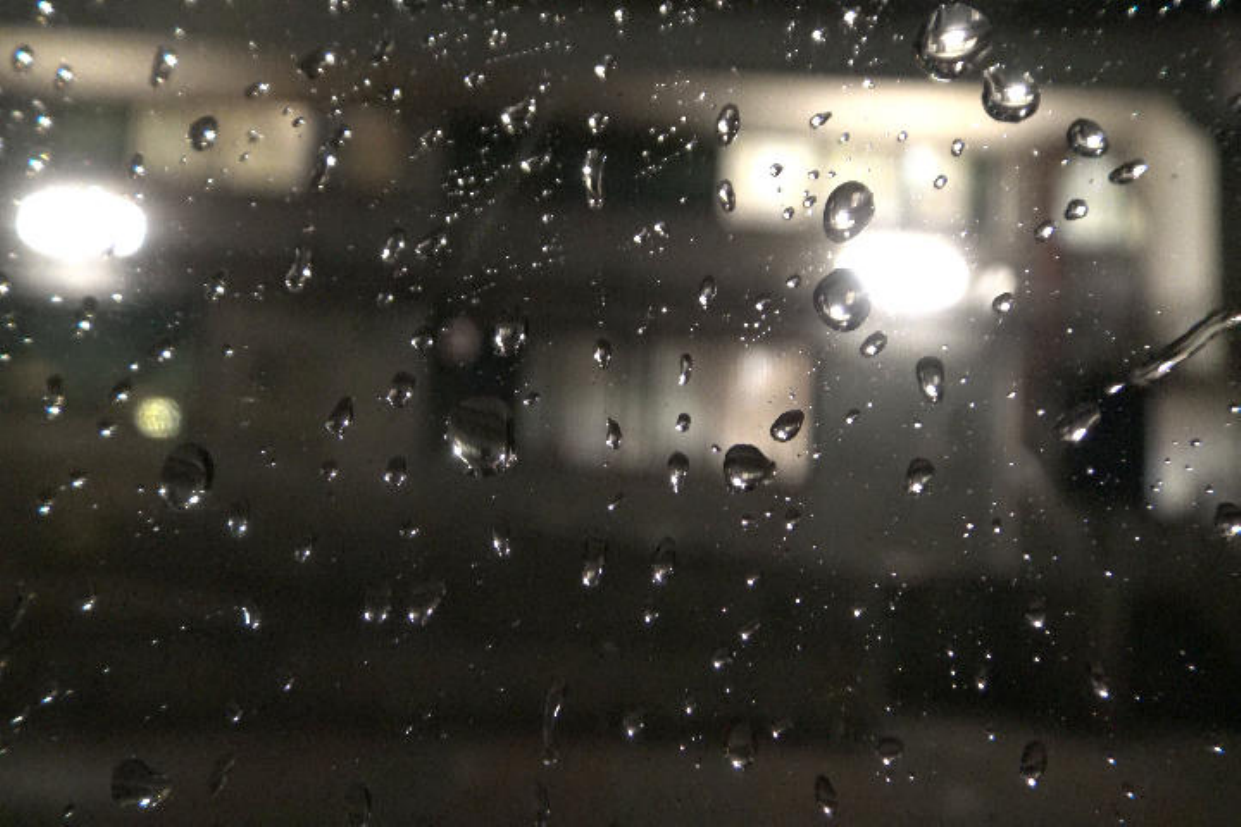} &
    \includegraphics[height=\priorsImgH,keepaspectratio]{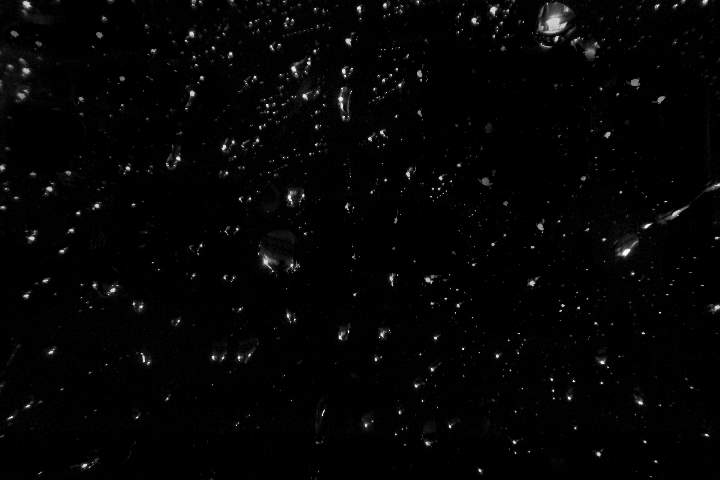} &
    \includegraphics[height=\priorsImgH,keepaspectratio]{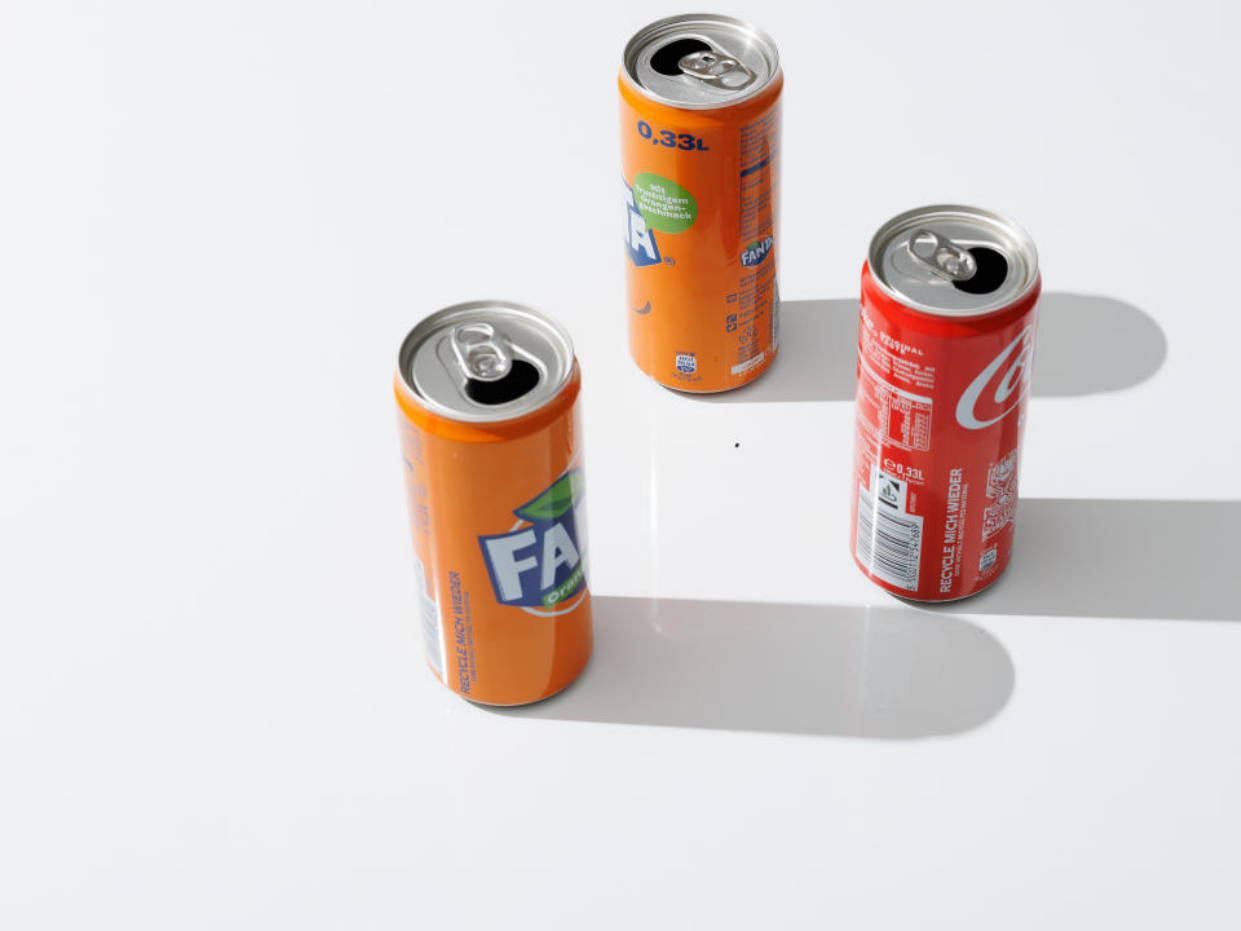} &
    \includegraphics[height=\priorsImgH,keepaspectratio]{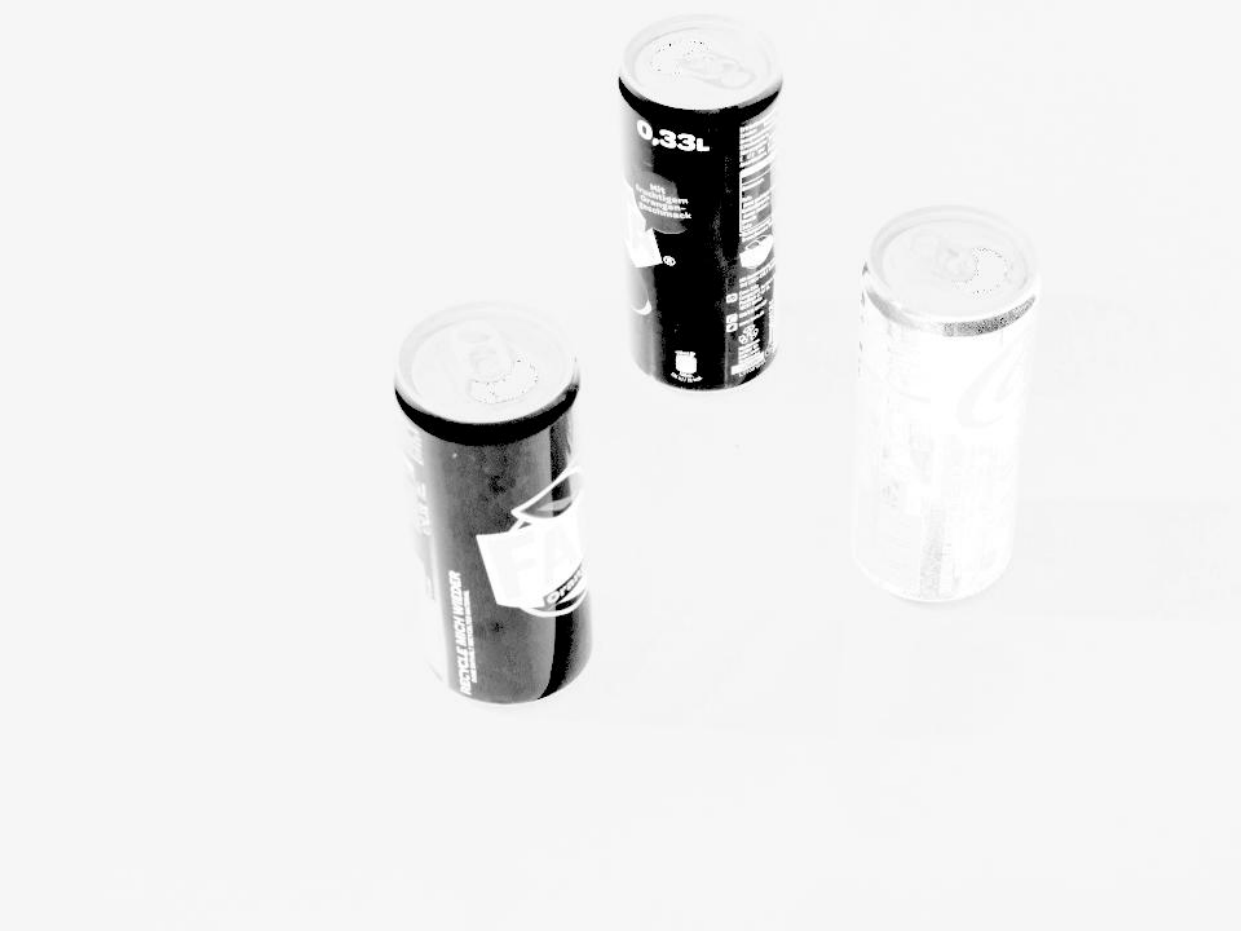} \\
    \scriptsize Input (Rain) & \scriptsize Prior (Rain) & \scriptsize Input (Shadow) & \scriptsize Prior (Shadow) \\

    \includegraphics[height=\priorsImgH,keepaspectratio]{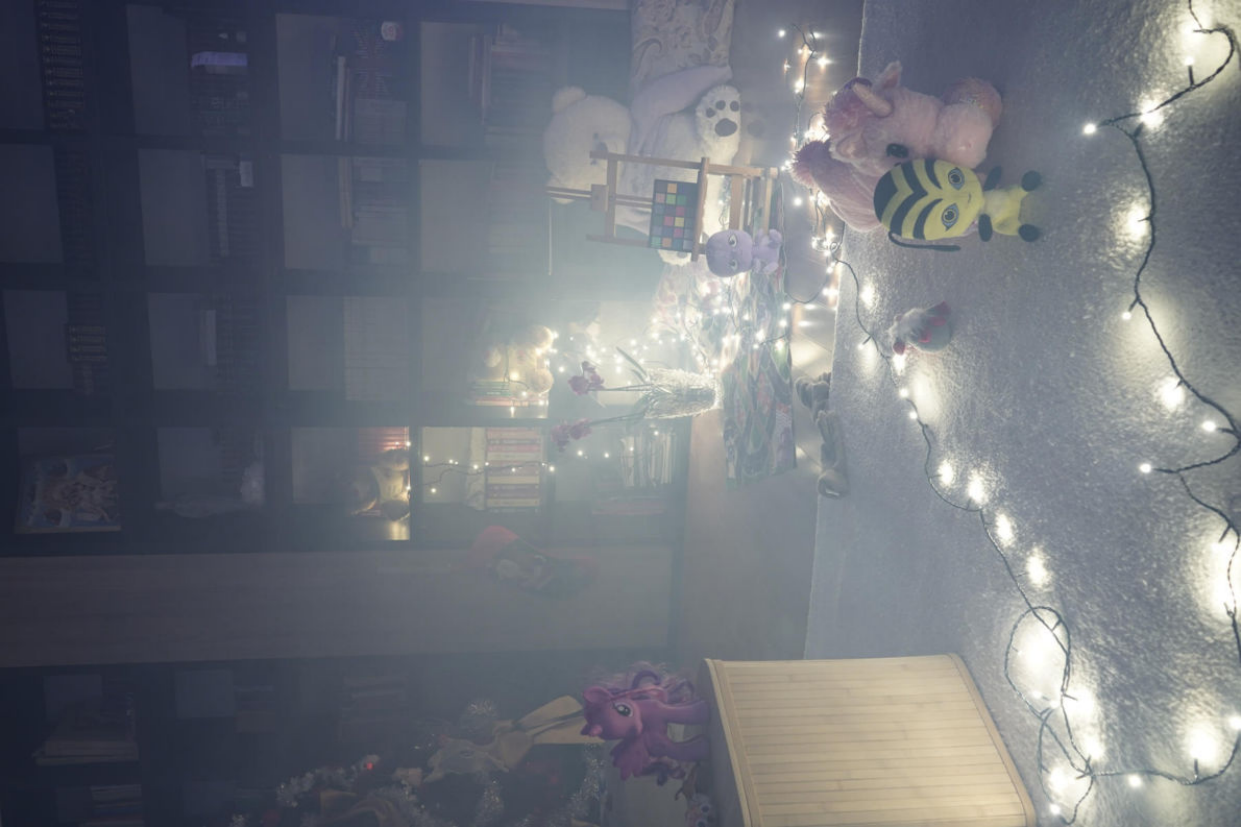} &
    \includegraphics[height=\priorsImgH,keepaspectratio]{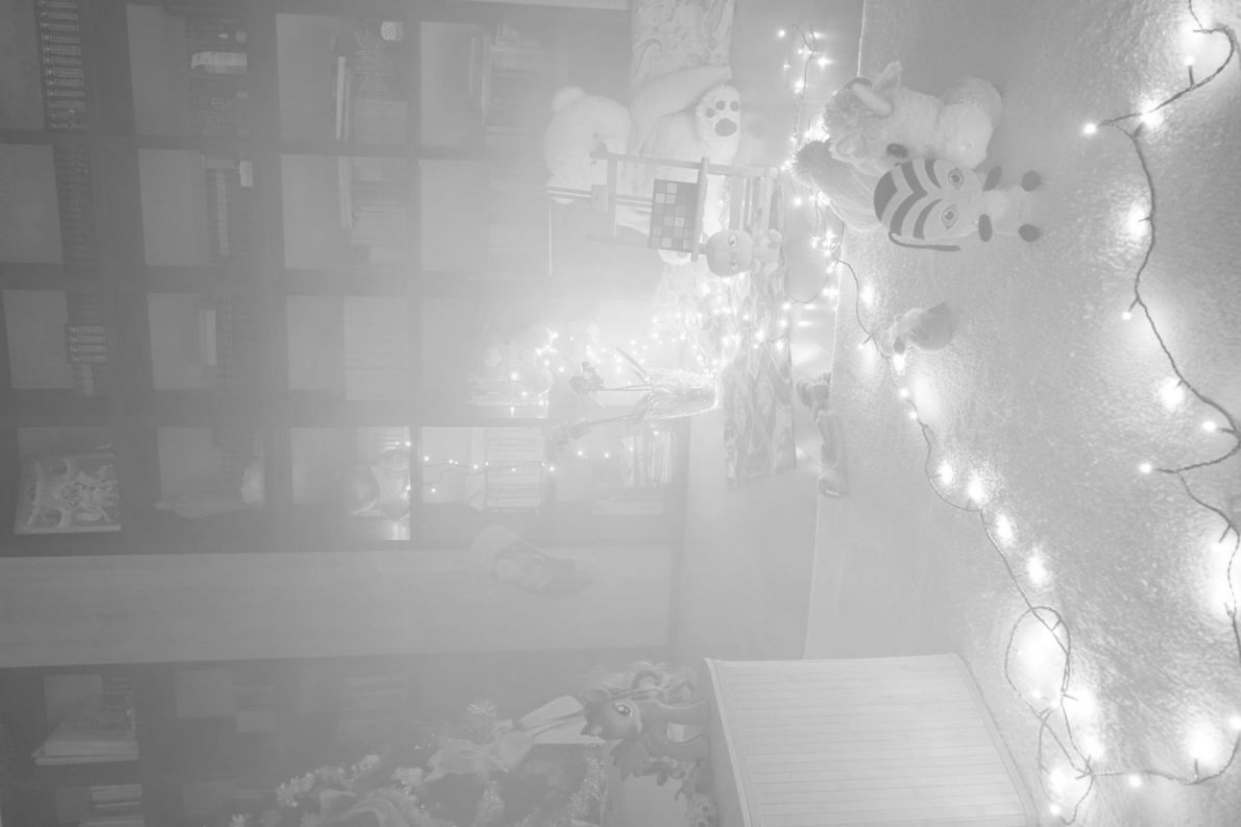} &
    \includegraphics[height=\priorsImgH,keepaspectratio]{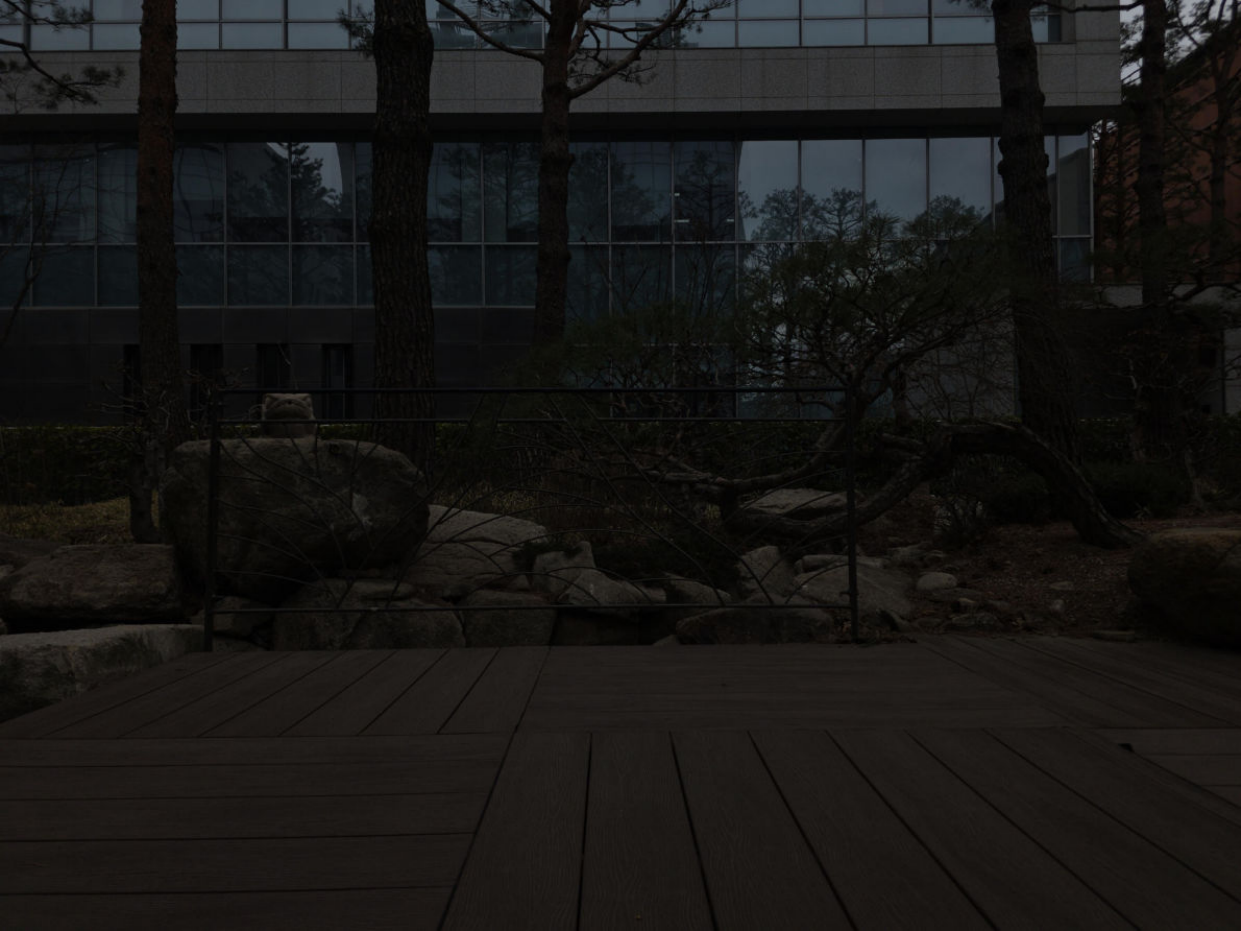} &
    \includegraphics[height=\priorsImgH,keepaspectratio]{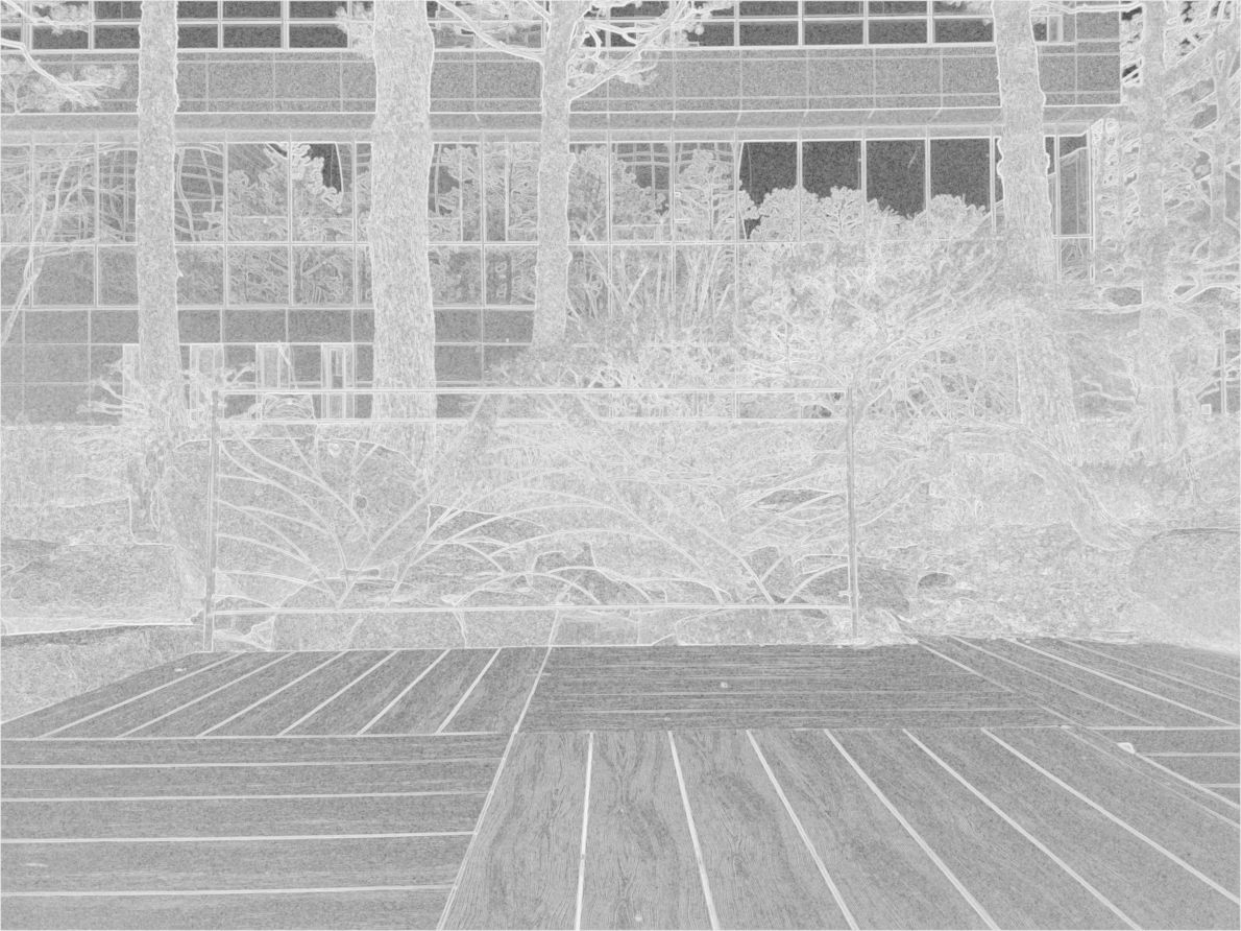} \\
    \scriptsize Input (Hazy) & \scriptsize Prior (Hazy) & \scriptsize Input (Noise LL) & \scriptsize Prior (Noise LL) \\
  \end{tabular}
  \vspace{-4mm}
  \caption{Inputs and corresponding priors for each task.}
  \label{fig:priors_examples}
\end{figure}

\subsection{Task-Specific Physical Grounding Module (TS-PGM)}
\label{subsec:TS-PGM}

The TS-PGM module extracts task-dependent physical priors as shown in Fig. \ref{fig:priors_examples}, guiding the restoration process using either degradation resistant components (as in shadow prior) or degradation sensitive components (as in rain prior) by modulating feature interactions through PC-MSA. Unlike general-purpose feature extraction, TS-PGM explicitly encodes degradation-specific physics to condition both reflection and illumination branches. Across all tasks, the physical prior is computed by fusing gradient and intensity-based representations:
\begin{equation}
    \mathbf{P}_m = \text{Fusion}(\text{Conv}_{\text{grad}}(I), \text{Conv}_{\text{int/struct}}(\text{Prior})),
\end{equation}
where $\text{Conv}_{\text{grad}}(I)$ captures image gradients to preserve high-frequency details, and the intensity or structure prior emphasizes task-specific degradation characteristics.

\textbf{Residual Rain Intensity Mask for Deraining.} Rain degradation manifests as localized artifacts with high spatial variance. We compute the residual rain mask by learning pixel-wise differences between rain-corrupted and rain-free regions. A shallow UNet $\Phi_{\text{rain}}$ trained on paired drops and blur images produces the rain intensity mask:
\begin{equation}
    \mathbf{M}_{\text{rain}} = \Phi_{\text{rain}}(I),
\end{equation}
where $\mathbf{M}_{\text{rain}}$ emphasizes regions with rain artifacts. The mask is generated offline by computing the boosted grayscale difference:
\begin{equation}
    \mathbf{M}_{\text{rain}}^{\text{gt}} = \alpha \cdot \text{Gray}(|I_{\text{drop}} - I_{\text{blur}}|),
\end{equation}
with boost factor $\alpha$ normalizing intensity variations.

\textbf{Structure-Aware Illumination Prior for Low-Light Enhancement.} Low-light degradation couples high noise with insufficient illumination, requiring joint recovery of structural clarity and brightness. We extract a structure-aware illumination prior inspired by \cite{dong2025towards} via color-invariant convolutions that remain robust to illumination changes. Specifically, we compute the Weberized invariant $W_{\text{inv}}$ through Gaussian color model transformations:
\begin{equation}
    \begin{aligned}
    \relax [E, E_l, E_{ll}] &= \mathbf{G}_{\text{cm}} \cdot I, \\
    W_{\text{inv}} &= \sum_{k \in \{0,1,2\}} \left(\frac{\nabla^{(k)} E_l}{\nabla^{(k)} E + \epsilon}\right)^2,
    \end{aligned}
\end{equation}
where $\mathbf{G}_{\text{cm}}$ is the Gaussian color model, $\nabla^{(k)}$ denotes $k$-th order derivatives via Gaussian filtering at learnable scales, and $\epsilon$ prevents numerical instability. This color-invariant representation captures local structure independent of absolute intensity, providing stable guidance under extreme underexposure.

\textbf{Dark Channel Prior for Nighttime Dehazing.} Haze and atmospheric scattering in nighttime scenes compound illumination deficiency, requiring joint dehazing and contrast enhancement. The dark channel prior, a fundamental haze prior from classical image restoration, captures pixels where at least one color channel exhibits low intensity due to scattering. We compute the dark channel as:
\begin{equation}
    \mathbf{D}_{c}(I) = \min_{c \in \{R,G,B\}} I_c,
\end{equation}
where $\mathbf{D}_{c}$ represents the minimum intensity across color channels at each pixel. This prior effectively identifies haze-dominated regions with low contrast.

\textbf{Log-Chromaticity Shadow-Resistant Prior for Shadow Removal.} Shadow regions exhibit localized illumination variations that distort reflectance estimates. We adopt a log-chromaticity shadow-resistant representation that isolates shadow cues independent of absolute brightness. Given an image $I$ normalized to $[0,1]$, we compute:
\begin{equation}
    \begin{aligned}
    \mu_g &= (I_R \cdot I_G \cdot I_B)^{1/3}, \\
    \rho_R &= \log(I_R / \mu_g), \quad \rho_B = \log(I_B / \mu_g), \\
    \mathbf{S}_{inv} &= \rho_R \cos(\theta) + \rho_B \sin(\theta), \\
    \mathbf{S}_{inv}^{\text{norm}} &= \text{clip}\left(\frac{\mathbf{S}_{inv} - p_2}{p_{98} - p_2}, 0, 1\right),
    \end{aligned}
\end{equation}
where $\mu_g$ is the geometric mean across color channels, $\rho_R, \rho_B$ are log-chromaticity features, $\theta$ is a learnable projection angle, and $p_{\alpha}$ denotes the $\alpha$-th percentile for robust normalization. This shadow-resistant invariant enables the illumination branch to recover shadow-free lighting cues while guiding the reflection branch to reconstruct shadow-occluded details.

\textbf{Motivation for Degradation-Specific Priors.} By extracting task-dependent structural representations rather than generic convolutional features, TS-PGM serves as a lightweight task-selector. The overarching motivation for utilizing distinct priors (rain mask, structure-aware illumination, dark channel, and log-chromaticity) stems from the observation that different degradations violate Retinex assumptions in fundamentally different ways. For instance, rain constitutes high-frequency localized occlusions violating reflectance continuity, whereas haze globally distorts illumination transmission. By injecting these carefully chosen physical mappings dynamically into PC-MSA, our model gracefully generalizes its single dual-branch backbone across drastically different conditions without requiring any task-specific rewriting of the core components.
 

\subsection{RetinexDualV2 Framework}
\label{subsec:Retiv2}

Inspired by RetinexDual \cite{kishawy2025retinexdualretinexbaseddualnature}, RetinexDualV2 follows Retinex theory, which decomposes a distorted image $I \in \mathbb{R}^{H \times W \times 3}$ as:
\begin{equation}
I = (R_{eff} + \hat{R}) \odot (L_{eff} + \hat{L})
\end{equation}
where $R_{eff}$, $L_{eff}$ are the effective reflection and illumination components, and $\hat{R}$, $\hat{L}$ represent distortion artifacts. Our Retinex decomposer $\psi_d$ extracts these components, and the restoration is formulated as:
\begin{equation}
\begin{aligned}
R_{eff}, L_{eff} =& \psi_d(I), \\
\hat{R} = -\mathcal{S}(R_{eff},P_m),& \quad \hat{L} = -\mathcal{F}(L_{eff},P_m),\\
I_{out} = (R_{eff} - \hat{R}) \odot& (L_{eff} - \hat{L})
\end{aligned}
\end{equation}
where $\mathcal{S}$ (PG-SAMBA) and $\mathcal{F}$ (PG-FCB) are complementary sub-networks designed to correct reflection and illumination distortions respectively. 

\textbf{Physical-Conditioned Multi-head Self-Attention (PC-MSA)}
The core component enabling physical grounding across both branches is the Physical-conditioned Multi-head Self-Attention (PC-MSA). PC-MSA extends standard multi-head self-attention by modulating feature interactions with task-specific physical guidance extracted from TS-PGM. 

Given input features $\mathbf{F}_{\text{in}} \in \mathbb{R}^{H \times W \times C}$, we first reshape to $\mathbf{X} \in \mathbb{R}^{HW \times C}$. The Query, Key, and Value representations are obtained through linear projections:
\begin{equation}
    \mathbf{Q} = \mathbf{X} \mathbf{W}_{Q}^{\top}, \quad
    \mathbf{K} = \mathbf{X} \mathbf{W}_{K}^{\top}, \quad
    \mathbf{V} = \mathbf{X} \mathbf{W}_{V}^{\top},
\end{equation}
where $\mathbf{W}_{Q}, \mathbf{W}_{K}, \mathbf{W}_{V} \in \mathbb{R}^{C \times C}$ are learnable projection matrices. Crucially, PC-MSA incorporates physical guidance $\mathbf{P}_m \in \mathbb{R}^{HW \times C}$ extracted from TS-PGM to modulate the value vectors, ensuring feature refinement respects physical constraints of the degradation task:
\begin{equation}
    \mathbf{V}' = \mathbf{V} \odot \mathbf{P}_m,
\end{equation}
where $\odot$ denotes element-wise multiplication. The physical-conditioned attention is then computed as:
\begin{equation}
    \text{Attention}(\mathbf{Q}, \mathbf{K}, \mathbf{V}') = \text{softmax}\left(\frac{\mathbf{K}^{\top} \mathbf{Q}}{\sigma}\right) \mathbf{V}',
\end{equation}
where $\sigma$ is a learnable per-head rescale parameter. Finally, positional embeddings via grouped convolutions capture spatial context, grounding self-attention in both physical and spatial domains.

\textbf{Physical-Grounded Scale Adaptive Mamba Block (PG-SAMB)}
PG-SAMB is a compact, coarse-to-fine block that injects TS-PGM priors into Mamba-style feature aggregation via PC-MSA. Instead of reworking the core scanning mechanism, we condition scale-wise attention and local convolutions on the physical prior so that the block emphasizes degradation-relevant regions across three scales and fuses these conditioned corrections into the residual path. A lightweight group state-space component supplies non-sequential semantic context to improve long-range consistency.

\textbf{Physical-Grounded Fourier Correction Block (PG-FCB)}
PG-FCB is a small, frequency-domain illumination corrector that applies PC-MSA conditioning before compact Fourier-domain processing. Guided by the physical prior, it modulates amplitude/phase via lightweight 1×1 convolutions in the spectral domain and returns an adaptive residual correction in image space. This design captures global contrast and color biases efficiently, keeping the illumination branch parameter-light (0.2M).

\subsection{Multi-level Training Objectives}
\label{subsec:TrainObj}

Following a deep supervision strategy, we employ a multi-level training with outputs at three resolutions: $\hat{I}_1$, $\hat{I}_2$, and $\hat{I}_3$ as shown in Fig. \ref{fig:RD_overview}. The total loss combines Charbonnier loss $\mathcal{L}_{cb}$ \cite{DBLP:journals/corr/abs-1710-01992}, SSIM loss $\mathcal{L}_{ssim}$ \cite{1284395}, FFT loss $\mathcal{L}_{fft}$, and perceptual loss $\mathcal{L}_p$ \cite{8578166} using AlexNet features \cite{NIPS2012_c399862d}:
\begin{equation}
    \begin{aligned}
    \mathcal{L}_{total} &= \sum_{i=1}^{3} w_i \Big( \mathcal{L}_{cb}(I_i, \hat{I}_i) + \lambda_{ssim}(1 - S_i) \\
    &\quad + \lambda_{fft}\mathcal{L}_{fft}(I_i, \hat{I}_i) + \lambda_p \mathcal{L}_p(I_i, \hat{I}_i) \Big),
    \end{aligned}
\end{equation}
where $w = [0.25, 0.5, 1.0]$ for $i = 1, 2, 3$ scales, $S_i = \text{SSIM}(I_i, \hat{I}_i)$, and $\lambda_{ssim} = 0.5$, $\lambda_{fft} = 0.0001$, $\lambda_p = 0.5$ are the loss weights. This multi-scale formulation progressively refines predictions while maintaining retinex decomposition constraints across all resolution levels.
\section{Experiments}
\label{sec:experiments}

\begin{table}[t]
\centering
\caption{4K-Rain13k Quantitative Comparison}
\label{tab:RainQuantitative}
\vspace{-2mm}
{\fontsize{8}{10}\selectfont
\begin{tabular}{l|c|c|c|c}
\hline
\textbf{Methods} & \textbf{Venue} & \textbf{PSNR$\uparrow$} & \textbf{SSIM$\uparrow$}  & \textbf{Param.$\downarrow$} \\
\hline
JORDER-E    & TPAMI'20& 30.46 & 0.912  & 4.21M  \\
RCDNet      & CVPR'20 & 30.83 & 0.921  & 3.17M  \\
SPDNet      & ICCV'21 & 31.81 & 0.922  & 3.04M  \\
IDT         & TPAMI'22& 32.91 & 0.948  & 16.41M \\
Restormer   & CVPR’22 & 33.02 & 0.933  & 26.12M \\
DRSformer   & CVPR'23 & 32.94 & 0.933  & 33.65M \\
UDR-S2Former& CVPR'23 & 33.36 & 0.946  & 8.53M  \\
UDR-Mixer   & arXiv'24& 34.28 & 0.951  & 8.53M  \\
ERR         & CVPR'25 & 34.48 & 0.952  & 1.131M \\
RetinexDual & Arxiv'25 & \underline{34.50} & \underline{0.967}  & 4.726M \\
\hline
\textbf{Ours} & - & \textbf{34.56} & \textbf{0.973} & 4.848M \\
\hline
\end{tabular}
}
\end{table}

\begin{table}[t]
\centering
\caption{UHD-LL Quantitative Comparison}
\label{tab:LLQuantitative}
\vspace{-3mm}
{\fontsize{8}{10}\selectfont
\begin{tabular}{l|c|c|c|c}
\hline
\textbf{Methods} & \textbf{Venue} & \textbf{PSNR$\uparrow$} & \textbf{SSIM$\uparrow$}  & \textbf{Param.$\downarrow$} \\
\hline
IFT         & ICCV'21 & 21.96 & 0.870  & 11.56M \\
SNR-Aware   & CVPR'22 & 22.72 & 0.877  & 40.08M \\
Uformer     & CVPR’22 & 19.28 & 0.849  & 20.62M \\
Restormer   & CVPR’22 & 22.25 & 0.871  & 26.11M \\
DiffLL      & TOG'23  & 21.36 & 0.872  & 17.29M \\
LLFormer    & AAAI’23 & 22.79 & 0.853  & 13.15M \\
UHDFour     & ICLR’23 & 26.23 & 0.900  & 17.54M \\
Wave-Mamba  & MM'24   & 27.35 & 0.913  & 1.258M \\
UHDFormer   & AAAI’24 & 27.11 & 0.927  & 0.339M \\
ERR         & CVPR'25 & 27.57 & \underline{0.932}  & 1.131M \\
RetinexDual & Arxiv'25 & \underline{28.86} & 0.920  & 4.726M \\
\hline
\textbf{Ours} & - & \textbf{28.91} & \textbf{0.935} & 4.848M \\
\hline
\end{tabular}
}
\end{table}

\begin{figure*}[ht]
  \centering
  \setlength{\tabcolsep}{2pt}
  \begin{tabular}{c@{\hspace{0.4mm}}c@{\hspace{0.4mm}}c@{\hspace{0.4mm}}c@{\hspace{0.4mm}}c@{\hspace{0.4mm}}c@{\hspace{0.4mm}}c}
    \textbf{Input} & \textbf{UHDFour} & \textbf{Wave-Mamba} & \textbf{ERR} & \textbf{RetinexDual} & \textbf{Ours} & \textbf{GT} \\
    \includegraphics[width=0.135\textwidth]{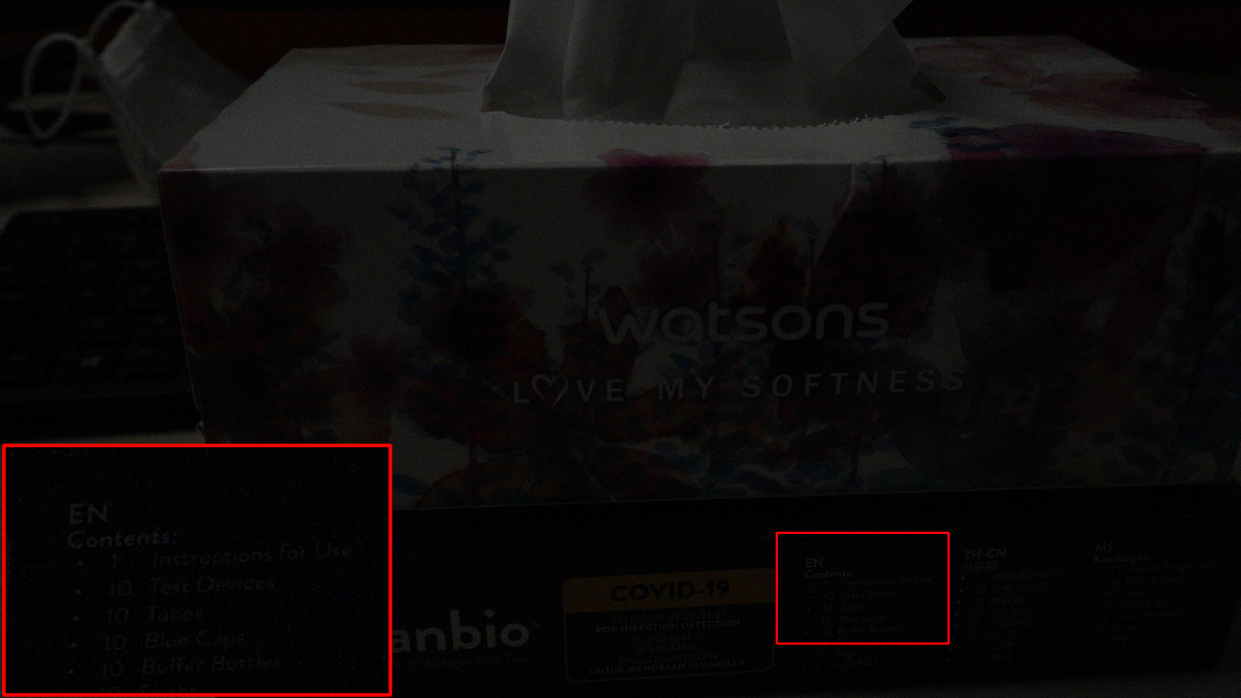} &
    \includegraphics[width=0.135\textwidth]{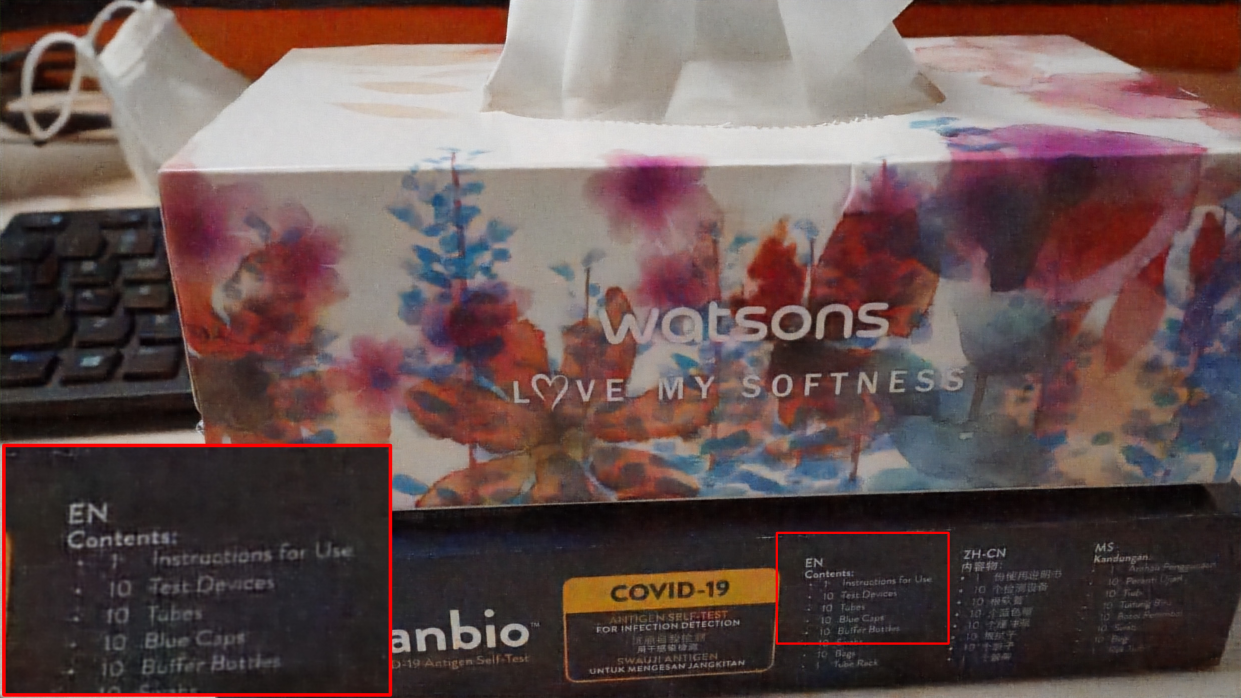} &
    \includegraphics[width=0.135\textwidth]{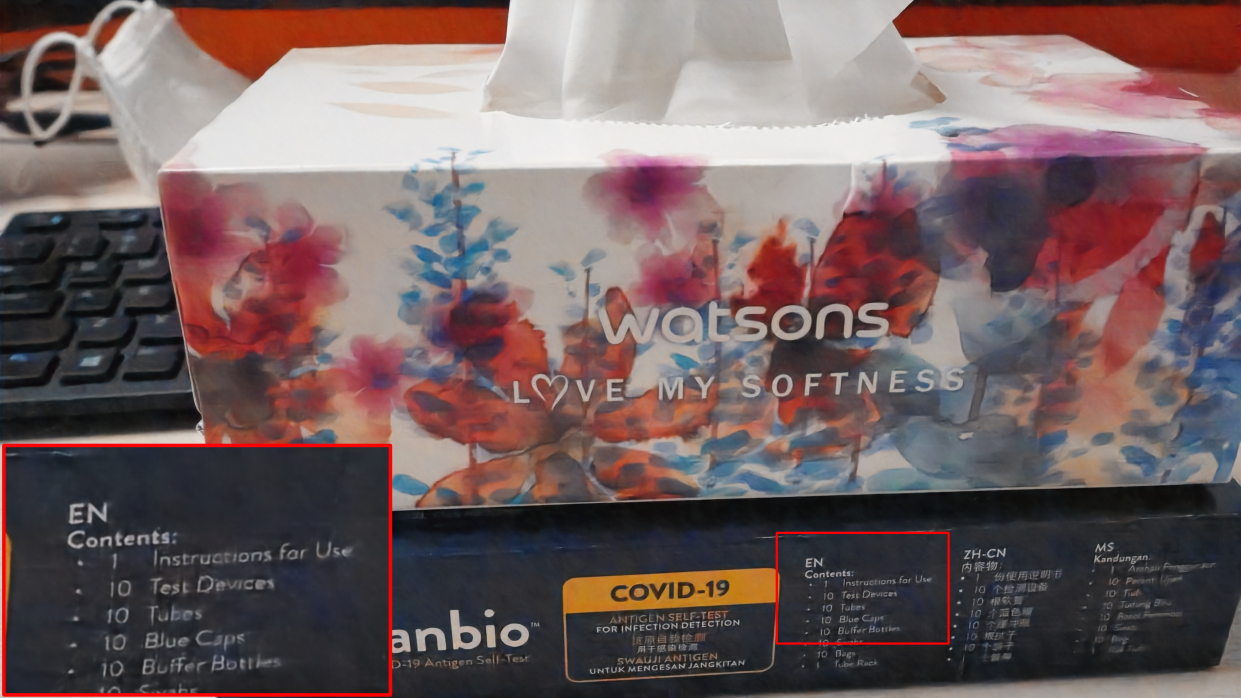} &
    \includegraphics[width=0.135\textwidth]{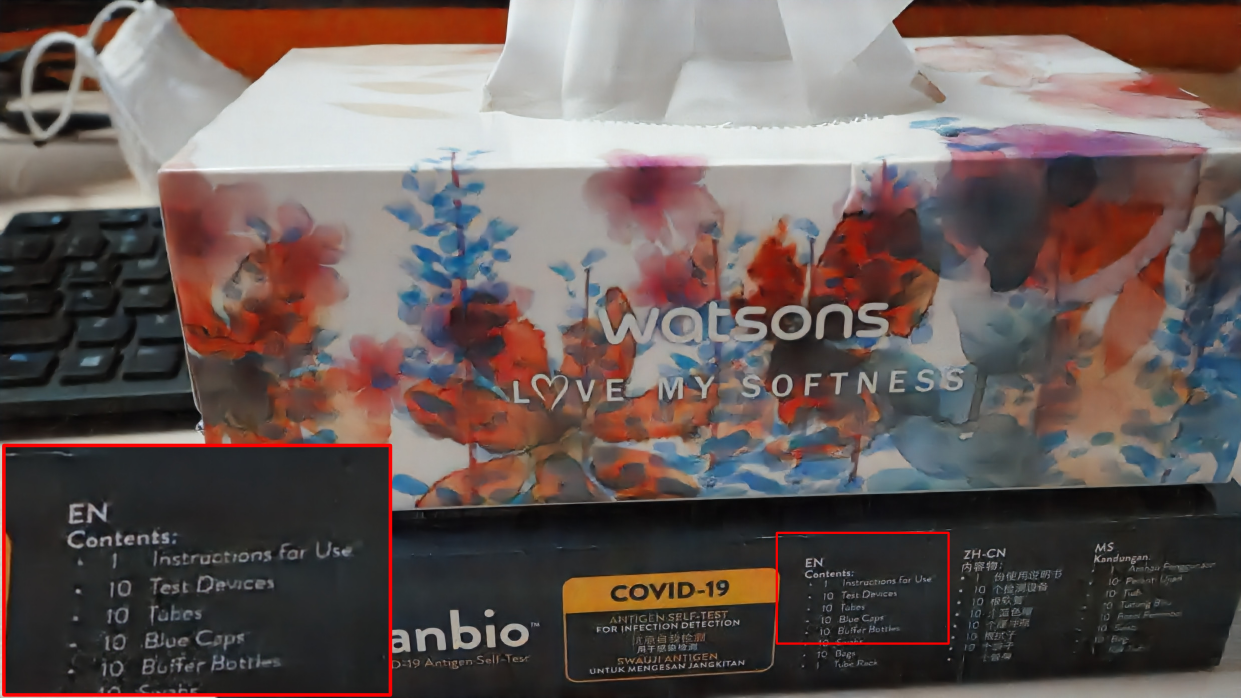} &
    \includegraphics[width=0.135\textwidth]{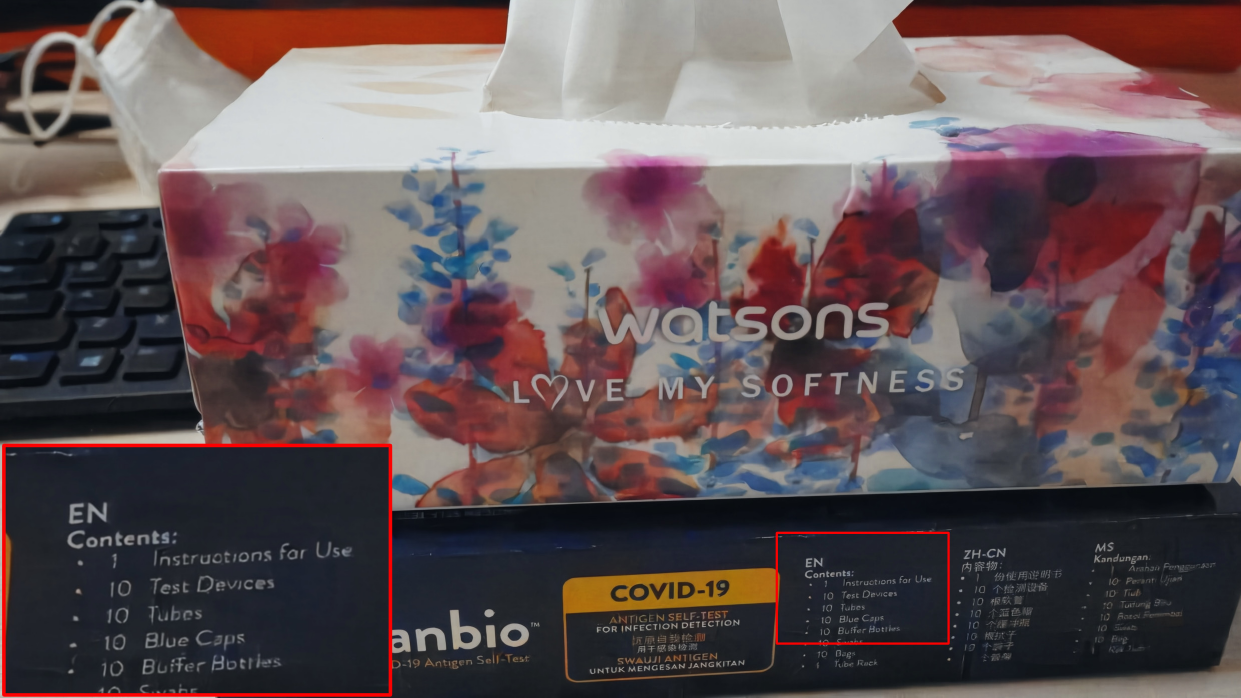} &
    \includegraphics[width=0.135\textwidth]{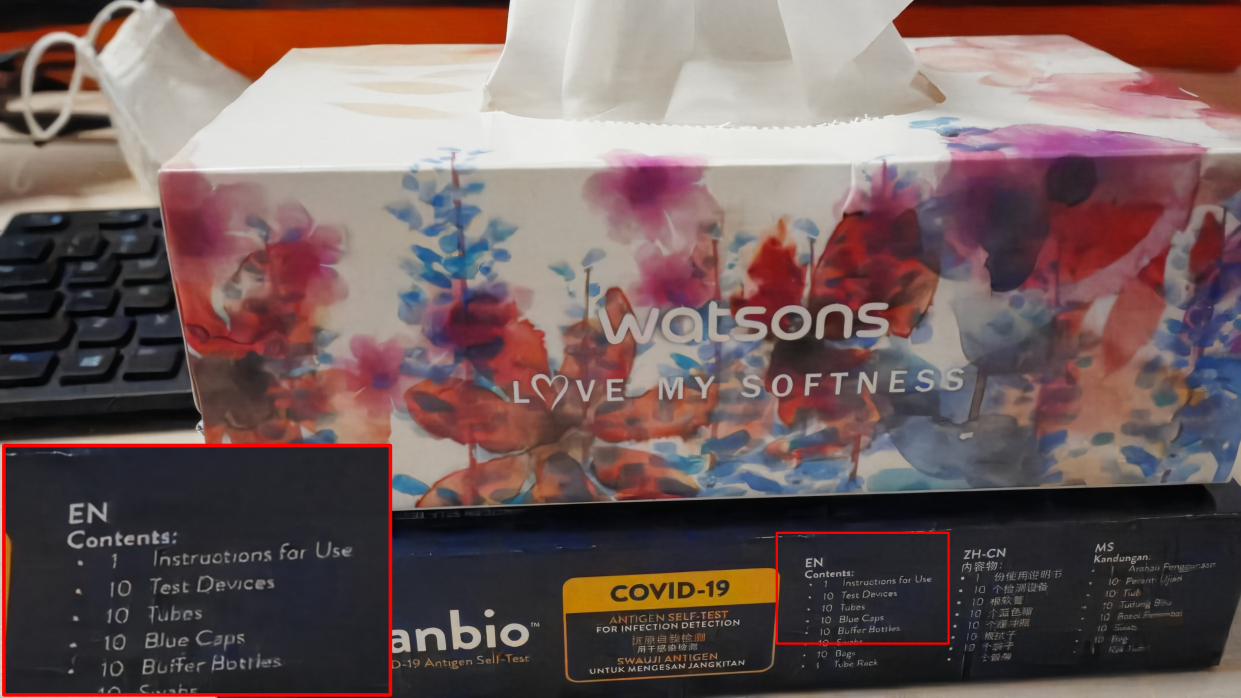} &
    \includegraphics[width=0.135\textwidth]{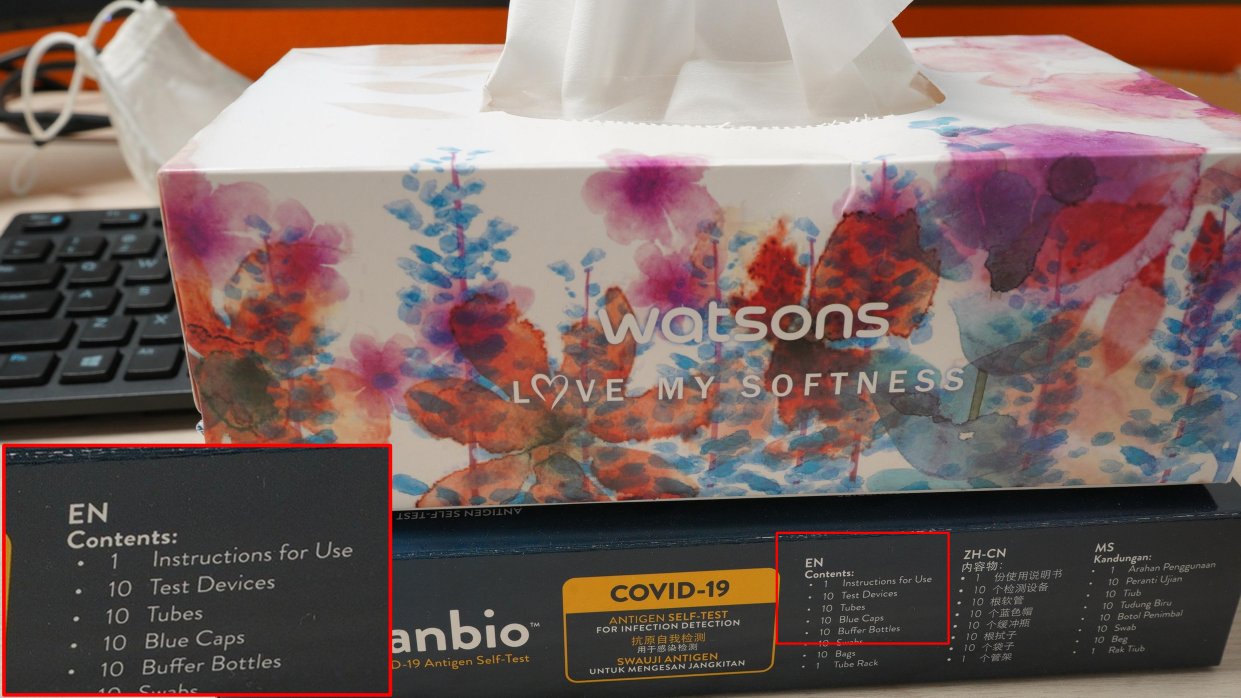} \\
    
    \includegraphics[width=0.135\textwidth]{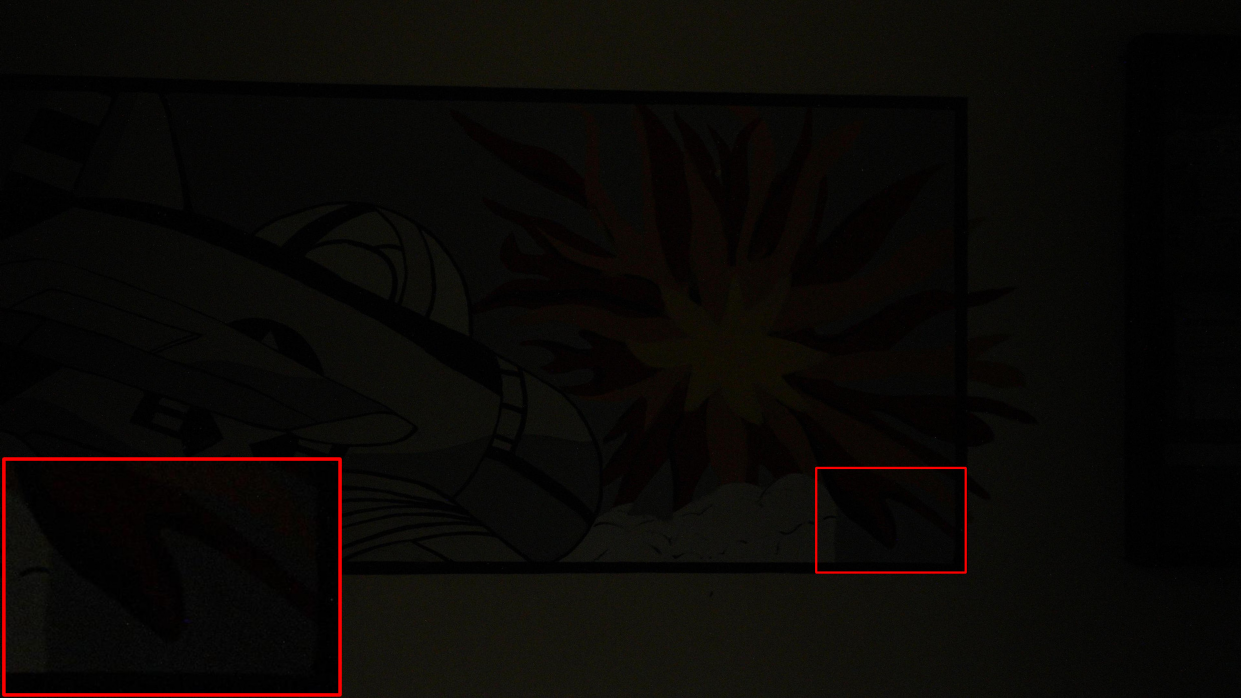} &
    \includegraphics[width=0.135\textwidth]{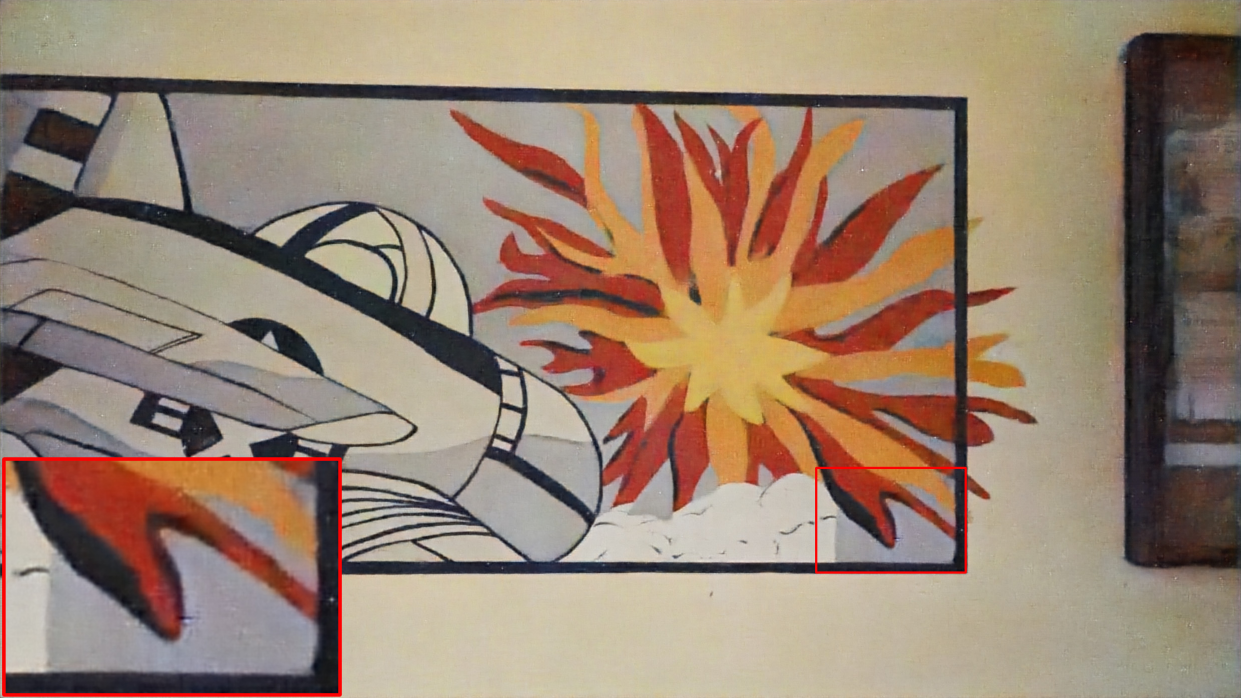} &
    \includegraphics[width=0.135\textwidth]{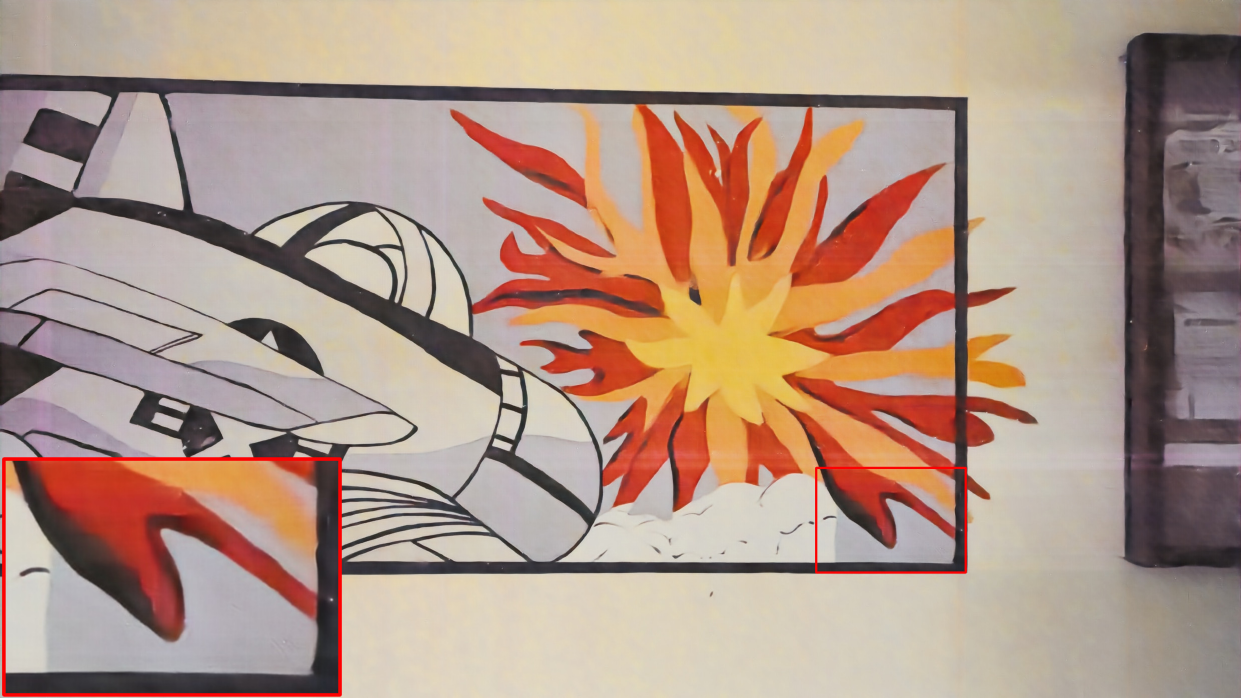} &
    \includegraphics[width=0.135\textwidth]{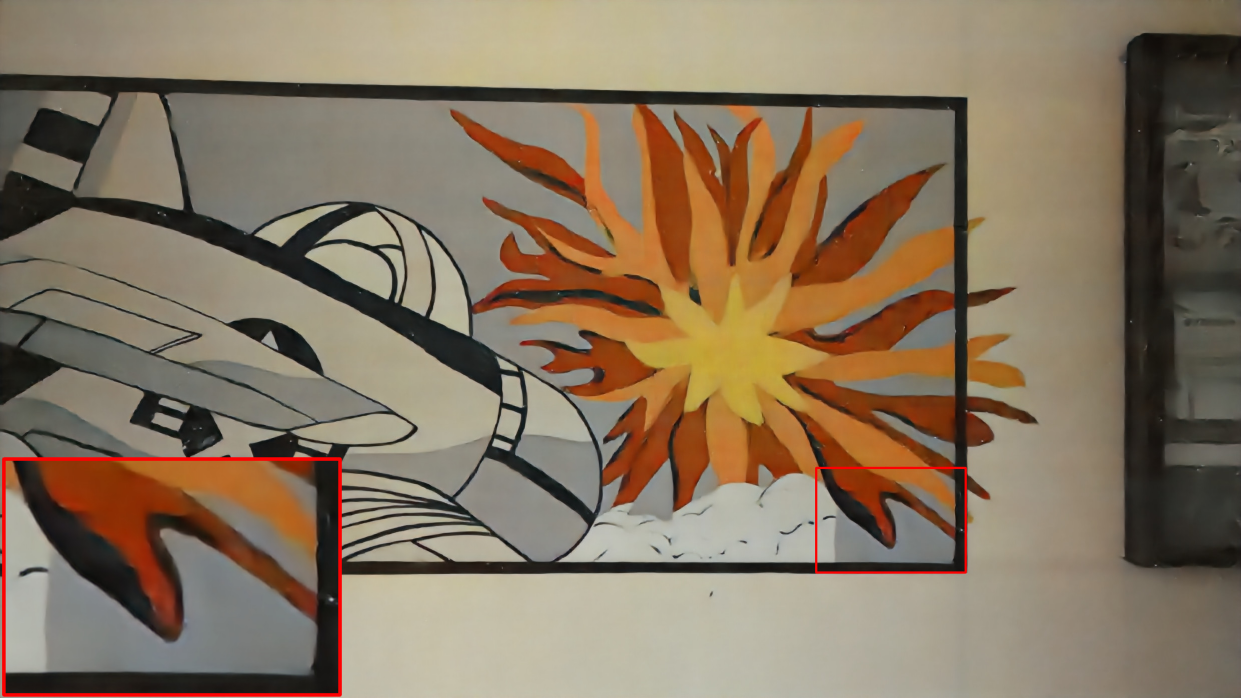} &
    \includegraphics[width=0.135\textwidth]{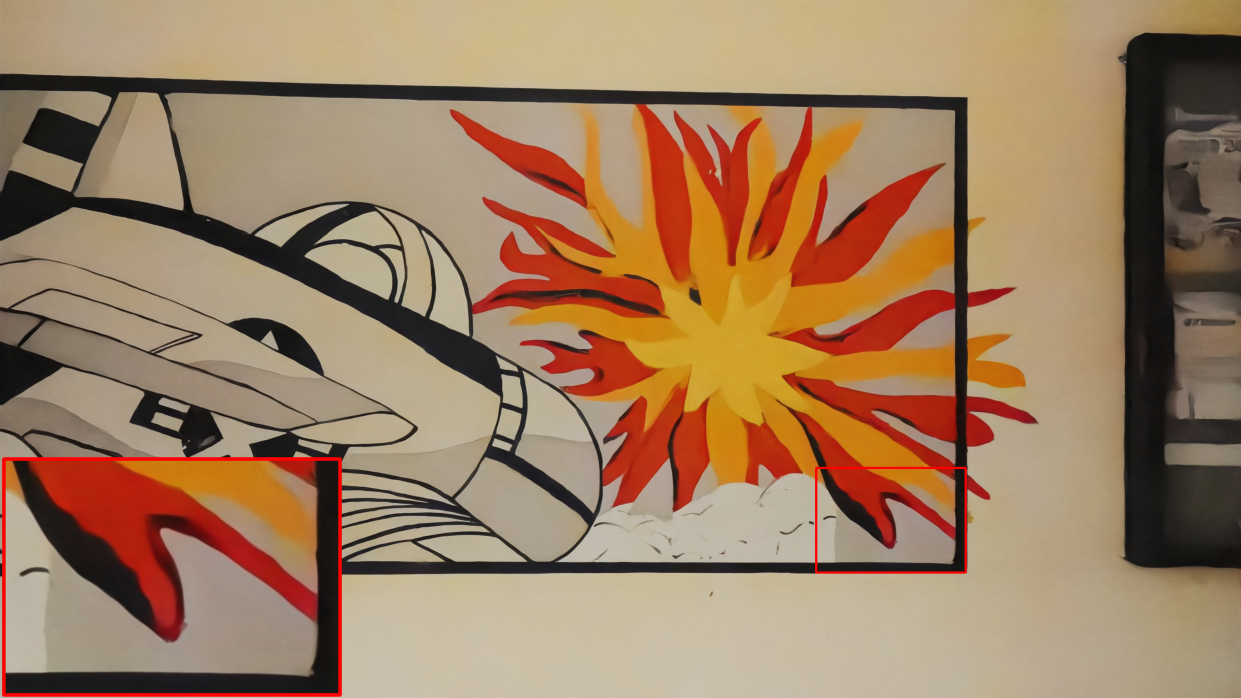} &
    \includegraphics[width=0.135\textwidth]{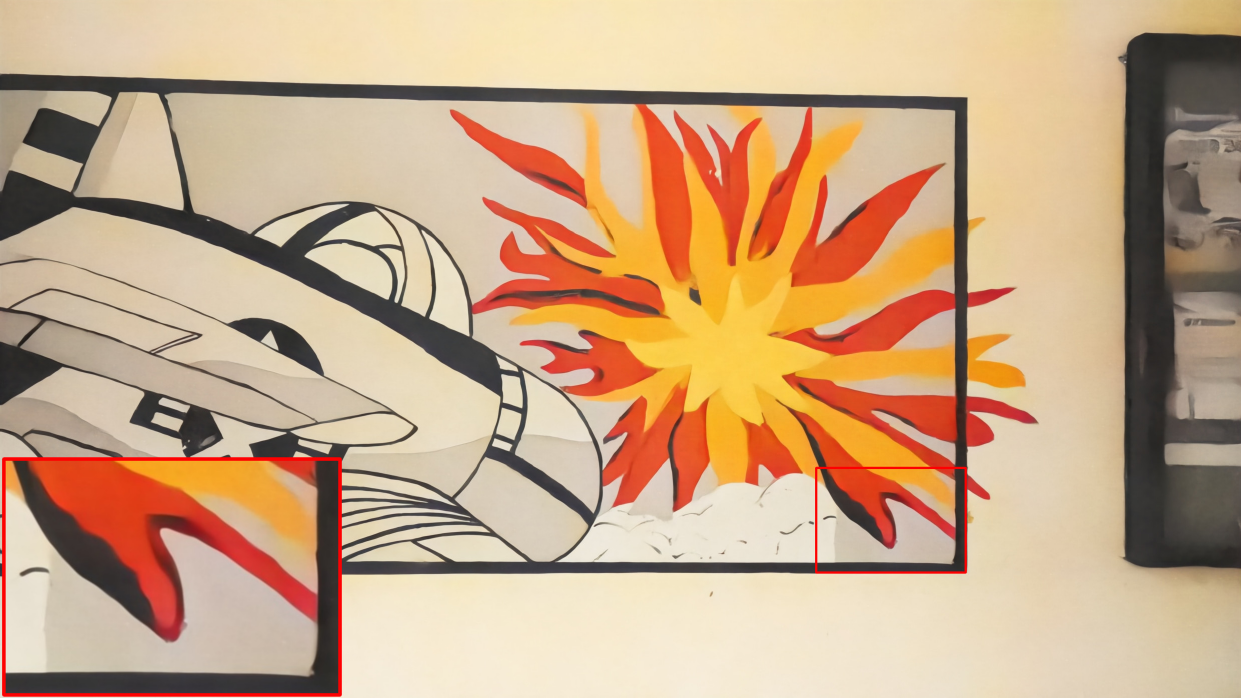} &
    \includegraphics[width=0.135\textwidth]{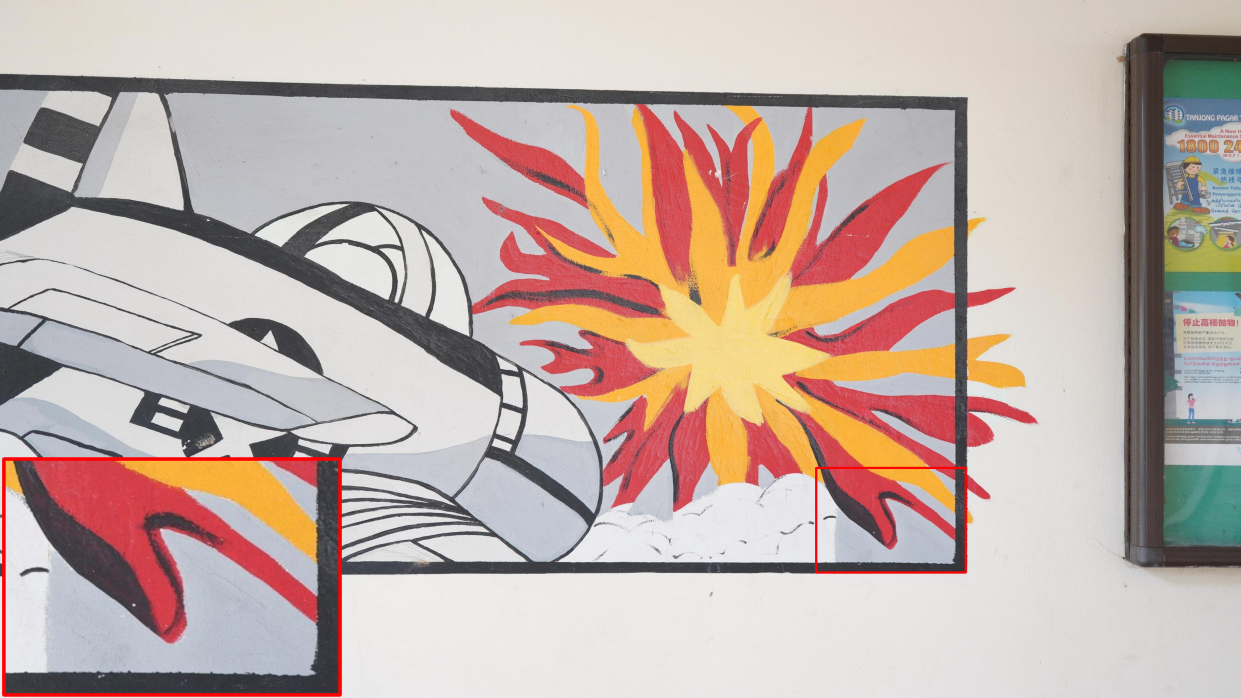} \\
  \end{tabular}

  \caption{Visual comparison between different architectures on UHD-LL \cite{Li2023ICLR}. Zoomed-in patches highlight our method's superiority in noise suppression and reliable structure preservation in comparison with literature methods.}
  \label{fig:VisualComparisonLLIE}
\end{figure*}

\begin{figure*}[ht]
  \centering
  \setlength{\tabcolsep}{2pt}
  \begin{tabular}{c@{\hspace{0.4mm}}c@{\hspace{0.4mm}}c@{\hspace{0.4mm}}c@{\hspace{0.4mm}}c@{\hspace{0.4mm}}c@{\hspace{0.4mm}}c}
    \textbf{Input} & \textbf{UDR-S2Former} & \textbf{UDR-mixer} & \textbf{ERR} & \textbf{RetinexDual} & \textbf{Ours} & \textbf{GT} \\
    \includegraphics[width=0.135\textwidth]{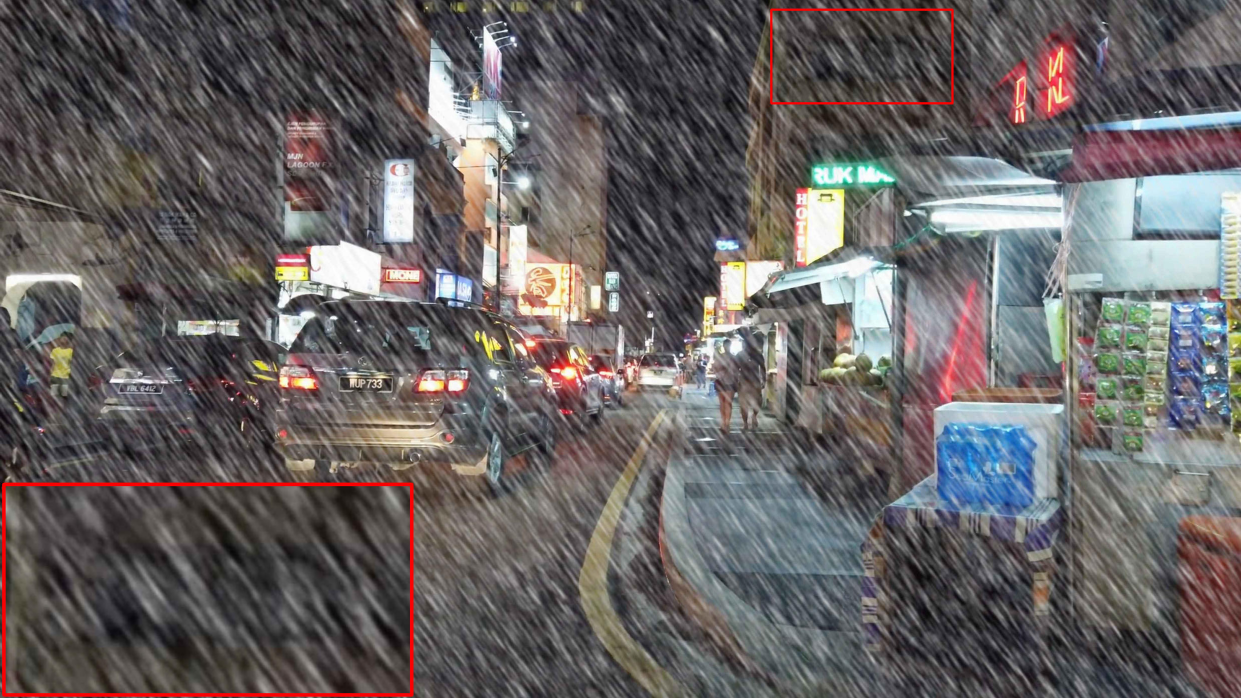} &
    \includegraphics[width=0.135\textwidth]{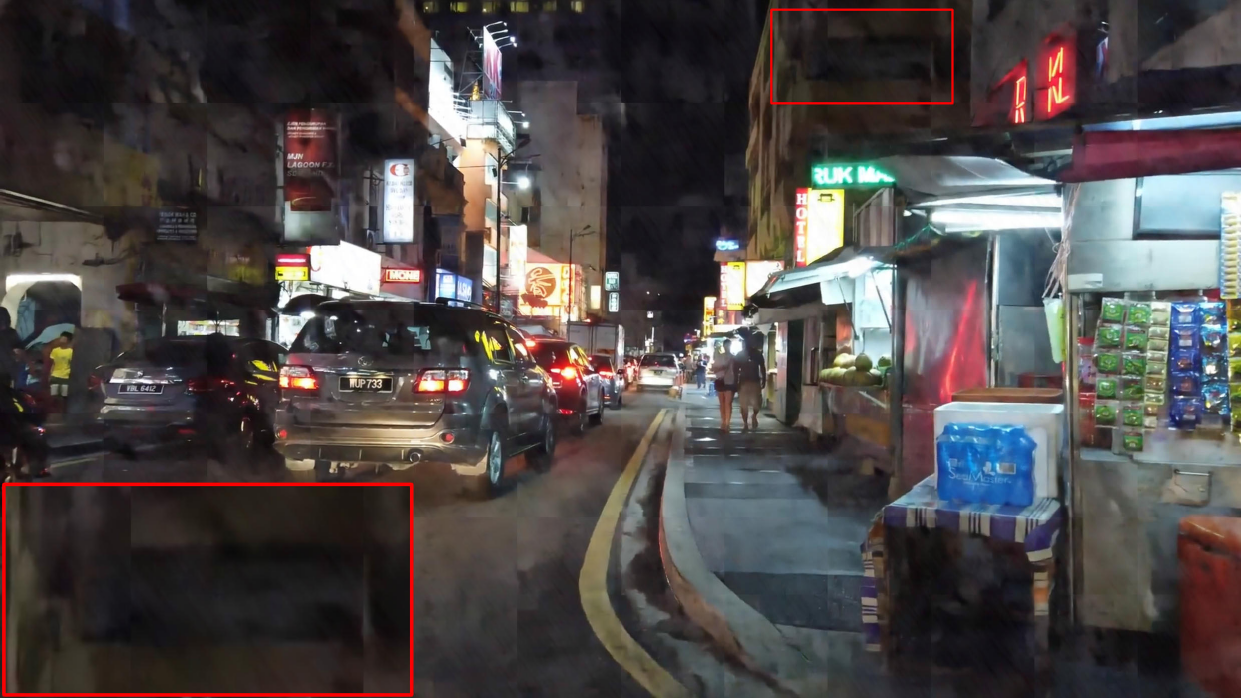} &
    \includegraphics[width=0.135\textwidth]{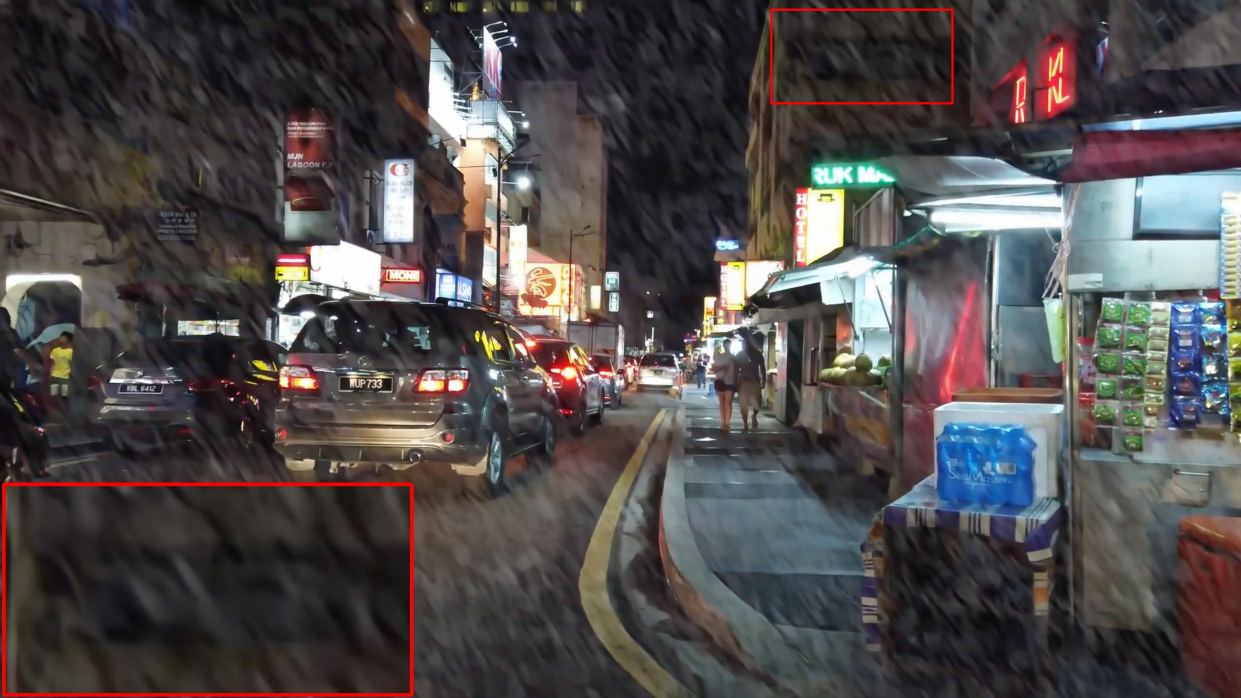} &
    \includegraphics[width=0.135\textwidth]{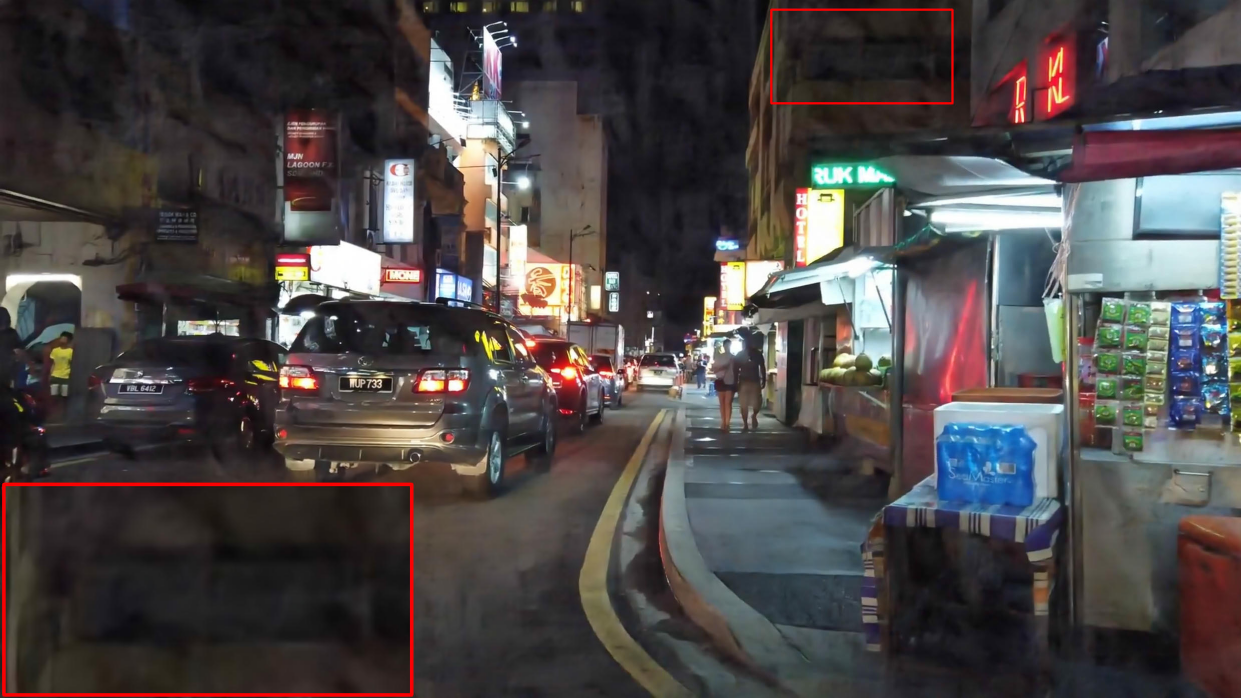} &
    \includegraphics[width=0.135\textwidth]{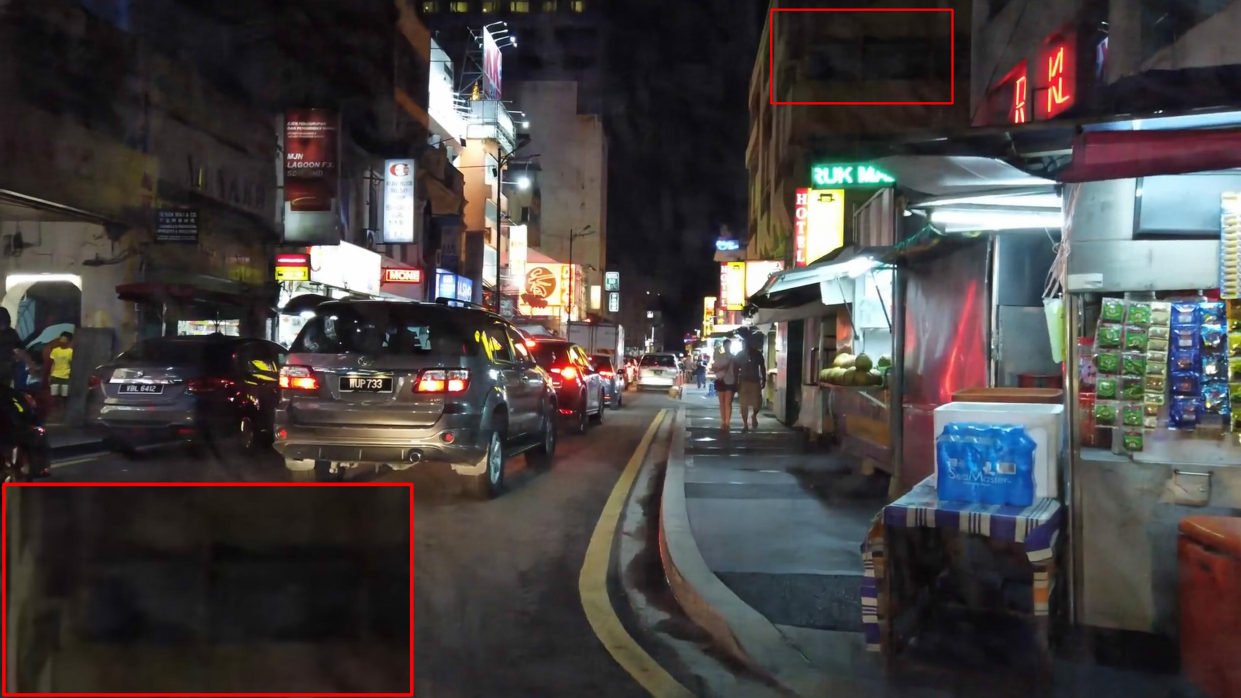} &
    \includegraphics[width=0.135\textwidth]{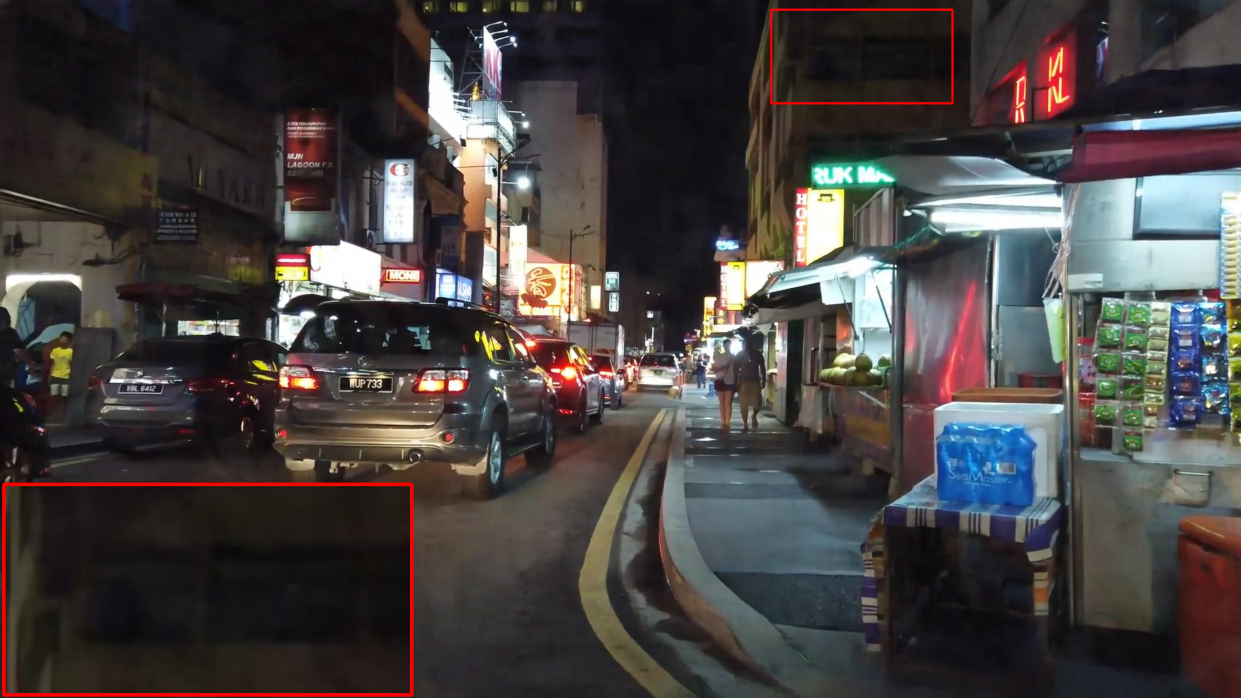} &
    \includegraphics[width=0.135\textwidth]{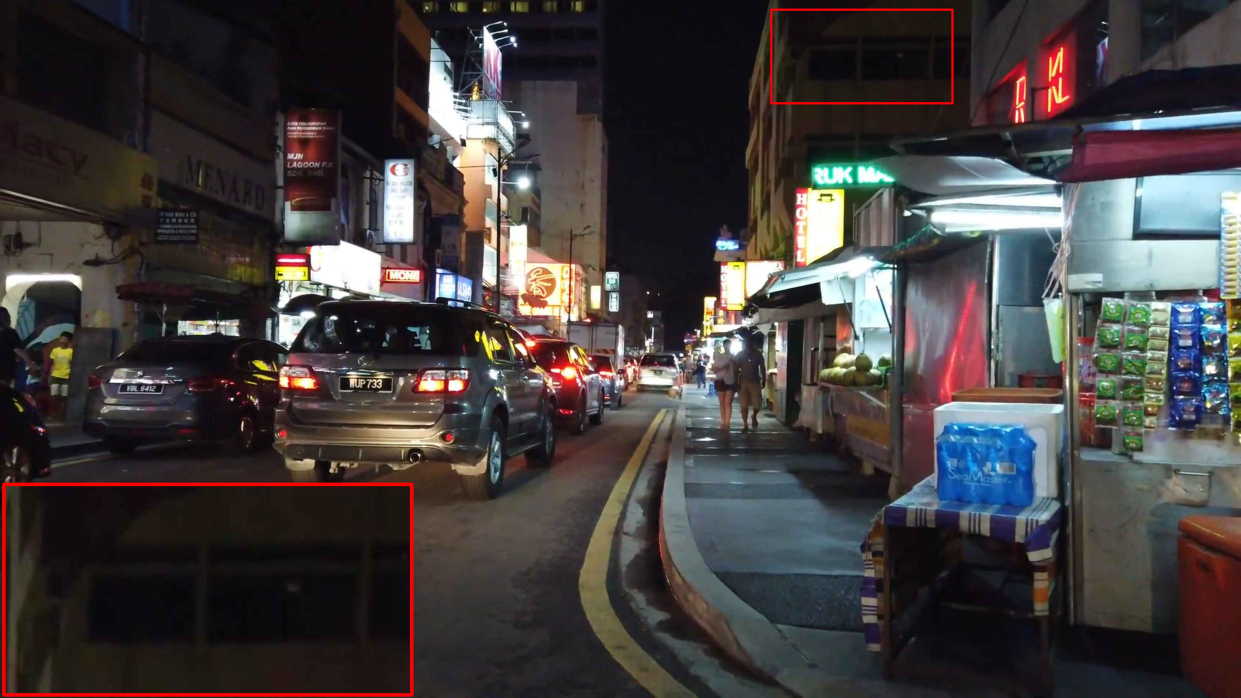} \\
    
    \includegraphics[width=0.135\textwidth]{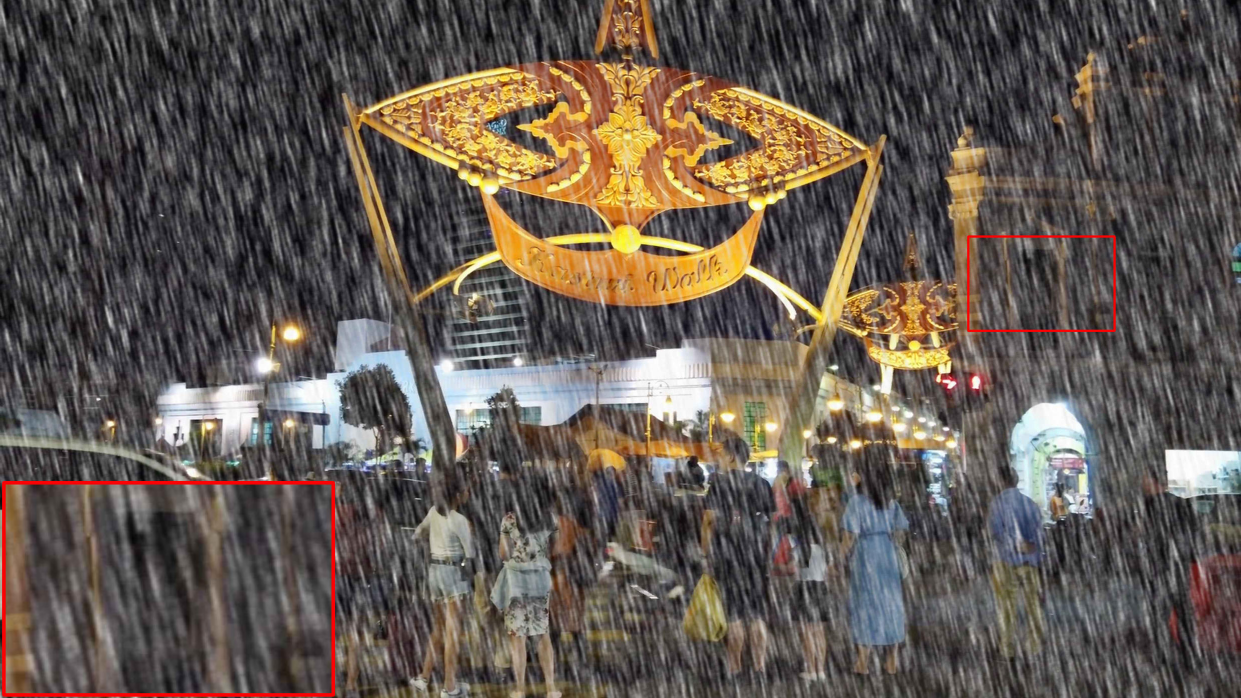} &
    \includegraphics[width=0.135\textwidth]{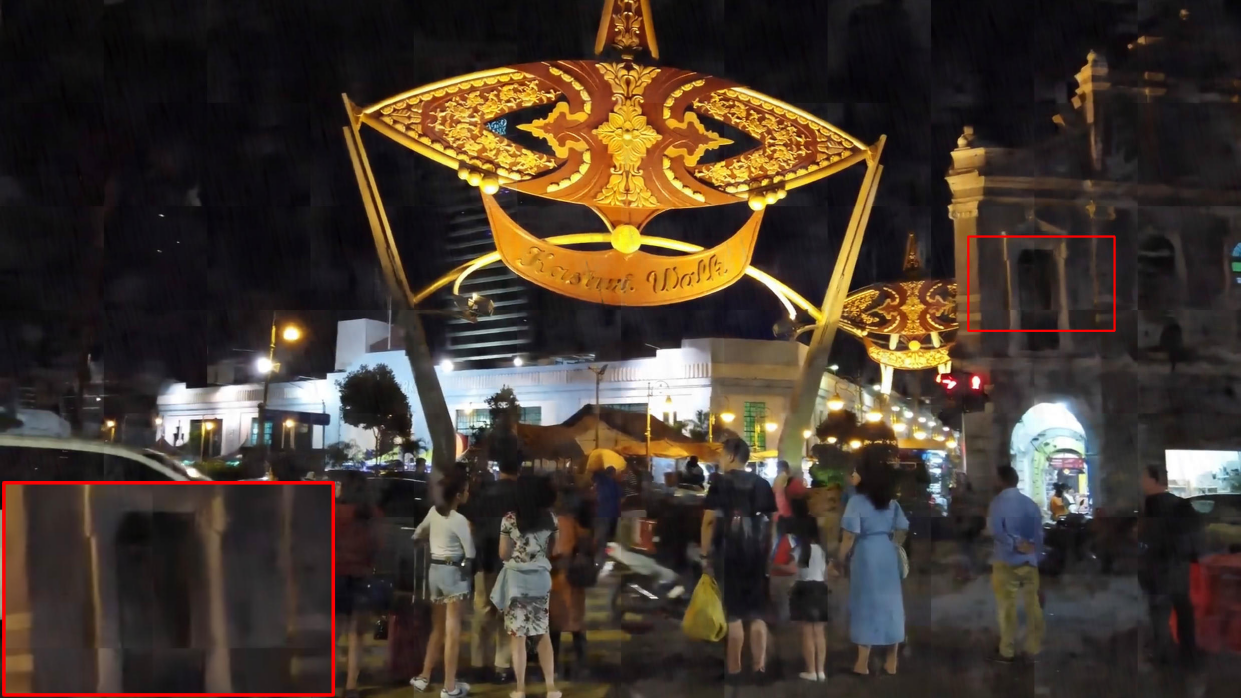} &
    \includegraphics[width=0.135\textwidth]{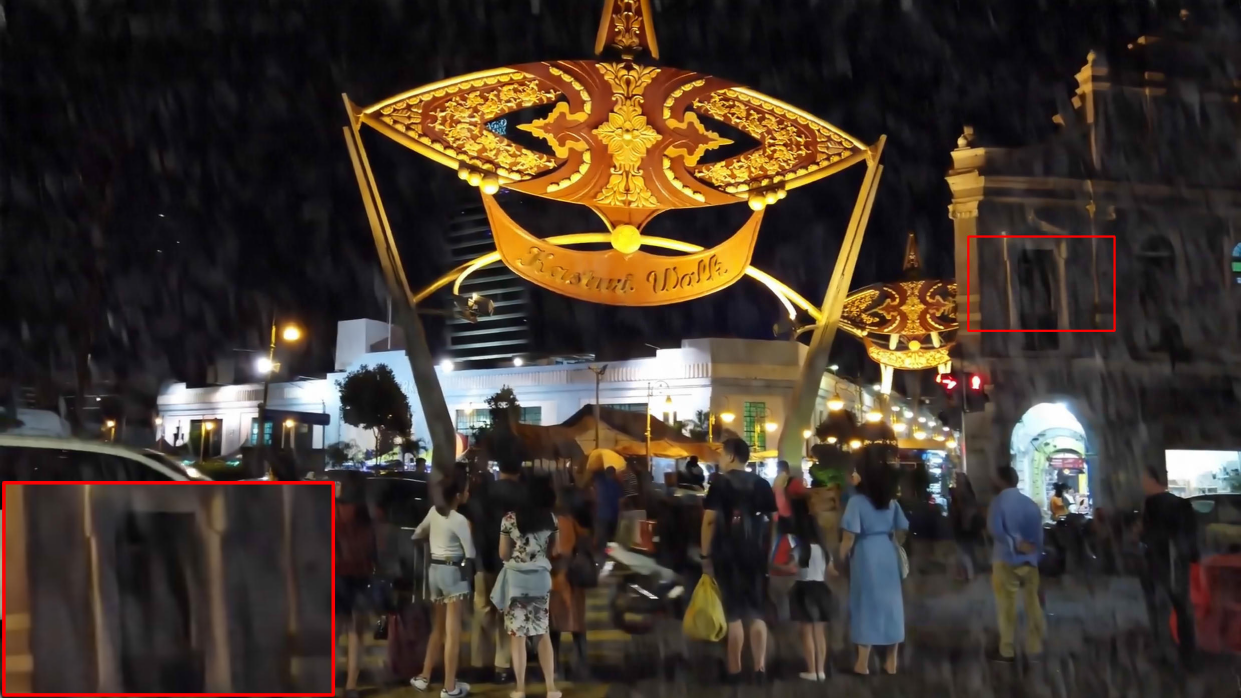} &
    \includegraphics[width=0.135\textwidth]{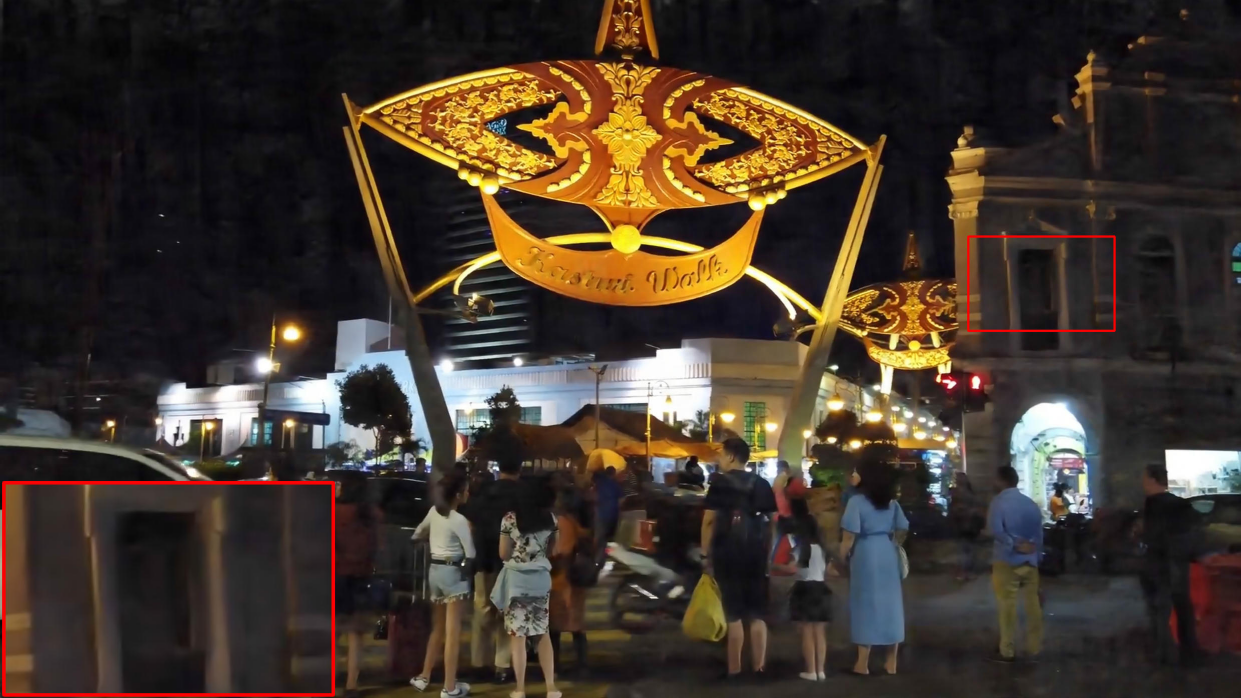} &
    \includegraphics[width=0.135\textwidth]{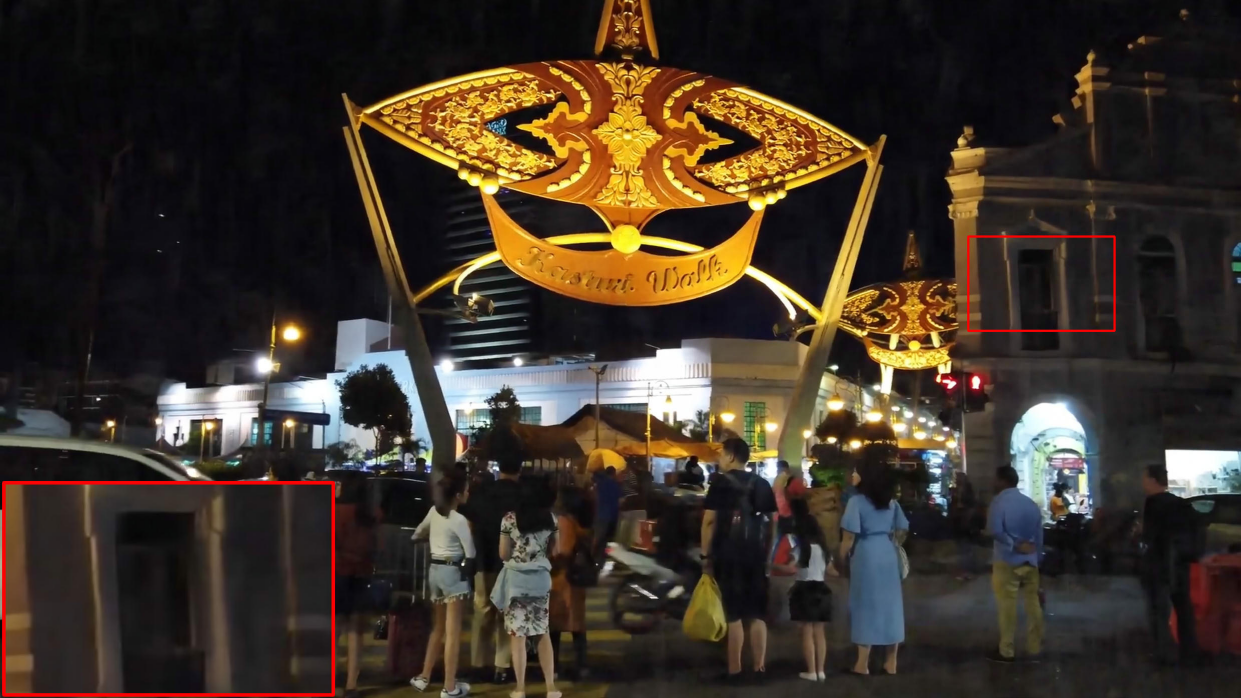} &
    \includegraphics[width=0.135\textwidth]{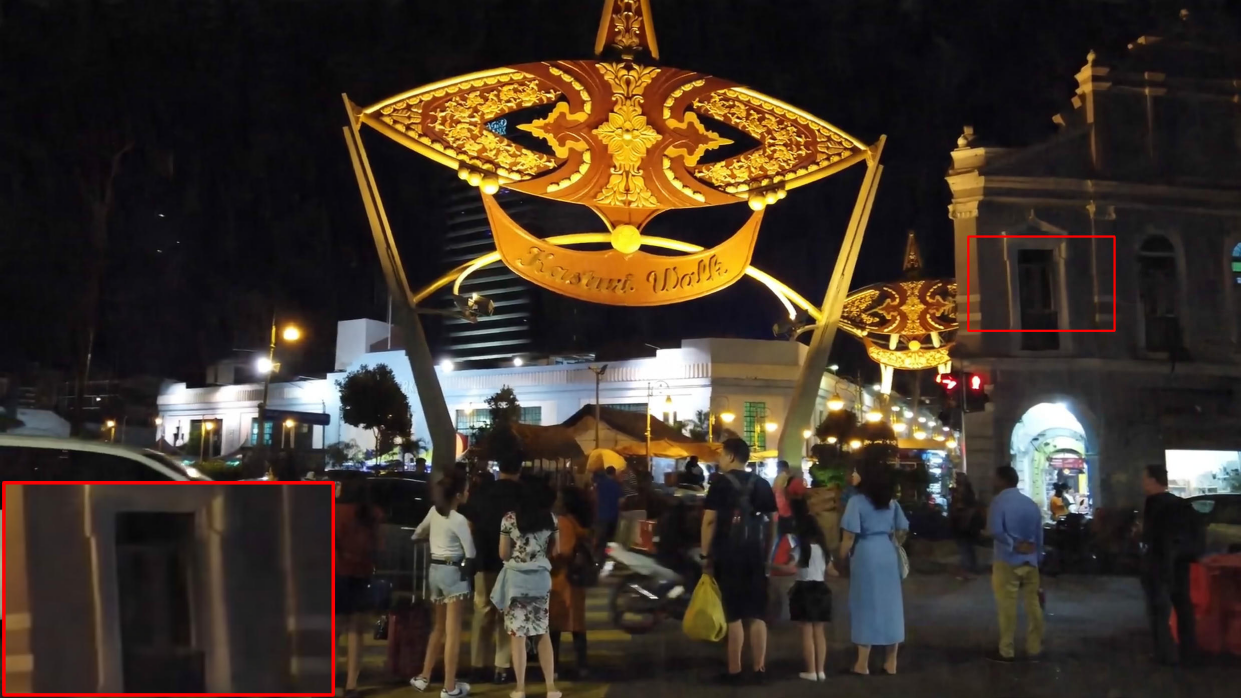} &
    \includegraphics[width=0.135\textwidth]{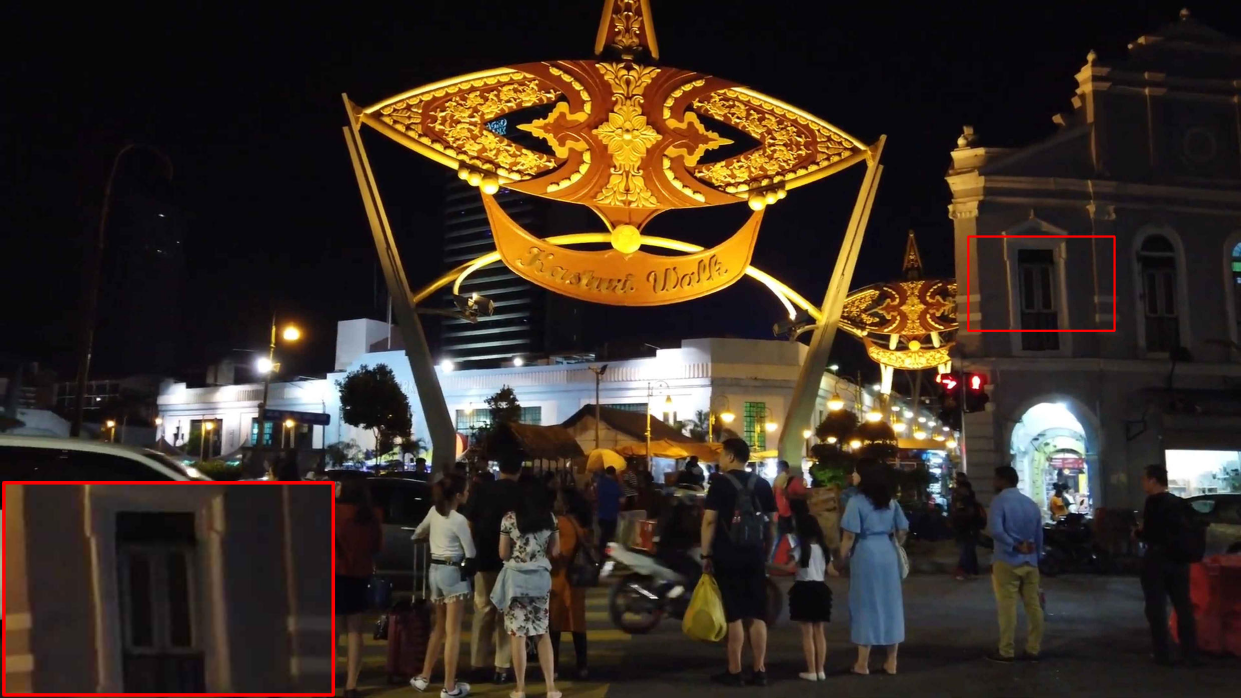} \\
  \end{tabular}

  \caption{Visual comparison between different architectures on 4K-Rain13K dataset \cite{chen2024towards}. Zoomed-in regions display that RetinexDualV2 removes localized rain artifacts optimally compared against recent literature.}
  \label{fig:VisualComparisonRain}
\end{figure*}

\begin{table}[t]
\centering
\begingroup\footnotesize
\setlength{\tabcolsep}{3pt}
\renewcommand{\arraystretch}{0.95}
\caption{Final ranking of NTIRE 2026 The Second Challenge on Day and Night Raindrop Removal for Dual-Focused Images \cite{ntire26ugcvideo}.}
\begin{tabular}{l r r r r r}
\toprule
Team & Final Score & PSNR & SSIM & LPIPS & Params (M) \\
\midrule
AIIA-Lab & 35.2378 & 28.3392 & 0.8265 & 0.2732 & 16.6 \\
raingod & 35.2186 & 28.2817 & 0.8255 & 0.2636 & 16.6 \\
BUU\_CV & 35.0360 & 28.1469 & 0.8222 & 0.2665 & 26.89 \\
\textbf{RetinexDualV2} & \textbf{33.8556} & \textbf{27.2379} & \textbf{0.8061} & \textbf{0.2887} & \textbf{4.8} \\
ULR & 33.7505 & 27.0572 & 0.7966 & 0.2547 & 593 \\
GU-day Mate & 32.9288 & 26.5547 & 0.7826 & 0.2903 & 2.14 \\
Derain & 32.7158 & 26.5915 & 0.7882 & 0.3515 & 16.1 \\
NTR & 32.1273 & 26.1084 & 0.7674 & 0.3310 & 142.5 \\
Cidaut AI & 31.9468 & 25.8394 & 0.7653 & 0.3092 & 2.95 \\
NCHU-CVLab & 31.9438 & 25.8288 & 0.7640 & 0.3050 & 29.38 \\
DGLTeam & 31.8563 & 25.7230 & 0.7682 & 0.3098 & 42.6 \\
MMAIrider & 31.6923 & 25.7243 & 0.7655 & 0.3373 & 10.17 \\
Rain-SVNIT & 31.6514 & 25.8245 & 0.7638 & 0.3623 & 26.1 \\
Just JiT & 31.2484 & 25.3451 & 0.7601 & 0.3396 & 953 \\
BITssvgg & 30.9433 & 25.1389 & 0.7495 & 0.3382 & 16.87 \\
\bottomrule
\end{tabular}
\label{tab:raindrop-results}
\endgroup
\end{table}

\begin{table*}[t]
\centering
\begingroup\footnotesize
\setlength{\tabcolsep}{5pt}
\renewcommand{\arraystretch}{0.92}
\caption{Final ranking of NTIRE 2026 Joint Noise Low-light enhancement Challenge \cite{ntire26llie}.}
\begin{tabular}{p{3.0cm} r r r r  r r r r r}
\toprule
 & \multicolumn{4}{c}{Reference-based} & \multicolumn{5}{c}{No-reference} \\
\cmidrule(lr){2-5} \cmidrule(lr){6-10}
Team & PSNR $\uparrow$ & SSIM $\uparrow$ & LPIPS $\downarrow$ & R.Score $\uparrow$ & NIQE $\downarrow$ & MUSIQUE $\uparrow$ & CLIPIA $\uparrow$ & MoS $\uparrow$ & W.Score $\uparrow$ \\
\midrule
BAU-Vision      & 20.84 & 0.68 & 0.44 & 98.70 & 7.30  & 23.64 & 0.19 & 8.25 & 65.79 \\
MiVideoDLLIE    & 20.75 & 0.67 & 0.47 & 94.02 & 5.82  & 23.10 & 0.26 & 5.50 & 53.22 \\
DH-XHDL-Team    & 19.87 & 0.70 & 0.48 & 90.46 & 7.15  & 24.88 & 0.27 & 8.50 & 75.34 \\
APRIL-AIGC      & 18.94 & 0.68 & 0.44 & 87.81 & 7.76  & 22.67 & 0.31 & 6.00 & 54.09 \\
\textbf{RetinexDualV2} & \textbf{18.69} & \textbf{0.65} & \textbf{0.50} & \textbf{77.46} & \textbf{7.49}  & \textbf{23.31} & \textbf{0.24} & \textbf{5.75} & \textbf{50.45} \\
Lucky one       & 20.23 & 0.67 & 0.60 & 76.62 & 10.83 & 19.43 & 0.30 & 5.25 & 34.46 \\
NTR             & 17.95 & 0.65 & 0.55 & 68.40 & 8.71  & 21.93 & 0.22 & 2.00 & 17.38 \\
VesperLux       & 17.94 & 0.59 & 0.54 & 64.78 & 8.74  & 21.88 & 0.26 & 6.05 & 47.72 \\
WIRNet          & 17.35 & 0.57 & 0.50 & 63.76 & 6.62  & 24.74 & 0.31 & 5.00 & 54.56 \\
PSU TEAM        & 17.18 & 0.59 & 0.54 & 60.47 & 7.12  & 23.53 & 0.27 & 4.50 & 44.91 \\
KLETech-CEVI    & 17.17 & 0.60 & 0.57 & 56.60 & 8.02  & 23.00 & 0.26 & 1.25 & 18.49 \\
SOMIS-LAB       & 16.75 & 0.57 & 0.59 & 50.30 & 6.95  & 23.51 & 0.26 & 3.25 & 35.98 \\
weichow         & 15.65 & 0.56 & 0.61 & 41.71 & 6.61  & 26.21 & 0.23 & 4.00 & 45.91 \\
MC2             & 15.13 & 0.25 & 0.74 & 0.00  & 3.53  & 26.86 & 0.46 & 4.00 & 68.97 \\
\bottomrule
\end{tabular}
\label{tab:joint-noise-results}
\endgroup
\end{table*}

\subsection{Experiment Settings}
\textbf{Datasets.} Our model was evaluated on \textbf{4K-Rain13K} dataset \cite{chen2024towards} for deraining, and \textbf{UHD-LL} dataset \cite{Li2023ICLR} for low-light enhancement. Additionally, we participated in NTIRE 2026 The Second Challenge on Day and Night Raindrop Removal for Dual-Focused Images, Joint Noise Low-light enhancement, Nighttime Dehazing, and Shadow Removal, where our method was evaluated on the respective challenge datasets.

\textbf{Implementation details.} RetinexDualV2 model was trained using an NVIDIA H100 GPU with the following settings. Using AdamW optimizer \cite{loshchilov2019decoupledweightdecayregularization} with an initial learning rate of 2e-5, which is gradually reduced to 1e-7 through a cosine annealing learning rate scheduler. The patch size used is 768×768 with a batch size of 5.

\textbf{Evaluation.} We evaluate architectures on the two benchmark datasets primarily using PSNR and SSIM\cite{1284395}, and report model parameter counts to indicate complexity; results on the challenge datasets follow each competition’s official metrics (the additional measures listed in the tables).

\subsection{Quantitative and Qualitative Results}
\textbf{Image Deraining Results.} We train the proposed RetinexDual architecture on the \textbf{4K-Rain13K} dataset for the deraining task. We compare RetinexDual against recent models in the literature, including JORDER-E \cite{yang2019joint}, RCDNet \cite{wang2020model}, SPDNet \cite{9710307}, IDT \cite{xiao2022image}, Restormer \cite{zamir2022restormer}, DRSformer \cite{chen2023learning}, UDR-S2Former \cite{chen2023sparse}, UDR-Mixer \cite{chen2024towards}, and ERR \cite{Zhao_2025_CVPR}. Quantitative results in Table \ref{tab:RainQuantitative} show small differences in PSNR and SSIM among UDR-Mixer, ERR, and our method; nevertheless, RetinexDual attains the highest reported score and delivers superior visual quality on challenging scenes (see Fig. \ref{fig:VisualComparisonRain}). Notably, RetinexDual reaches these results with a substantially smaller model size compared to some competitors (e.g., DRSformer).

\textbf{Low-Light Image Enhancement Results.} Table \ref{tab:LLQuantitative} compares RetinexDual with recent SOTA methods on the \textbf{UHD-LL} benchmark. Among the compared approaches, IFT, SNR-Aware \cite{xu2022snr}, Uformer \cite{wang2022uformer}, Restormer \cite{zamir2022restormer}, DiffLL \cite{jiang2023low} and LLFormer provide varied performance, while UHDFour \cite{Li2023ICLR}, Wave-Mamba \cite{zou2024wavemamba}, UHDFormer \cite{wang2024uhdformer}, ERR \cite{Zhao_2025_CVPR}, and RetinexDual are more competitive for UHD low-light enhancement. Our model outperforms prior SOTA in PSNR and achieves improved SSIM compared to ERR, with qualitative examples shown in Fig. \ref{fig:VisualComparisonLLIE}.

\begin{figure*}[ht]
  \centering
  \setlength{\tabcolsep}{1pt}
  \begin{tabular}{c cc@{\hspace{2pt}} cc@{\hspace{2pt}} cc}
    & \textbf{Input} & \textbf{Ours} & \textbf{Input} & \textbf{Ours} & \textbf{Input} & \textbf{Ours} \\
    \multirow{2}{*}{\rotatebox{90}{\scriptsize Deraining}} &
    \includegraphics[width=0.155\textwidth]{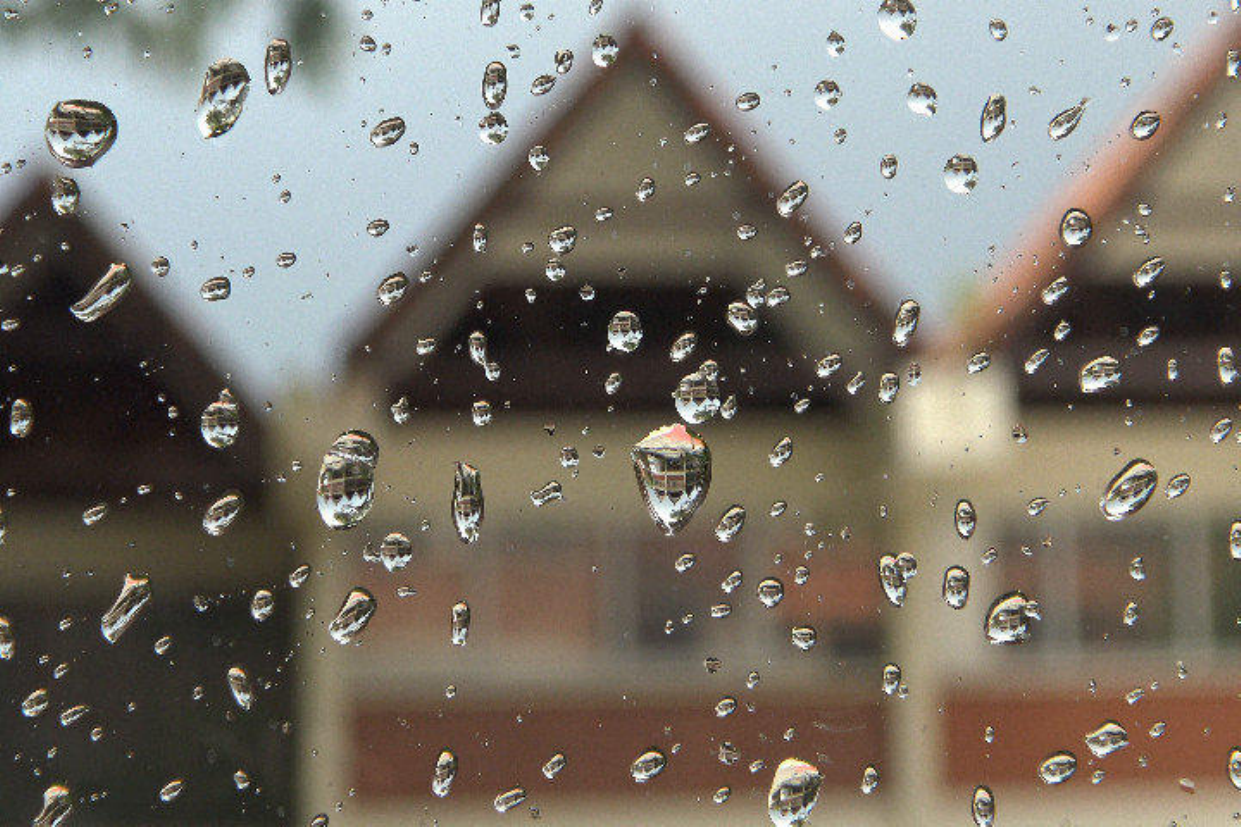} &
    \includegraphics[width=0.155\textwidth]{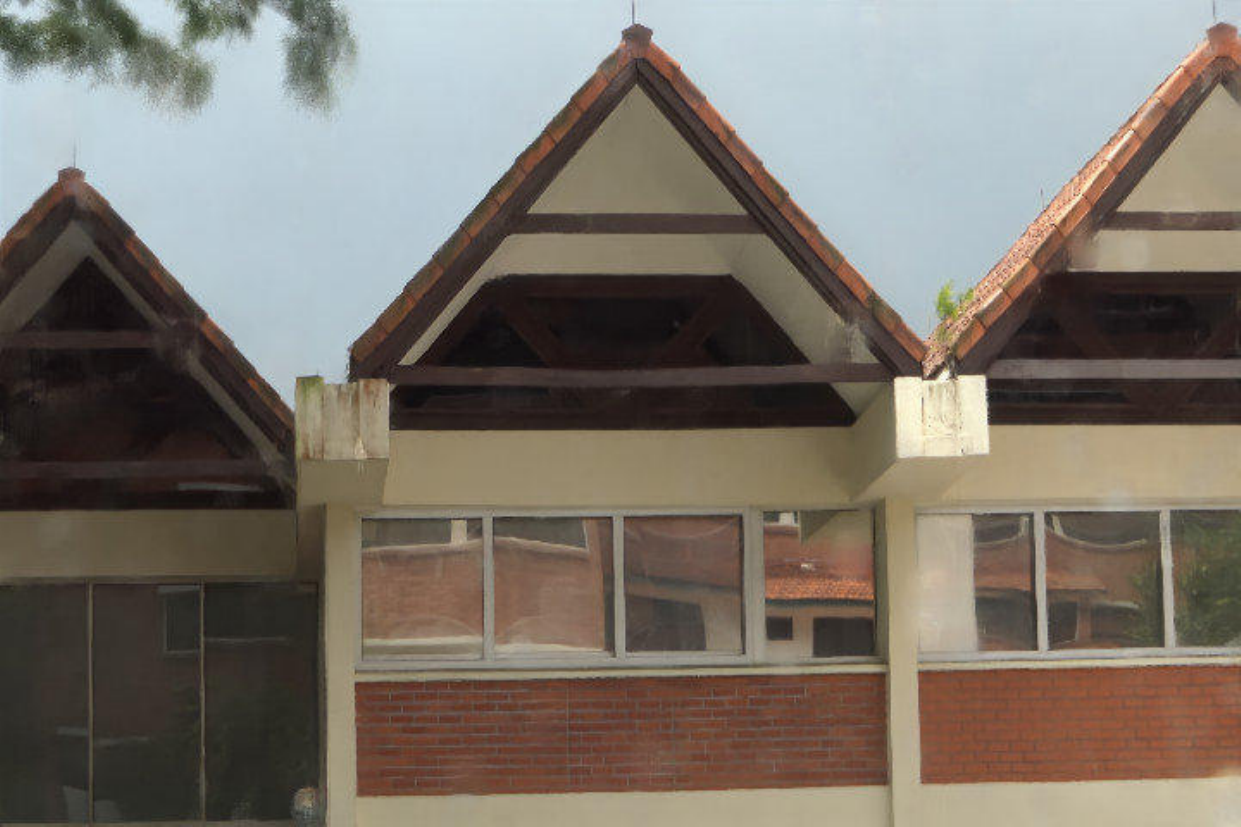} &
    \includegraphics[width=0.155\textwidth]{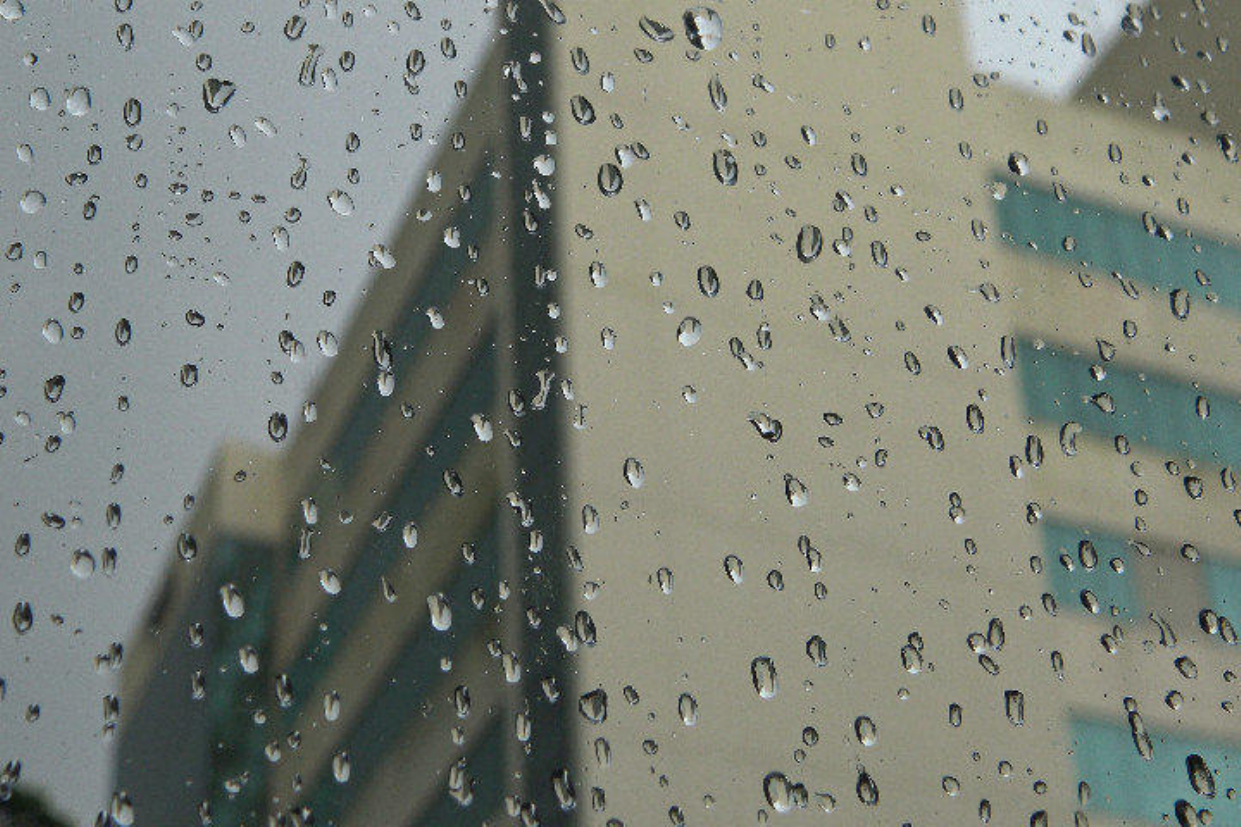} &
    \includegraphics[width=0.155\textwidth]{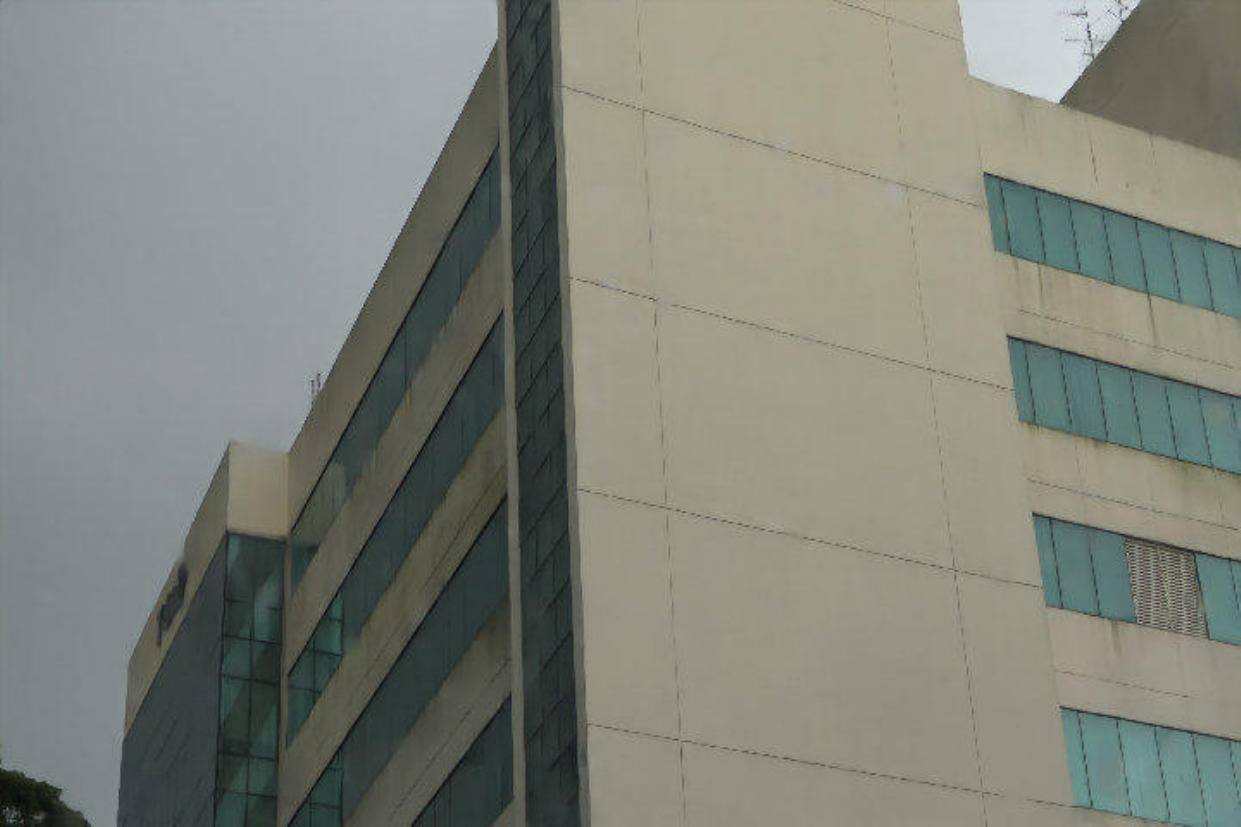} &
    \includegraphics[width=0.155\textwidth]{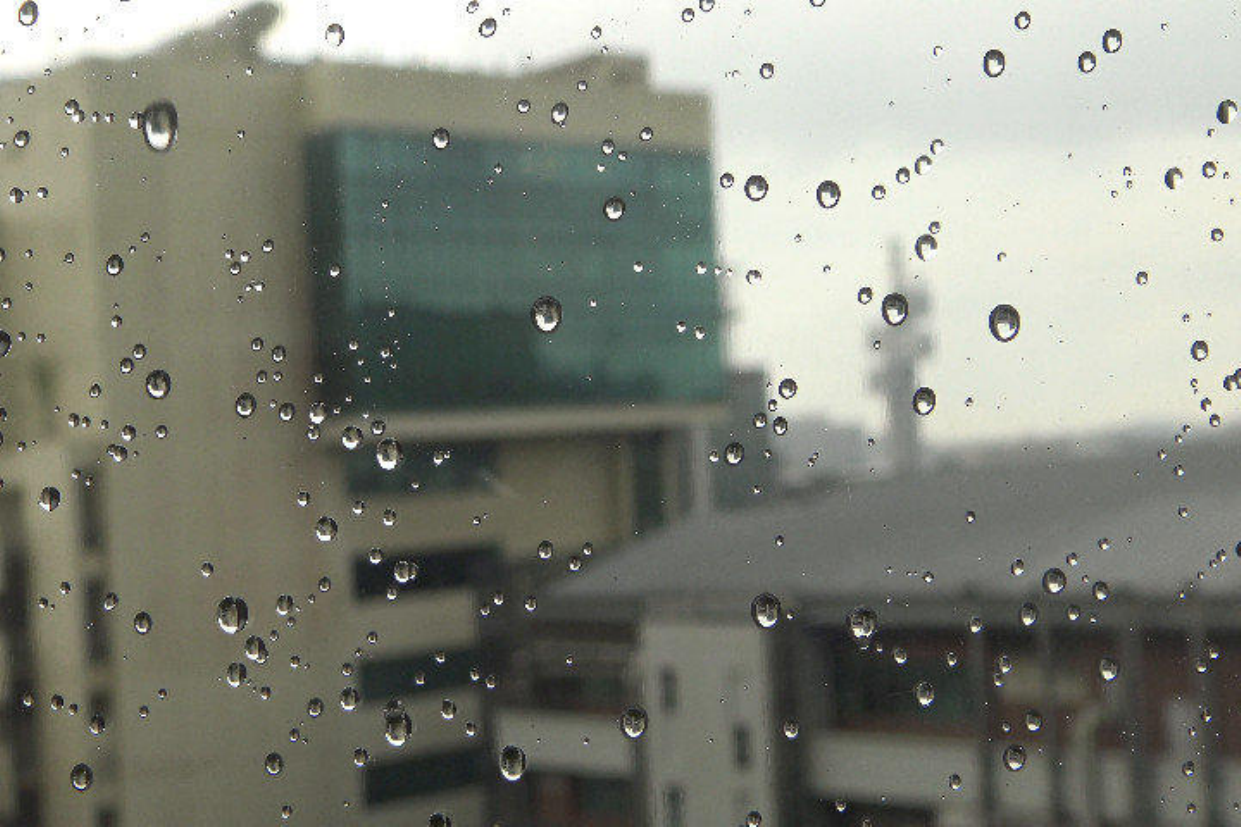} &
    \includegraphics[width=0.155\textwidth]{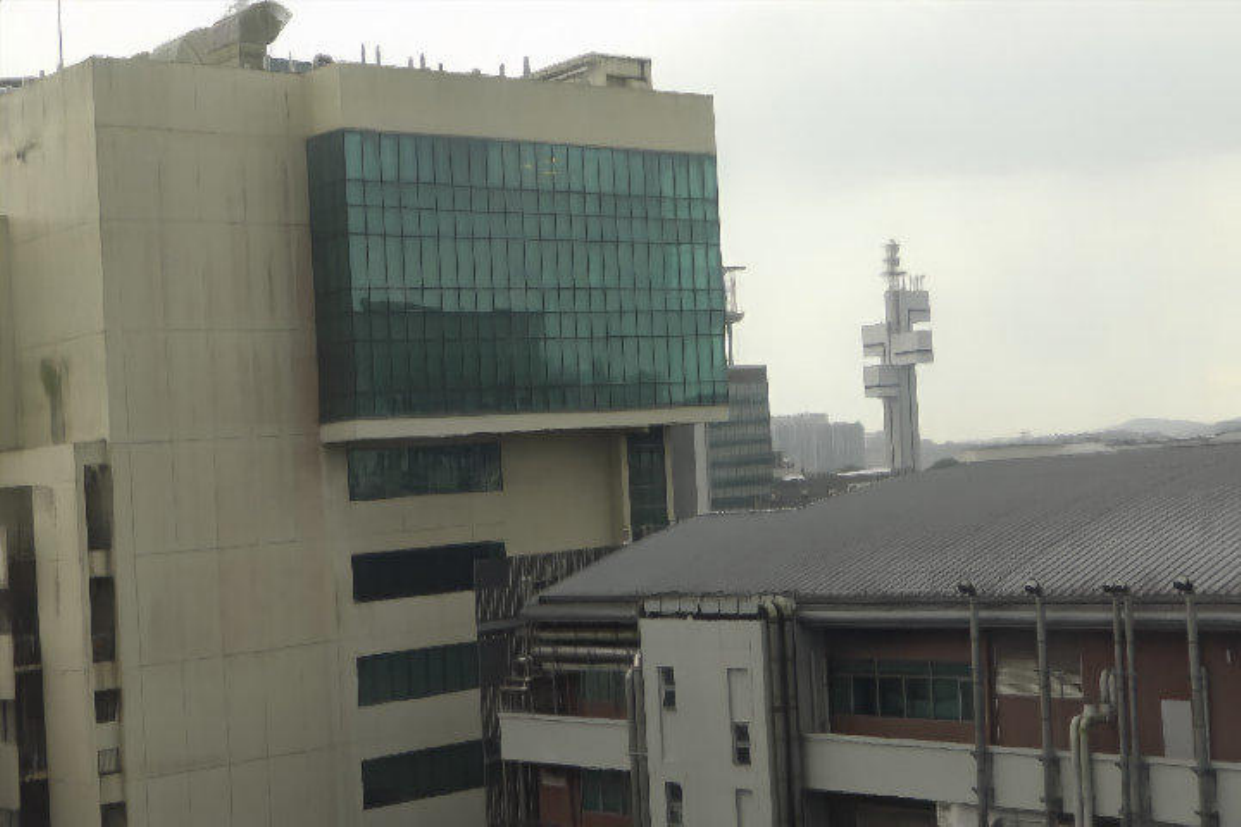} \\[-1mm]
    &
    \includegraphics[width=0.155\textwidth]{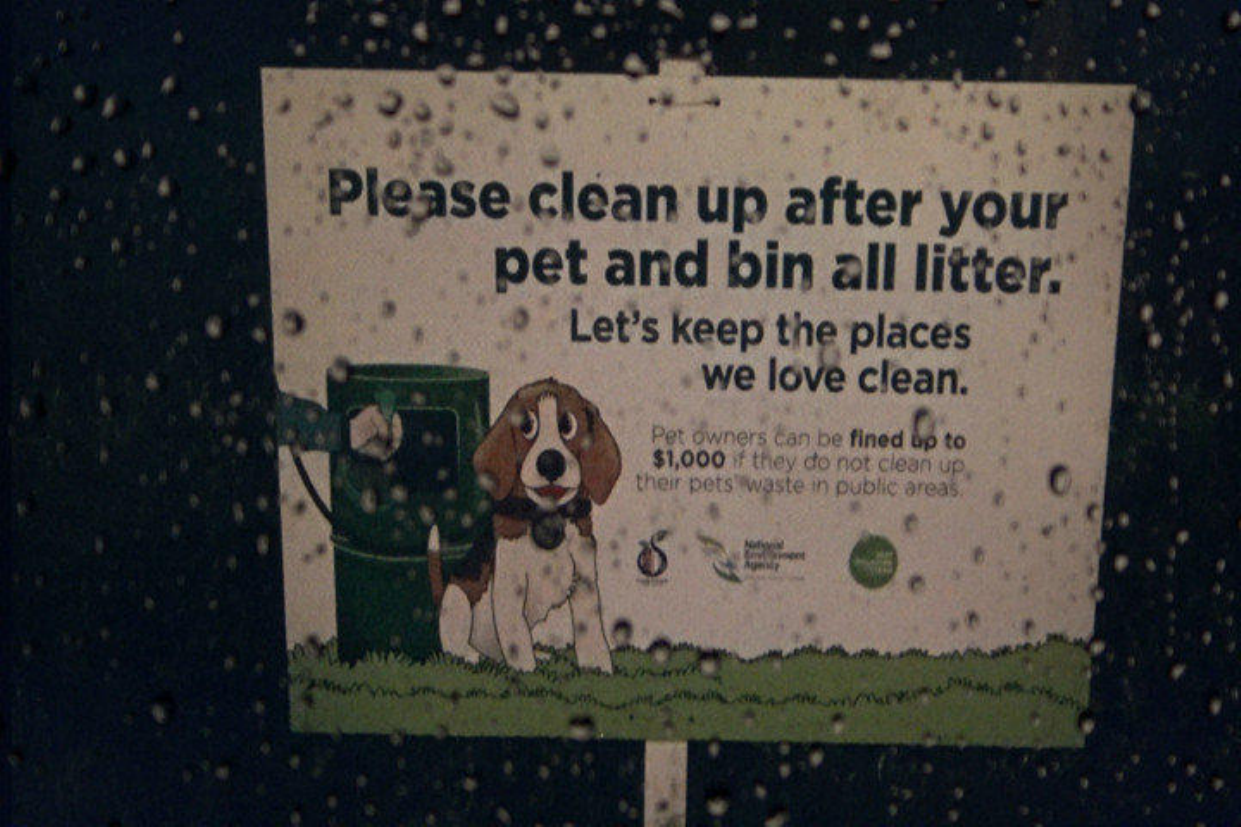} &
    \includegraphics[width=0.155\textwidth]{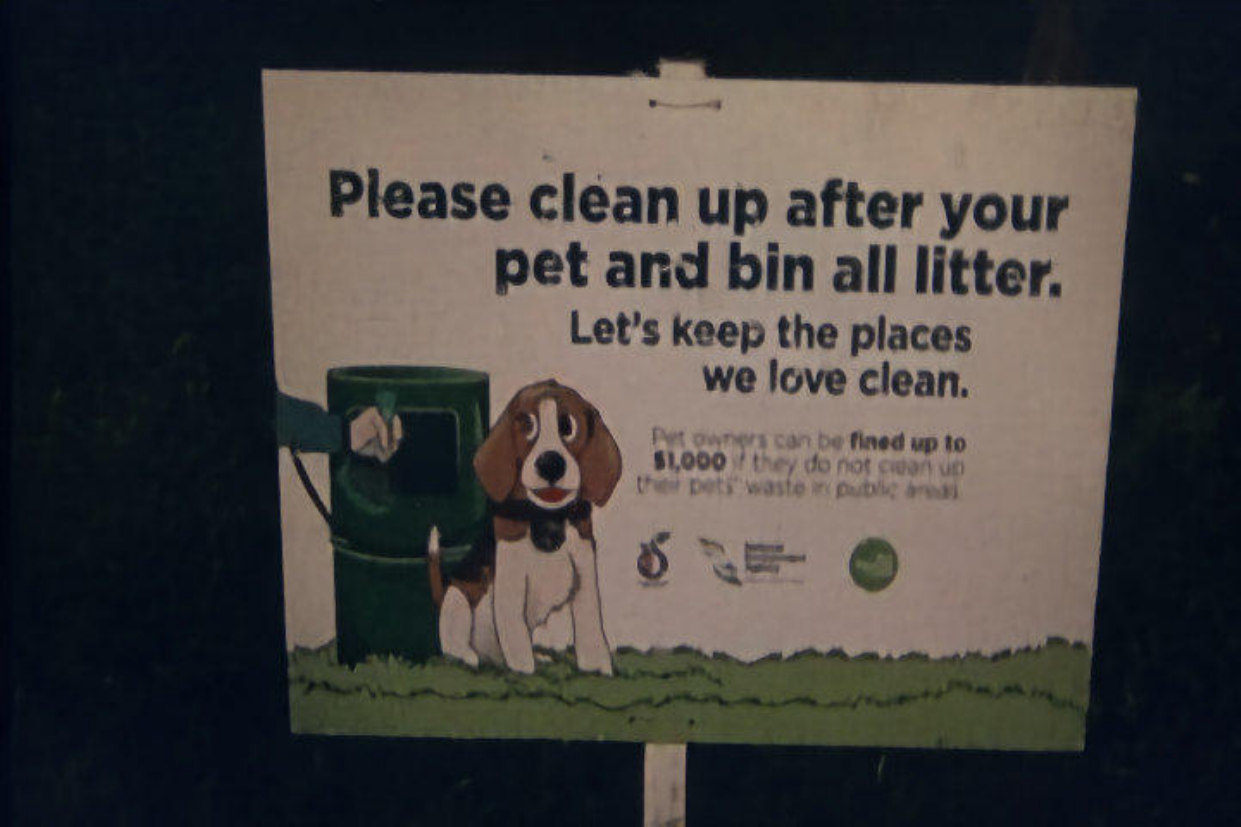} &
    \includegraphics[width=0.155\textwidth]{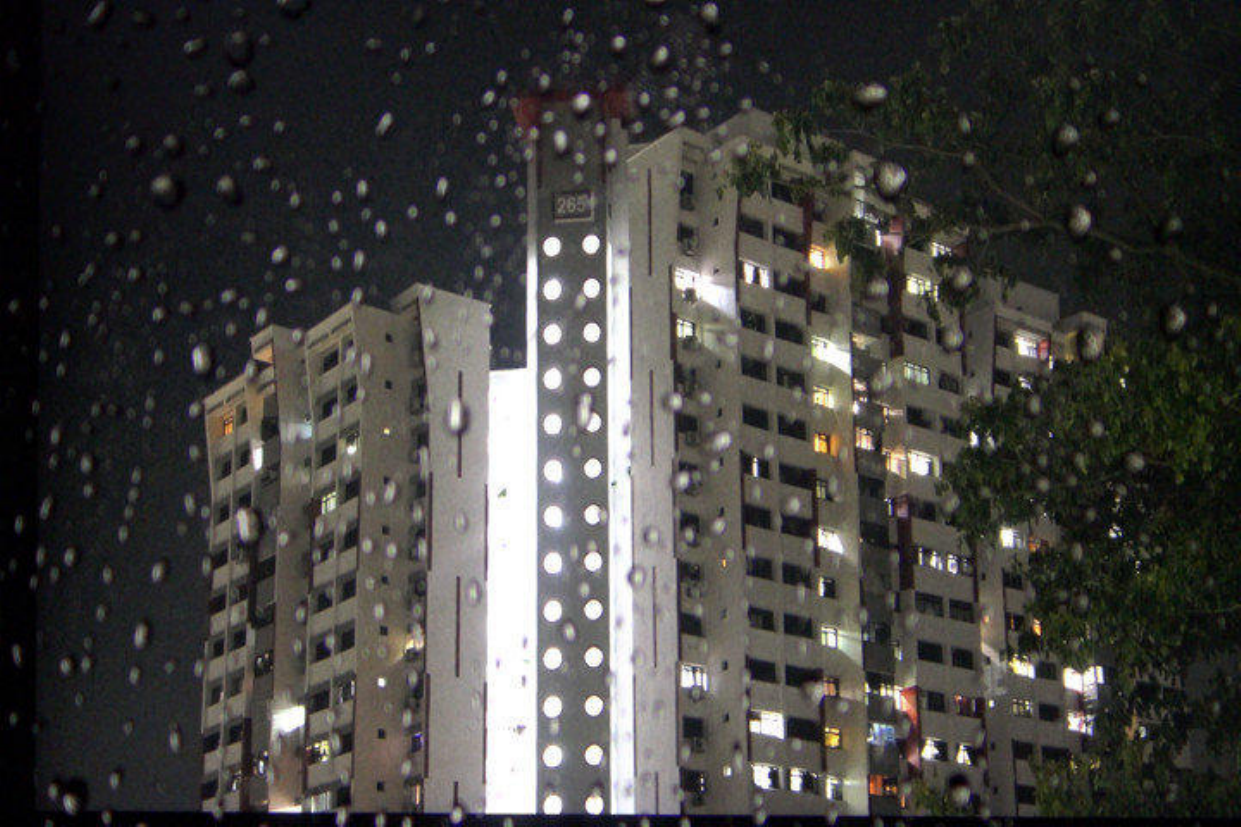} &
    \includegraphics[width=0.155\textwidth]{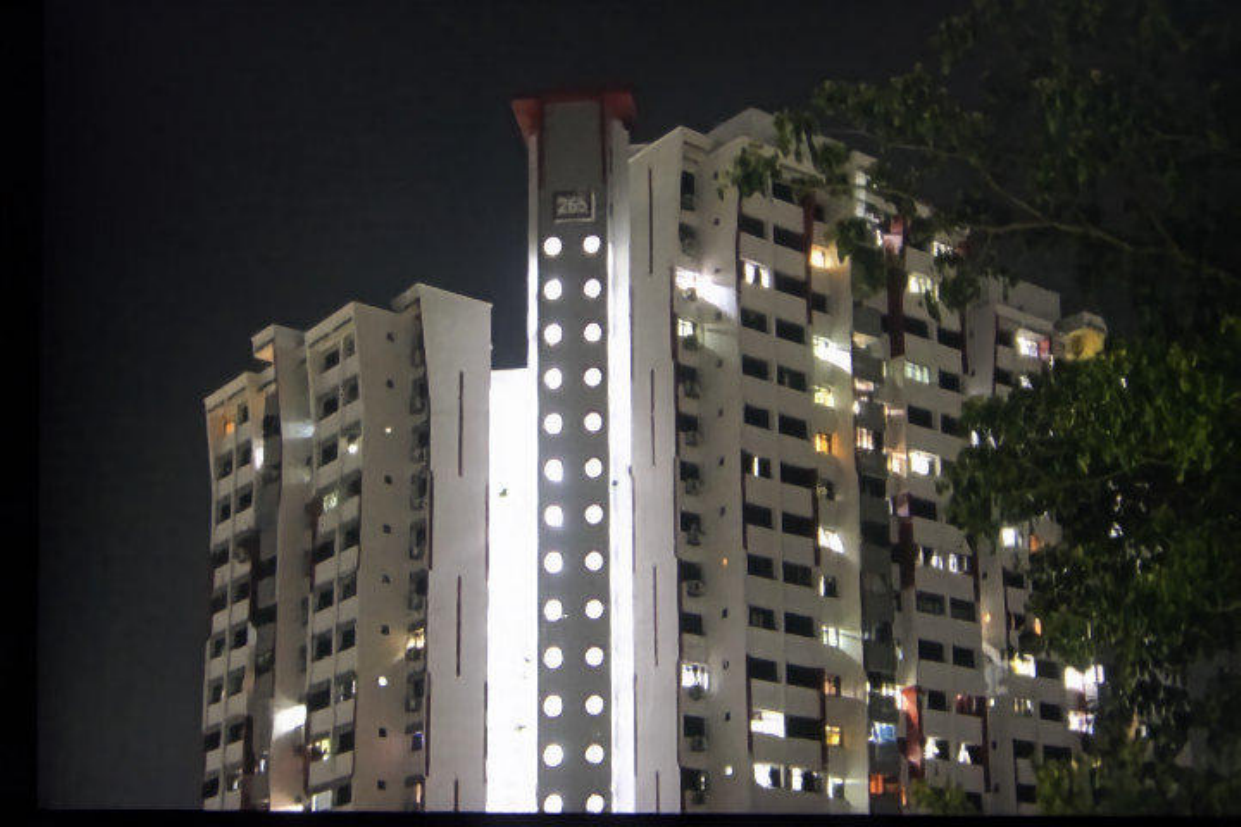} &
    \includegraphics[width=0.155\textwidth]{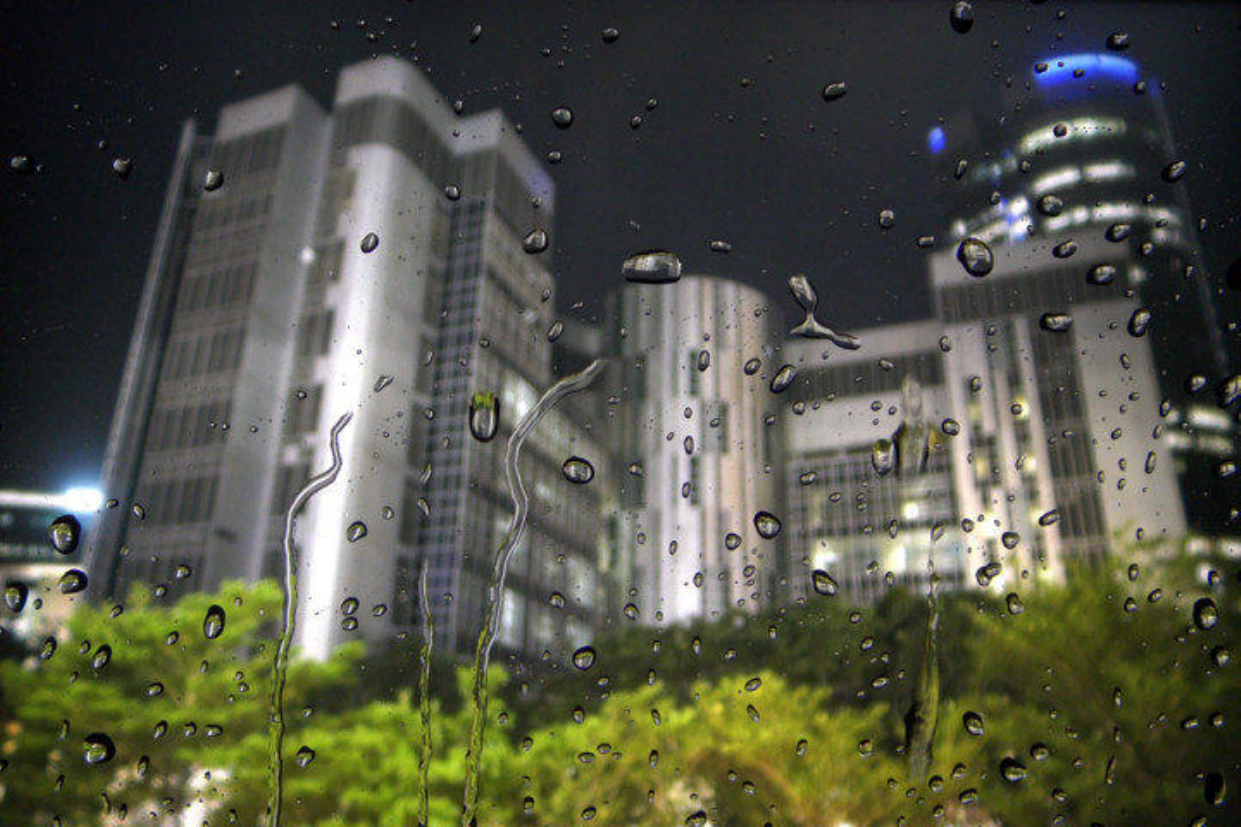} &
    \includegraphics[width=0.155\textwidth]{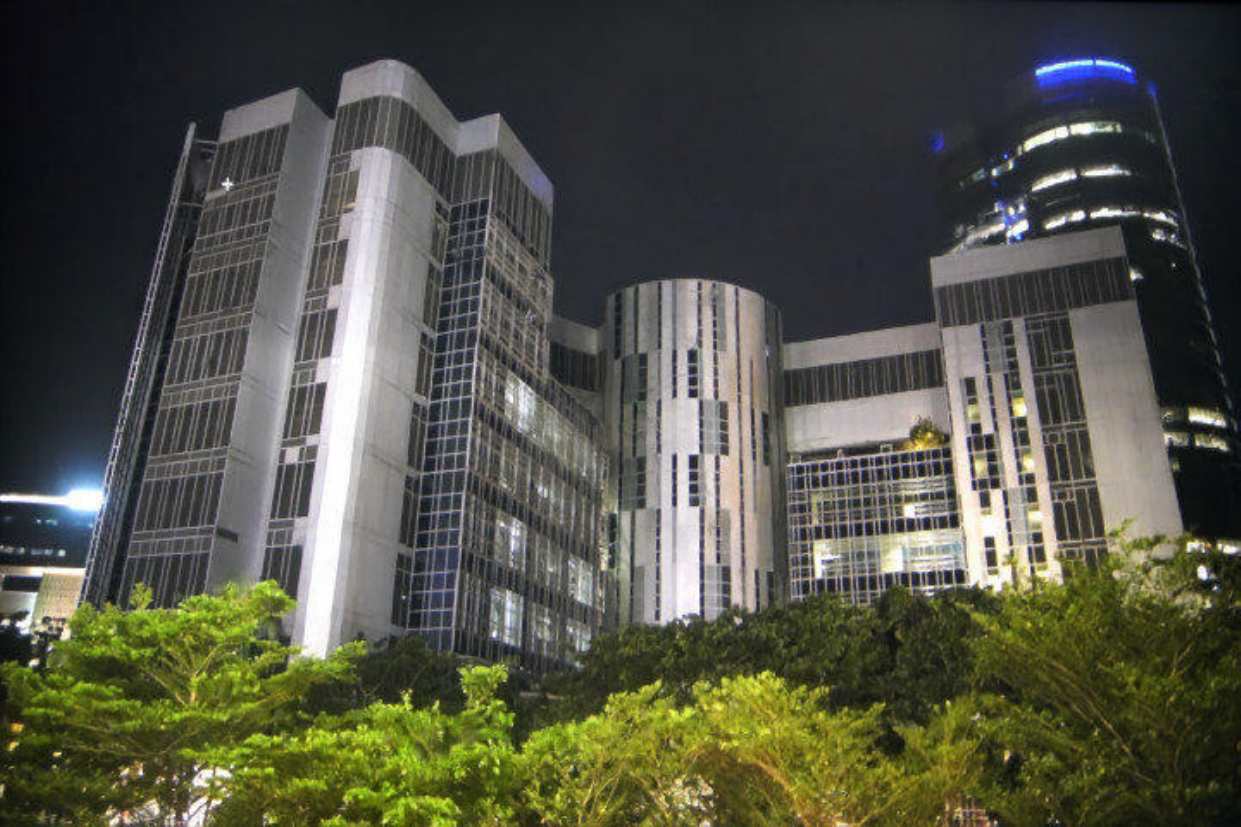} \\[1.5mm]
    
    \multirow{2}{*}{\rotatebox{90}{\scriptsize JNLLIE}} &
    \includegraphics[width=0.155\textwidth]{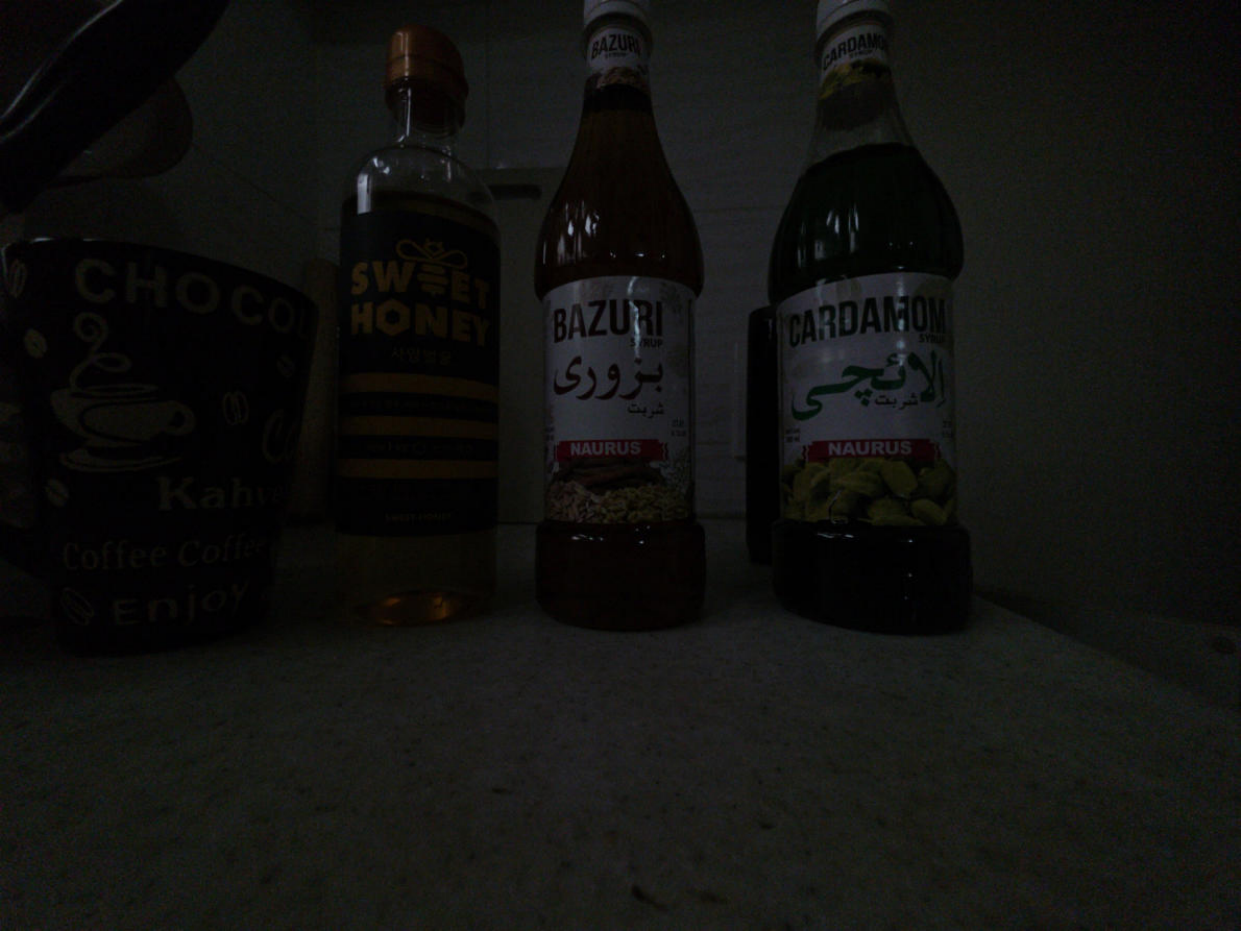} &
    \includegraphics[width=0.155\textwidth]{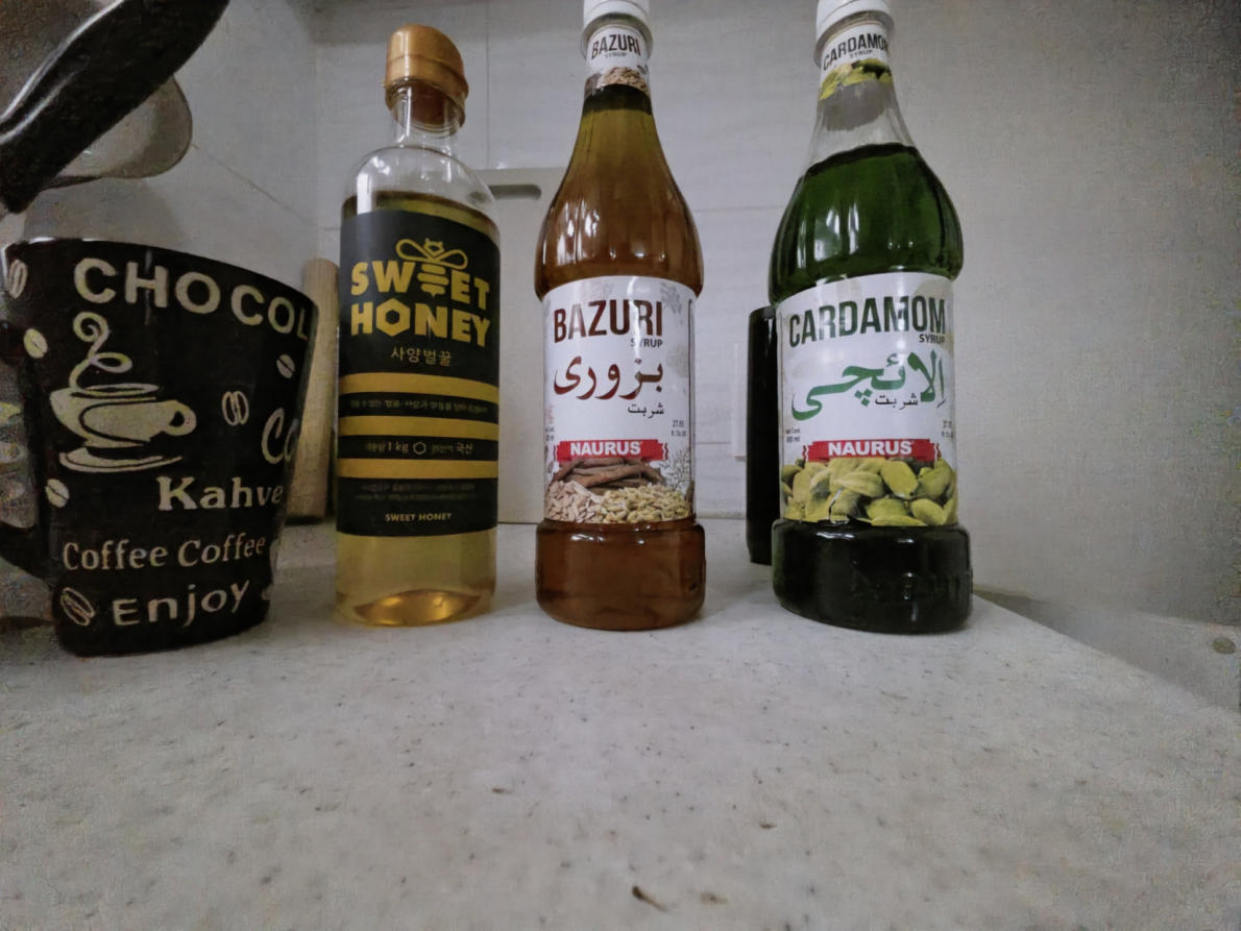} &
    \includegraphics[width=0.155\textwidth]{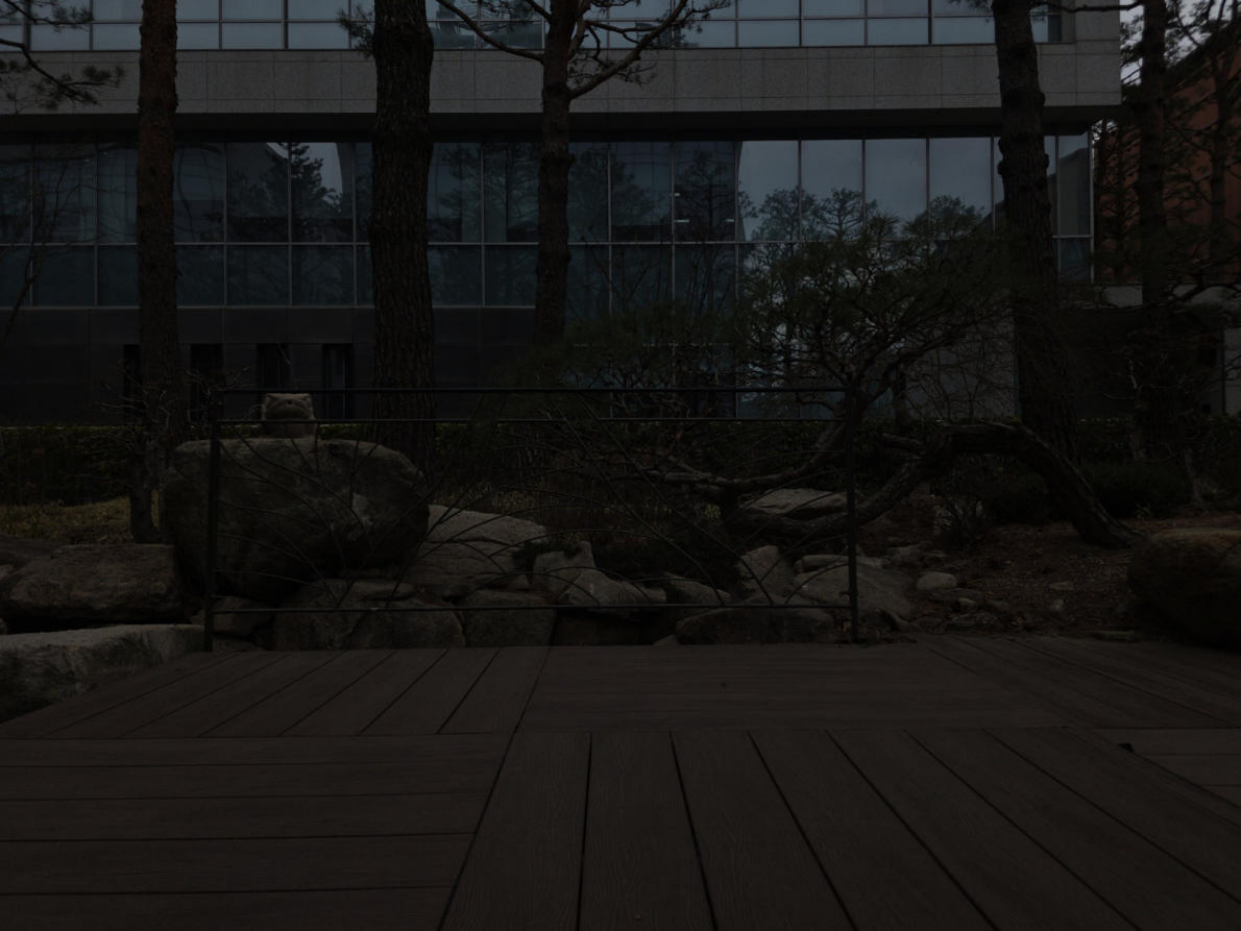} &
    \includegraphics[width=0.155\textwidth]{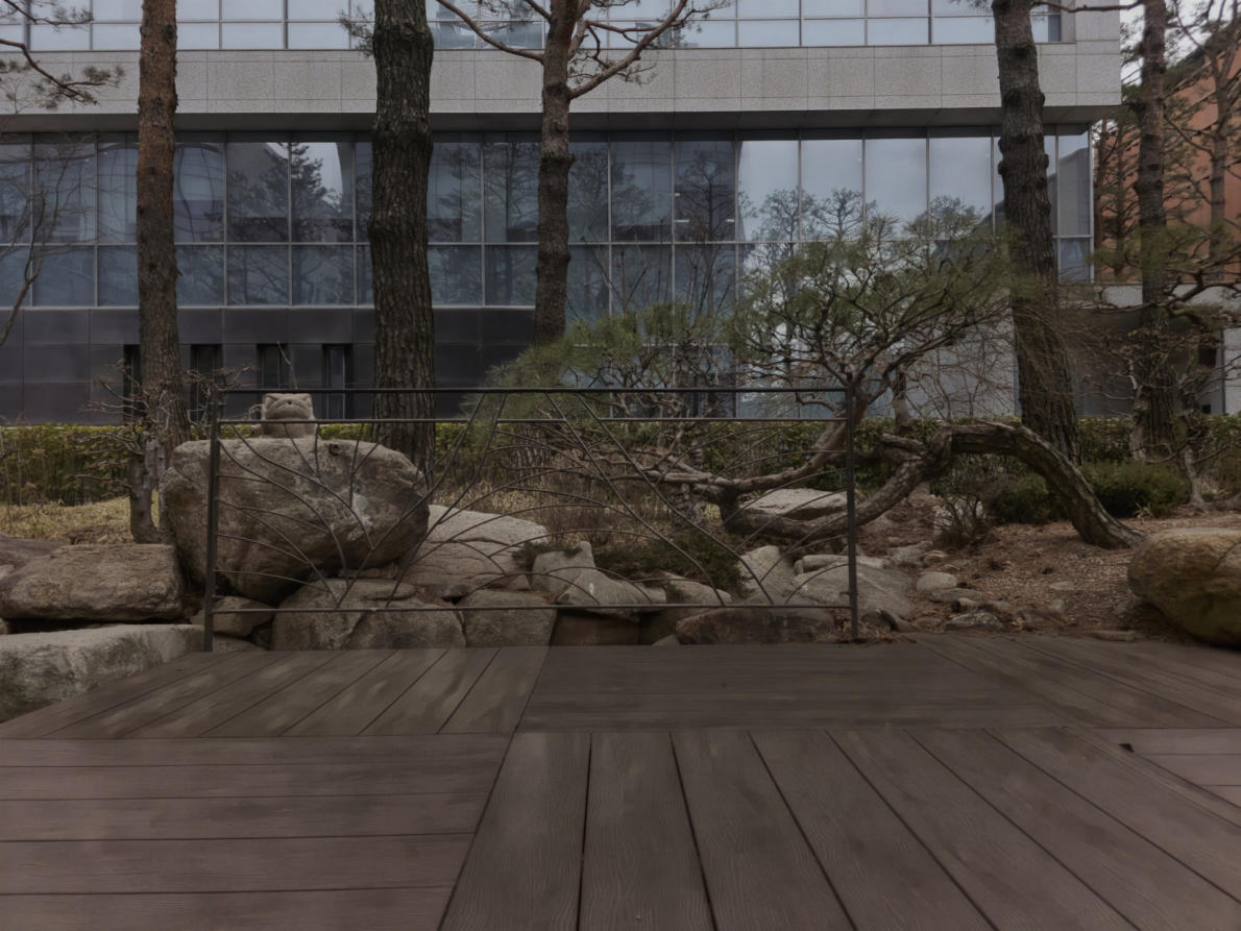} &
    \includegraphics[width=0.155\textwidth]{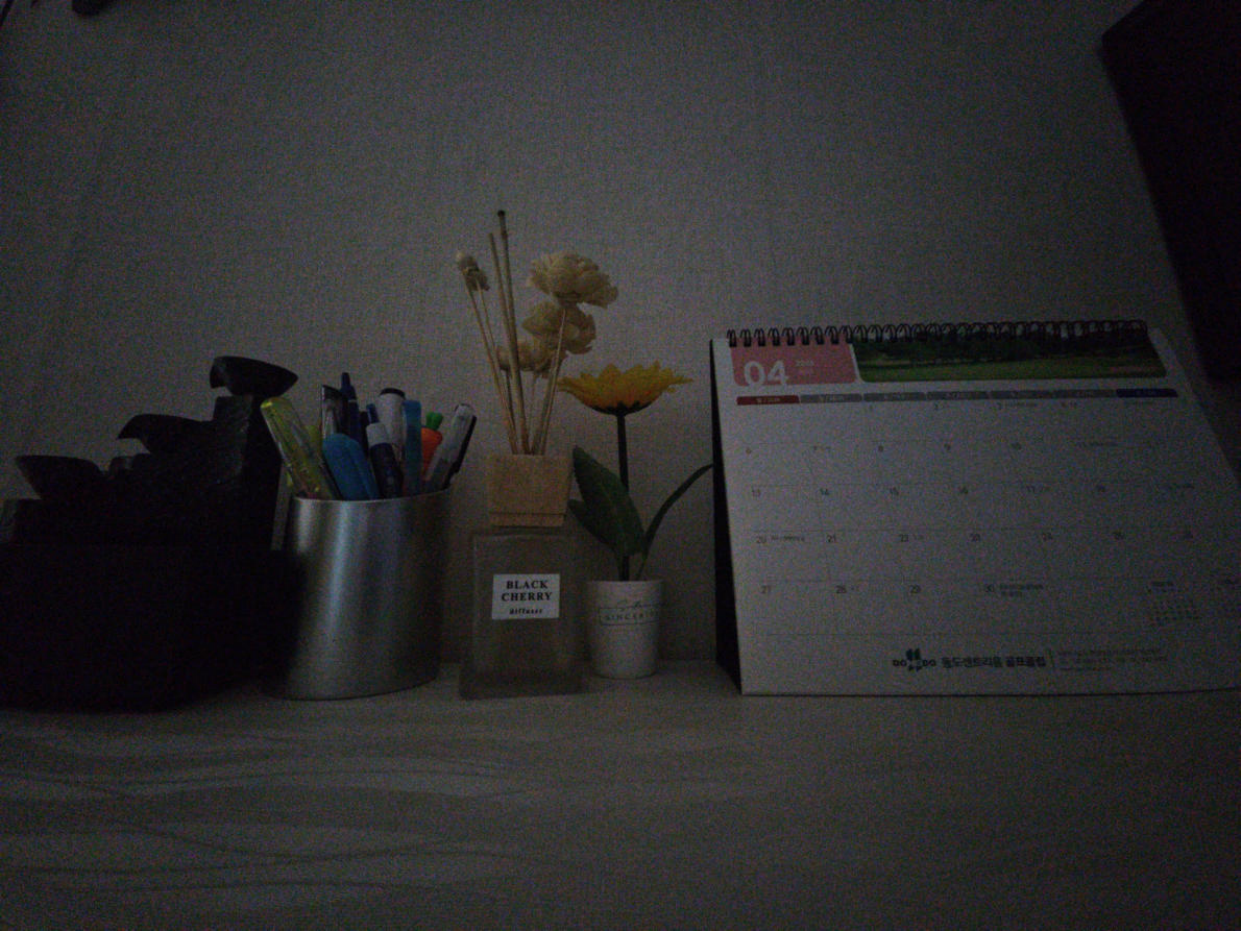} &
    \includegraphics[width=0.155\textwidth]{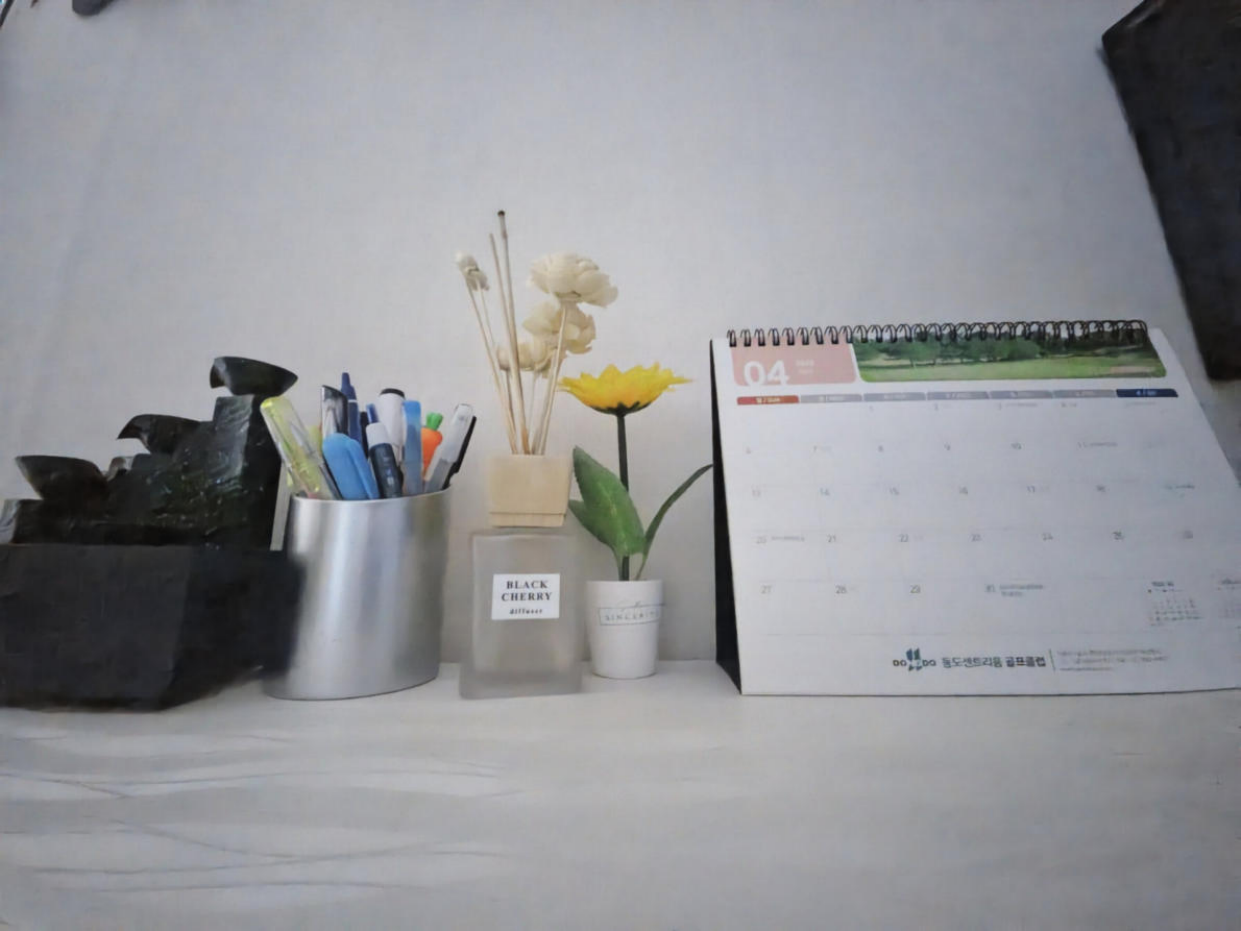} \\[-1mm]
    &
    \includegraphics[width=0.155\textwidth]{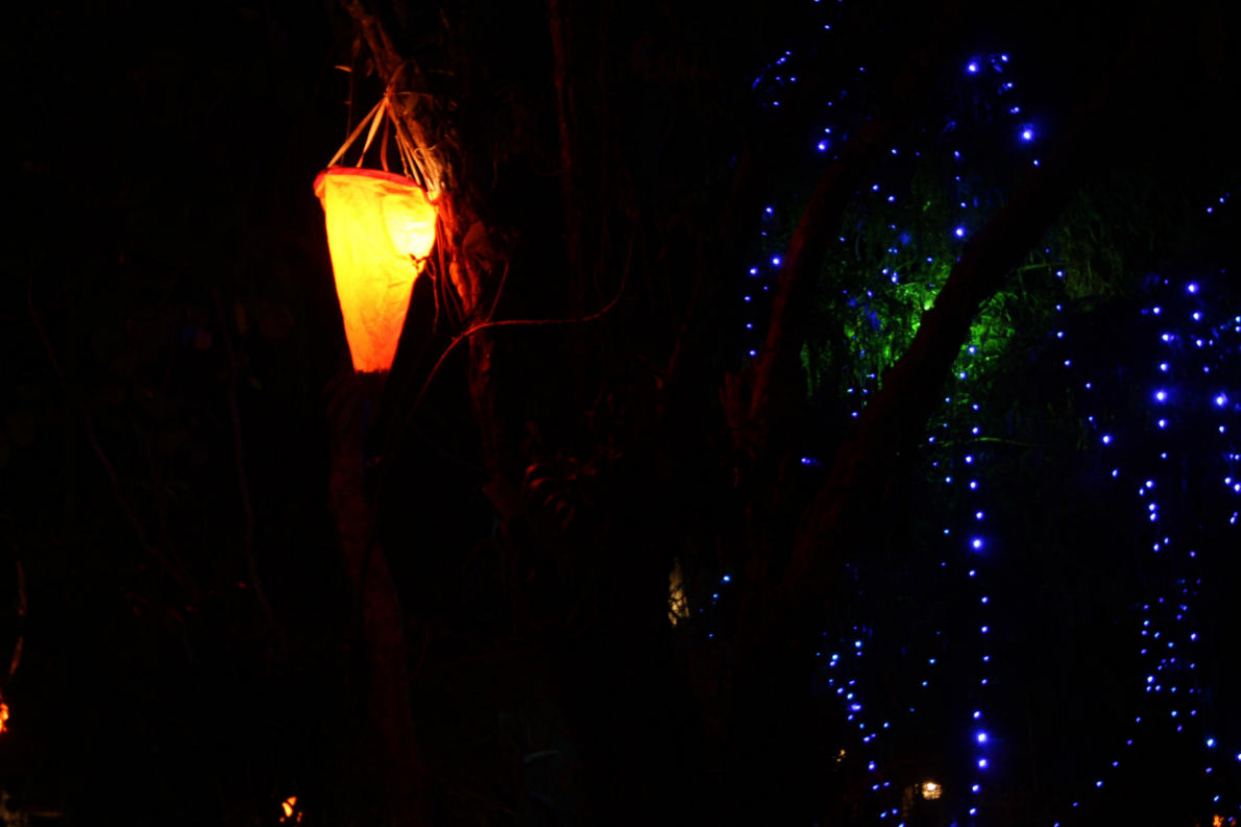} &
    \includegraphics[width=0.155\textwidth]{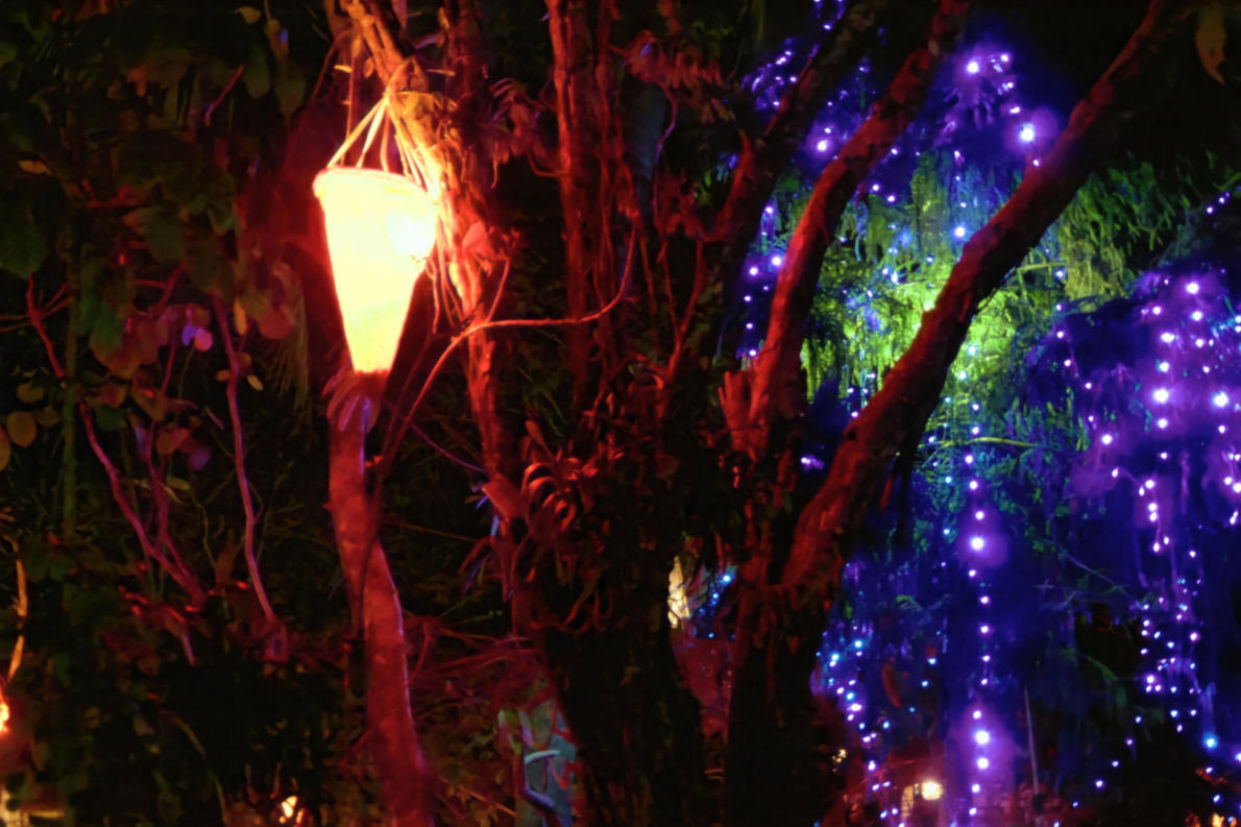} &
    \includegraphics[width=0.155\textwidth]{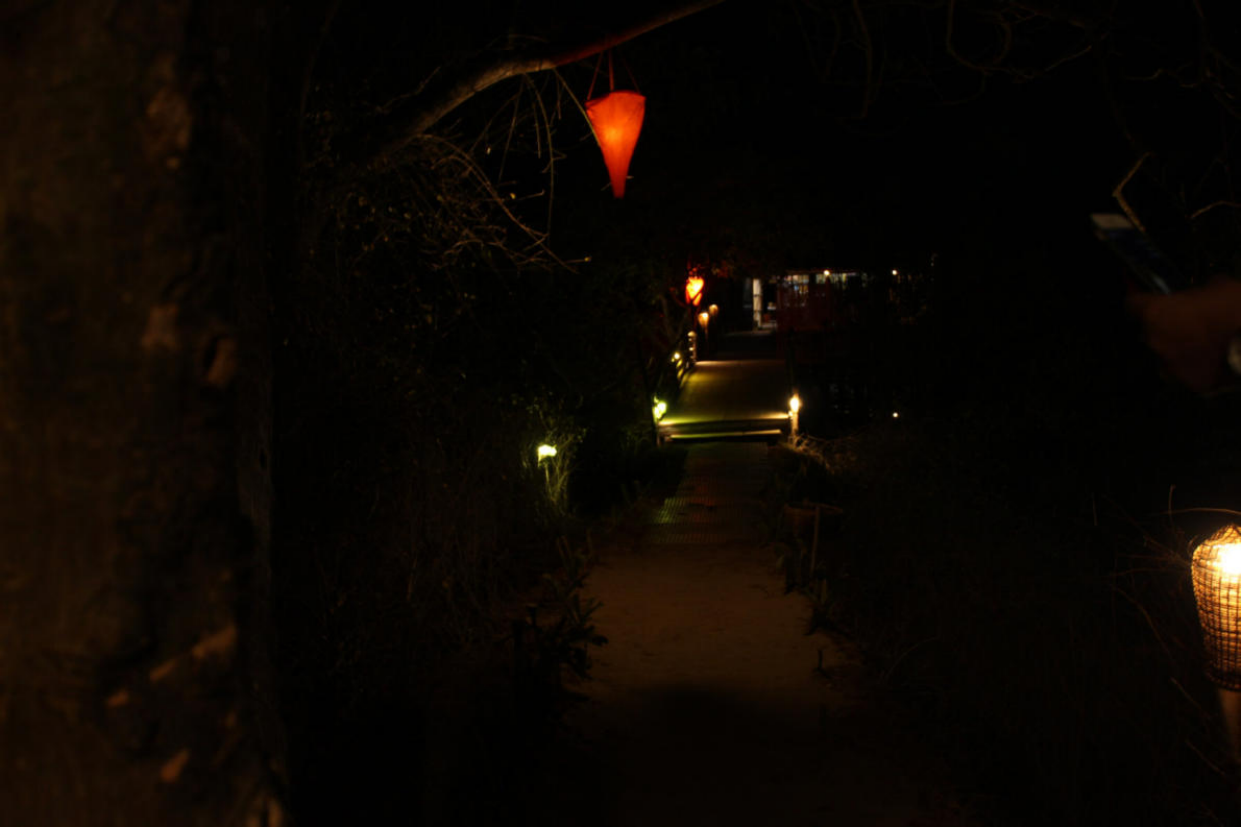} &
    \includegraphics[width=0.155\textwidth]{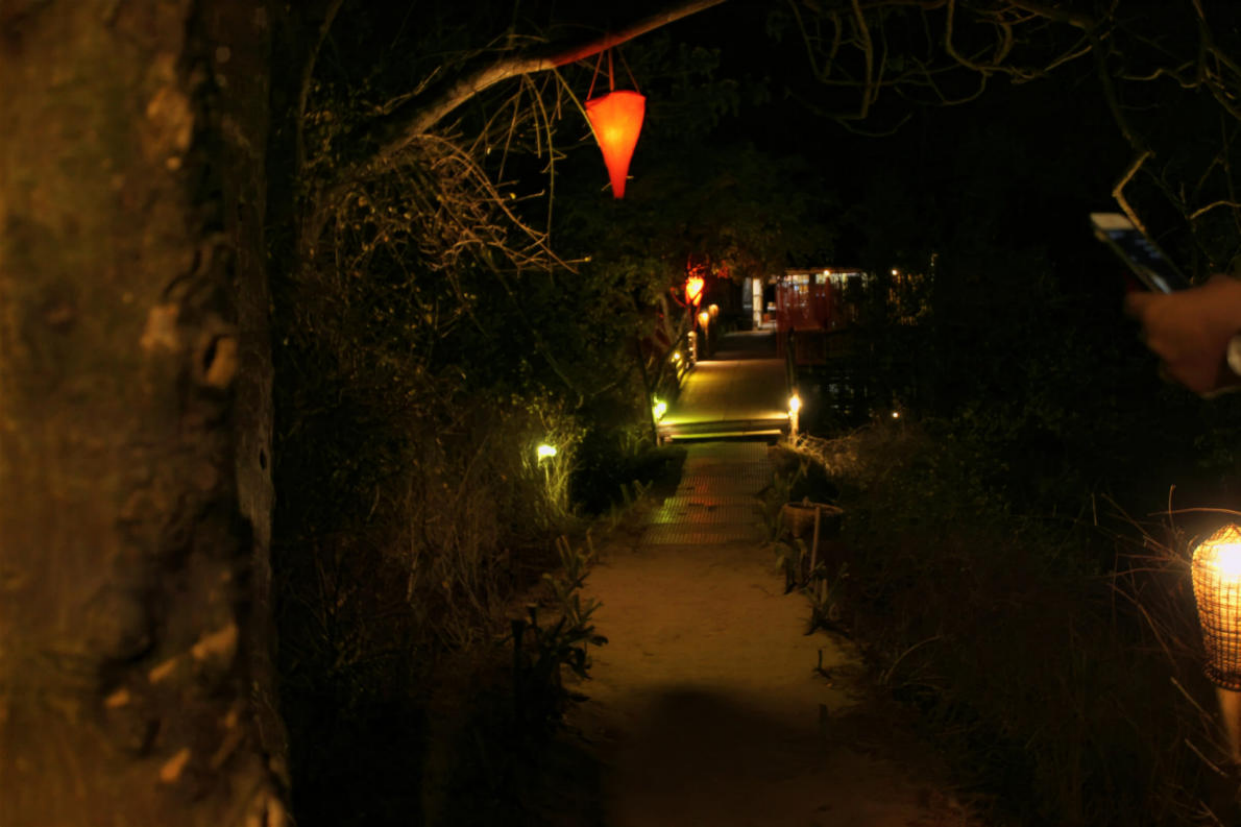} &
    \includegraphics[width=0.155\textwidth]{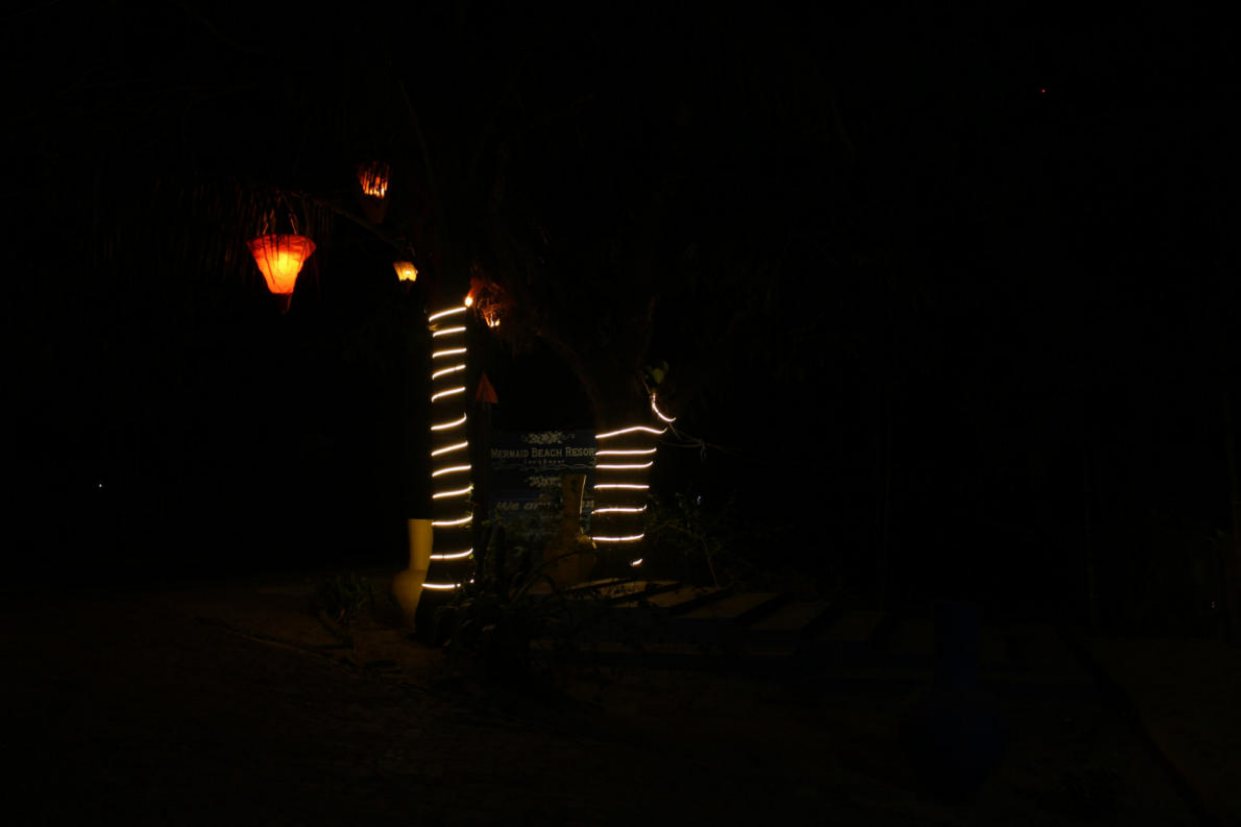} &
    \includegraphics[width=0.155\textwidth]{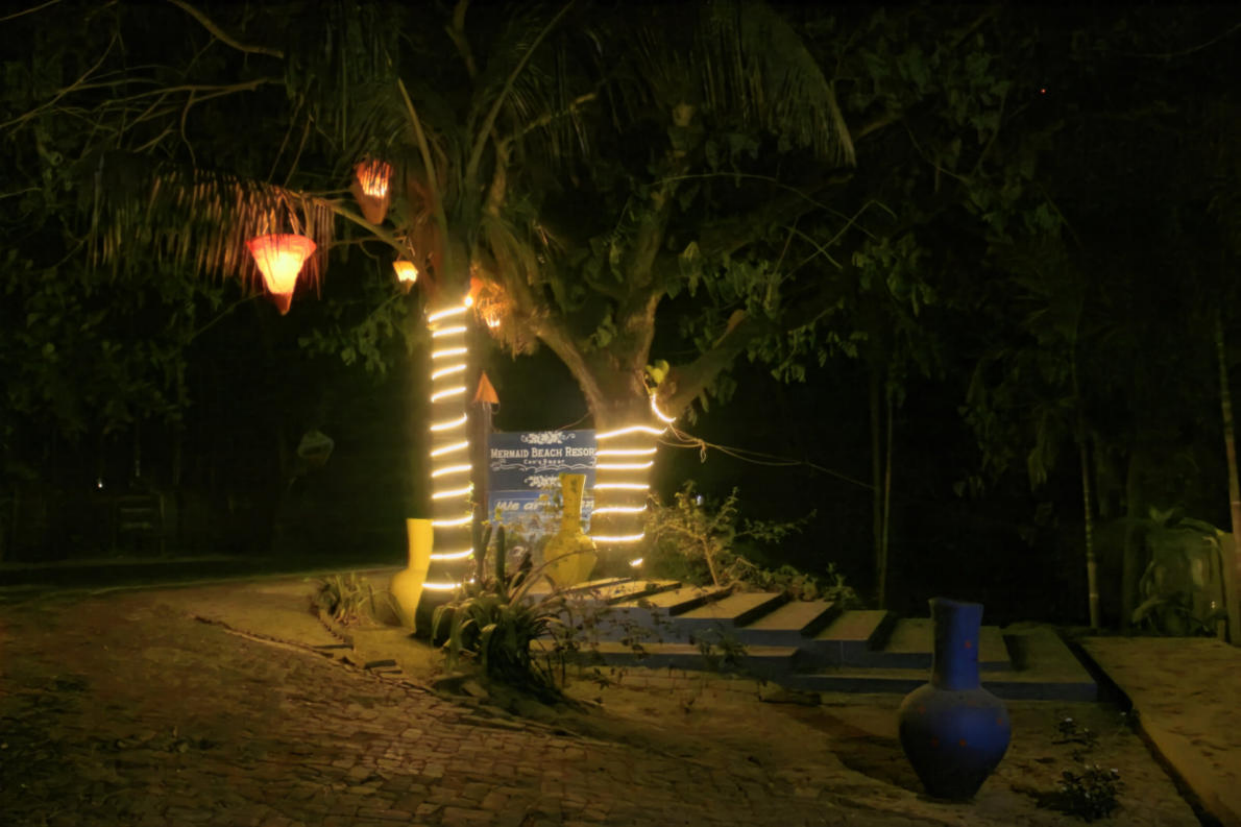} \\[1.5mm]
    
    \multirow{2}{*}{\rotatebox{90}{\scriptsize Deshadowing}} &
    \includegraphics[width=0.155\textwidth]{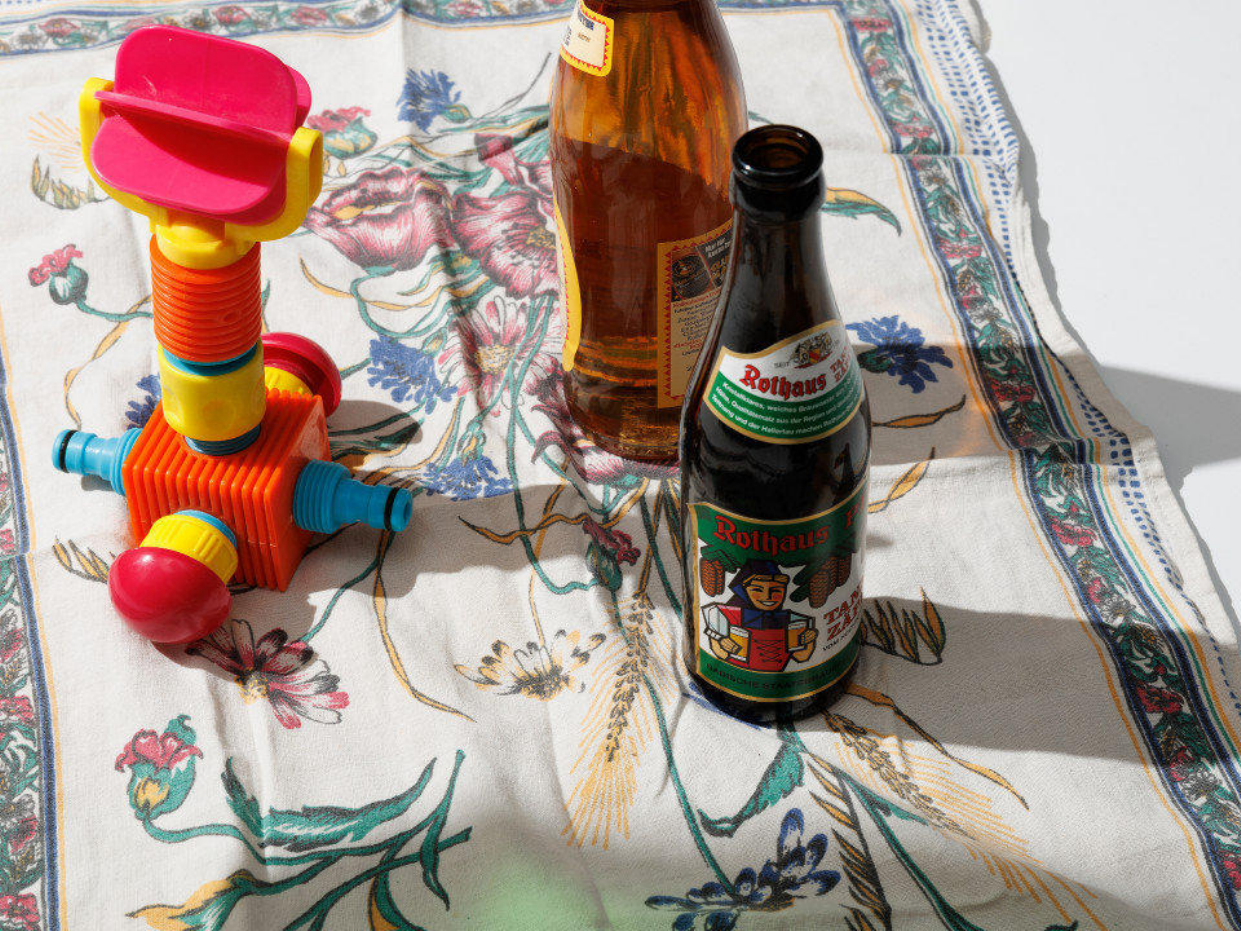} &
    \includegraphics[width=0.155\textwidth]{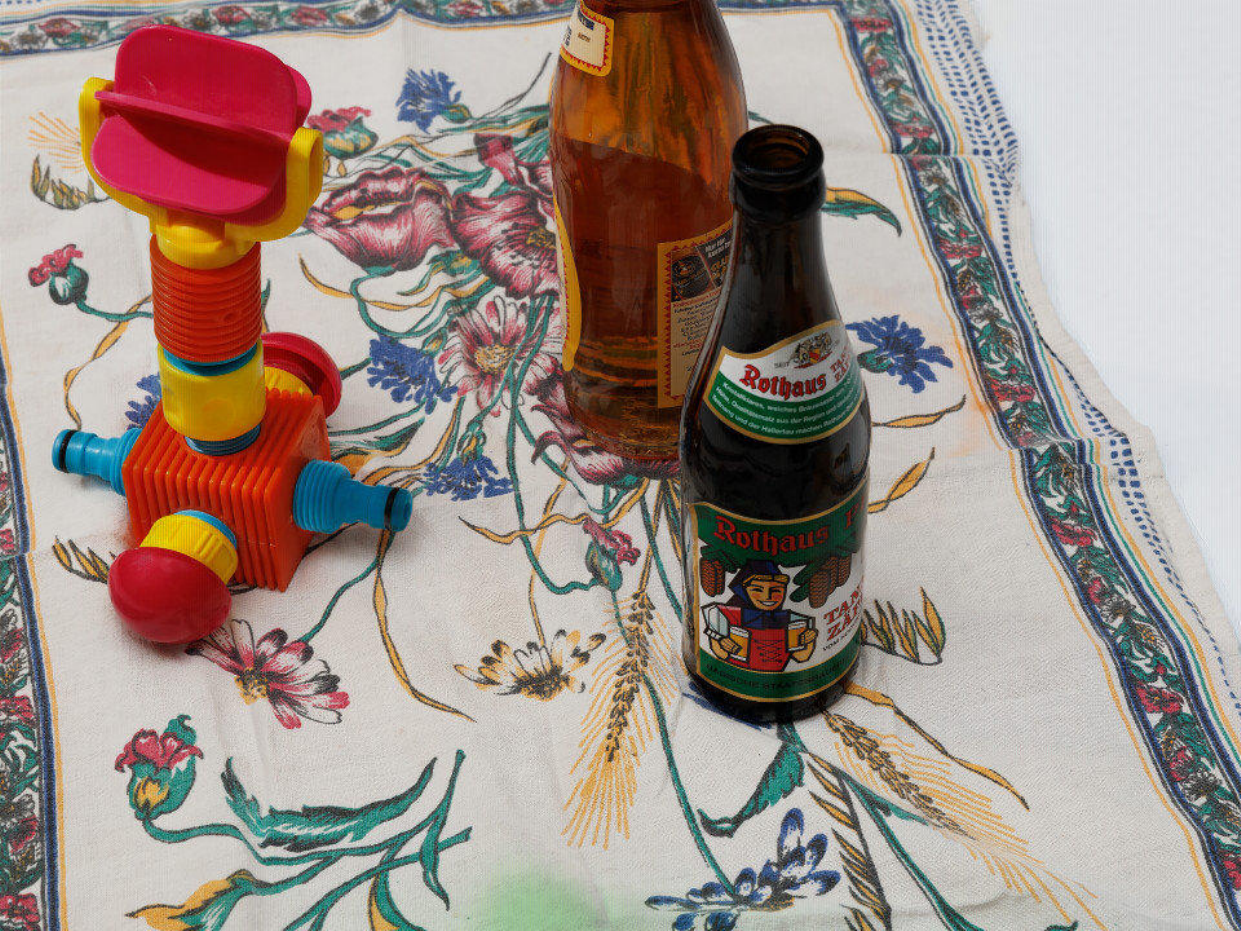} &
    \includegraphics[width=0.155\textwidth]{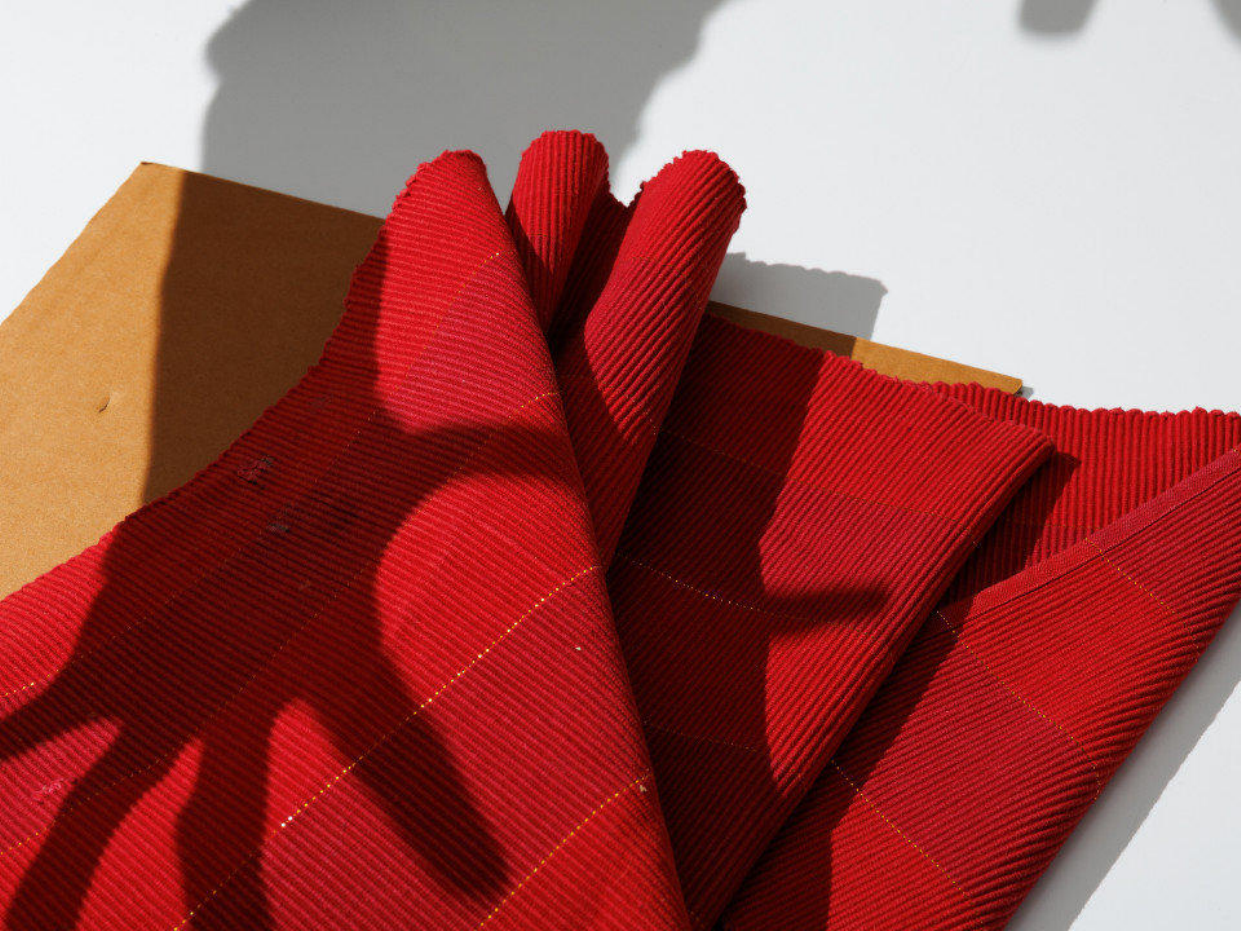} &
    \includegraphics[width=0.155\textwidth]{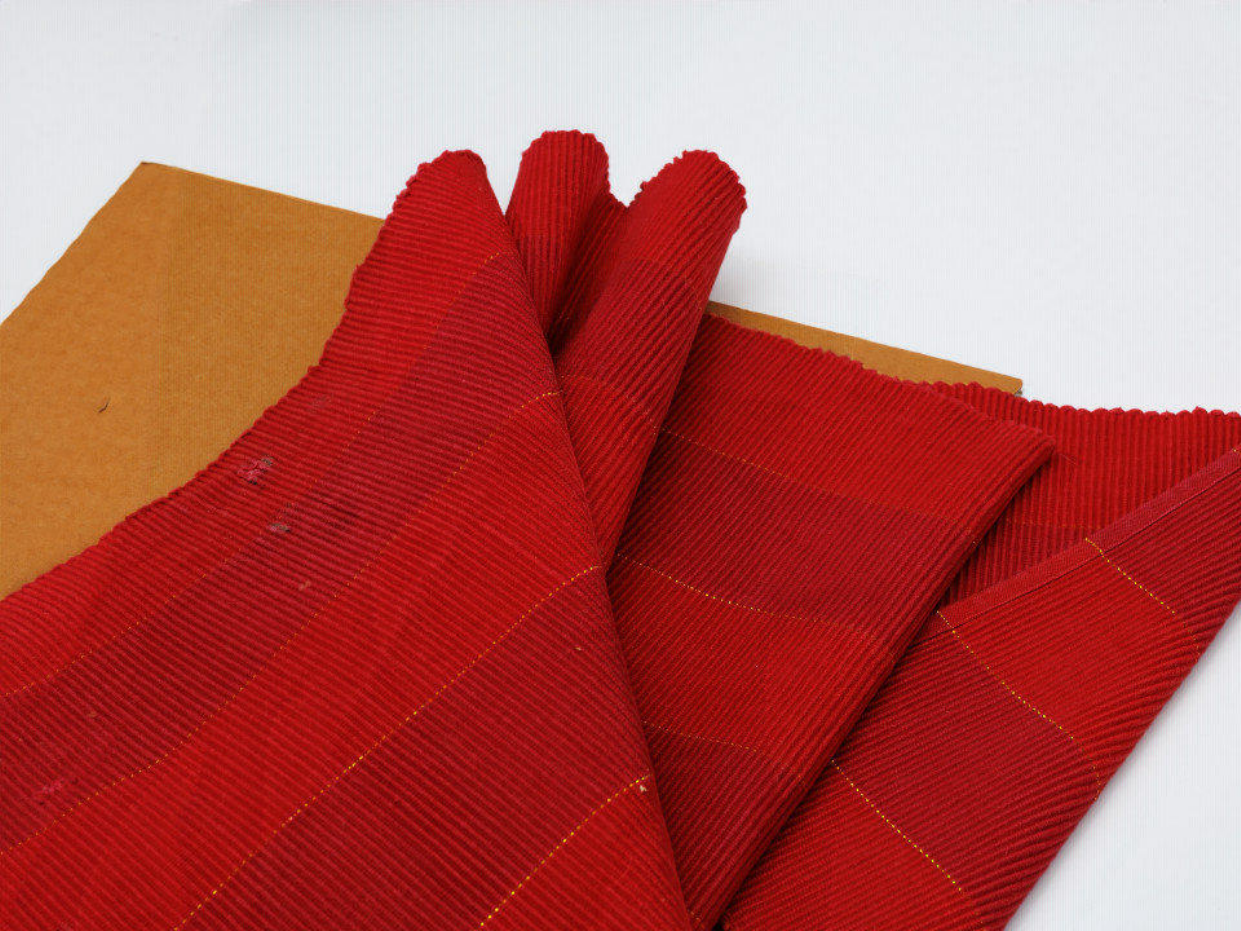} &
    \includegraphics[width=0.155\textwidth]{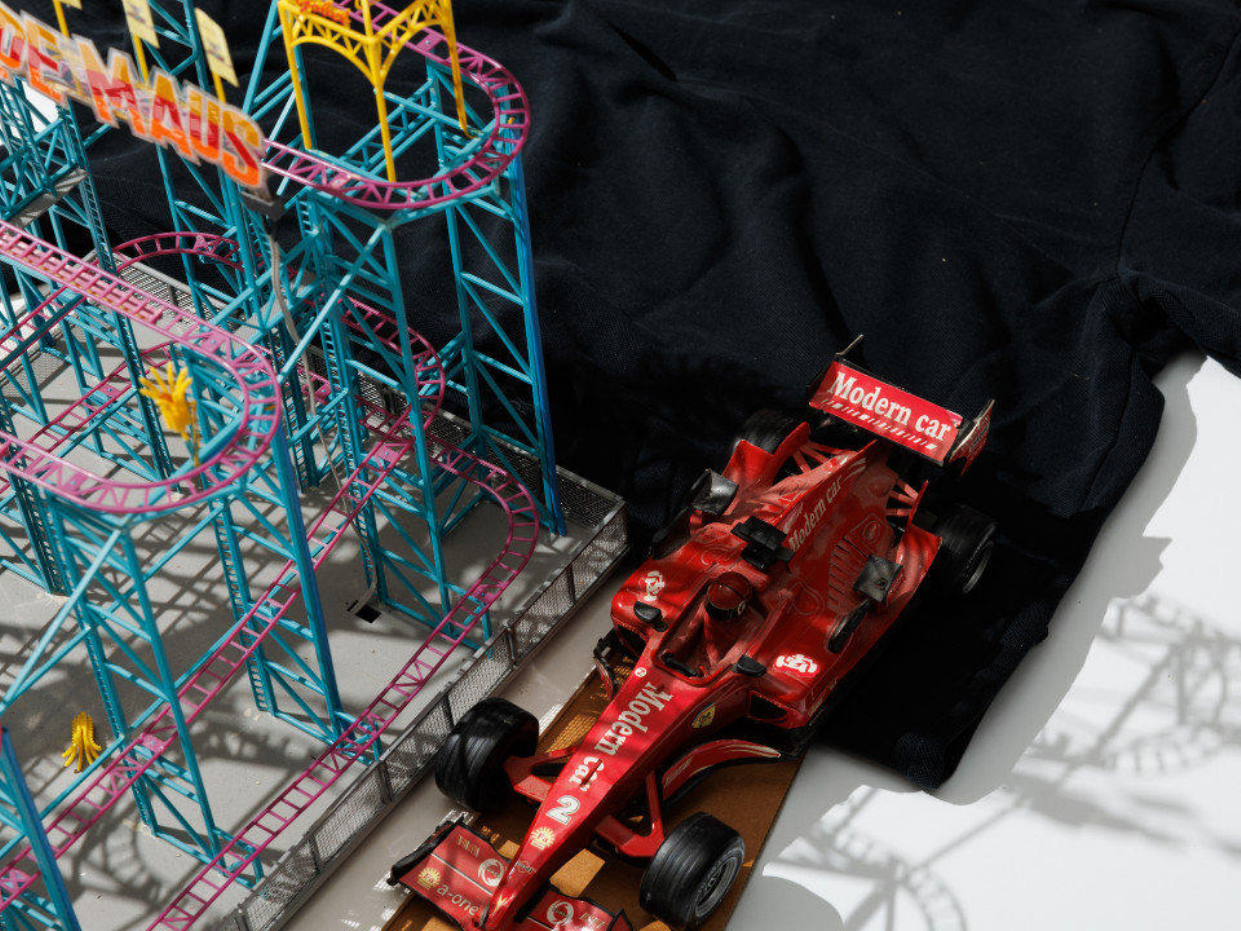} &
    \includegraphics[width=0.155\textwidth]{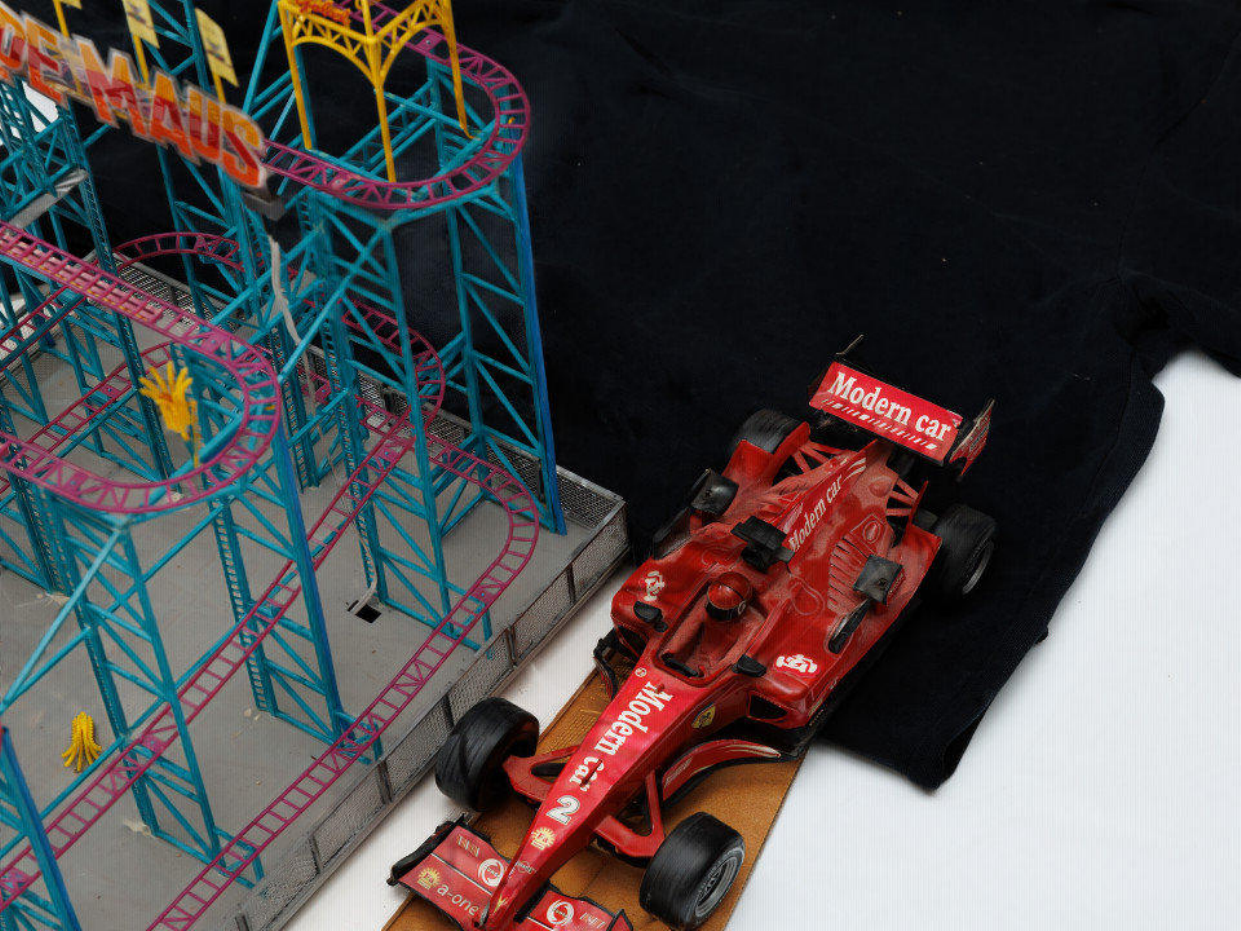} \\[-1mm]
    &
    \includegraphics[width=0.155\textwidth]{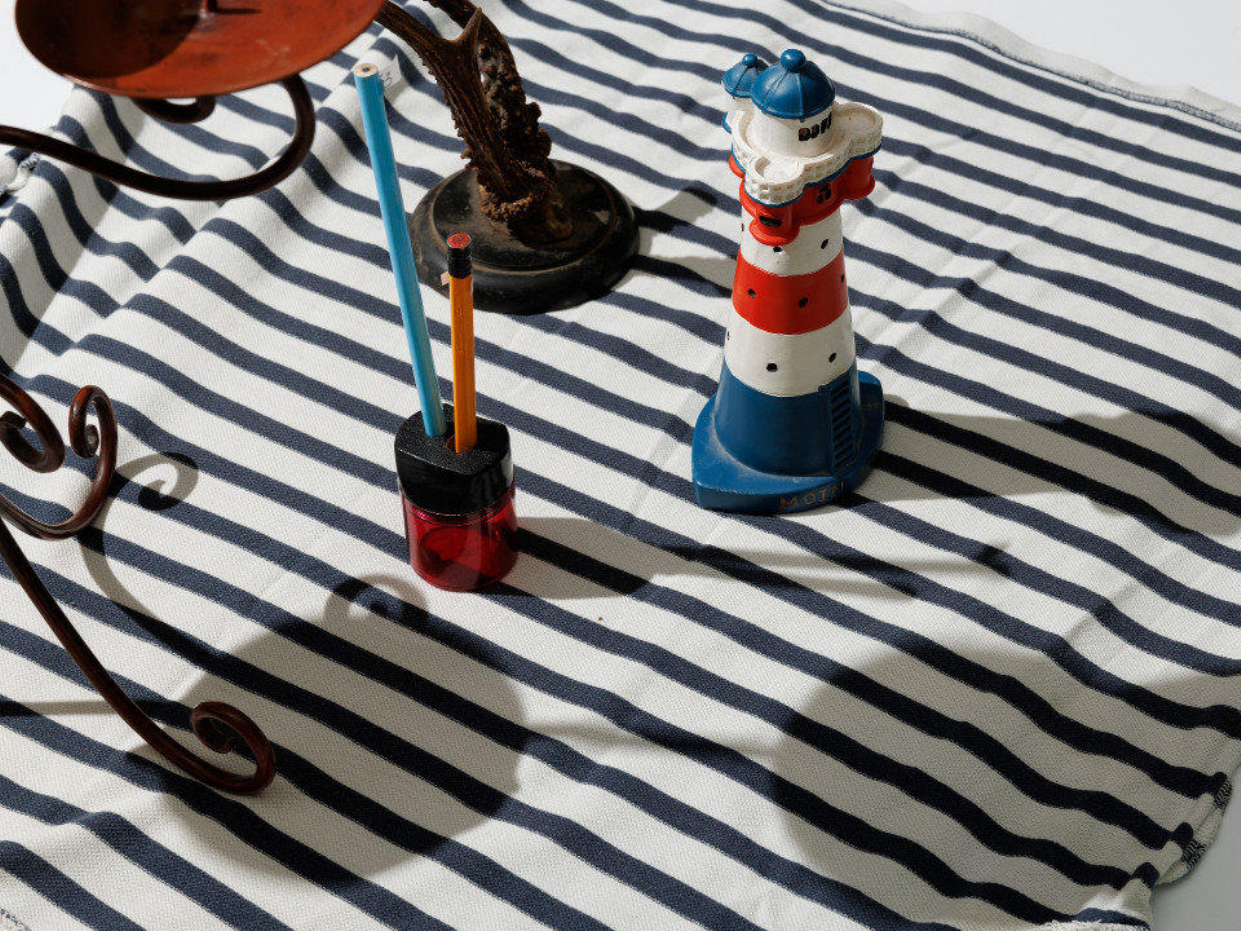} &
    \includegraphics[width=0.155\textwidth]{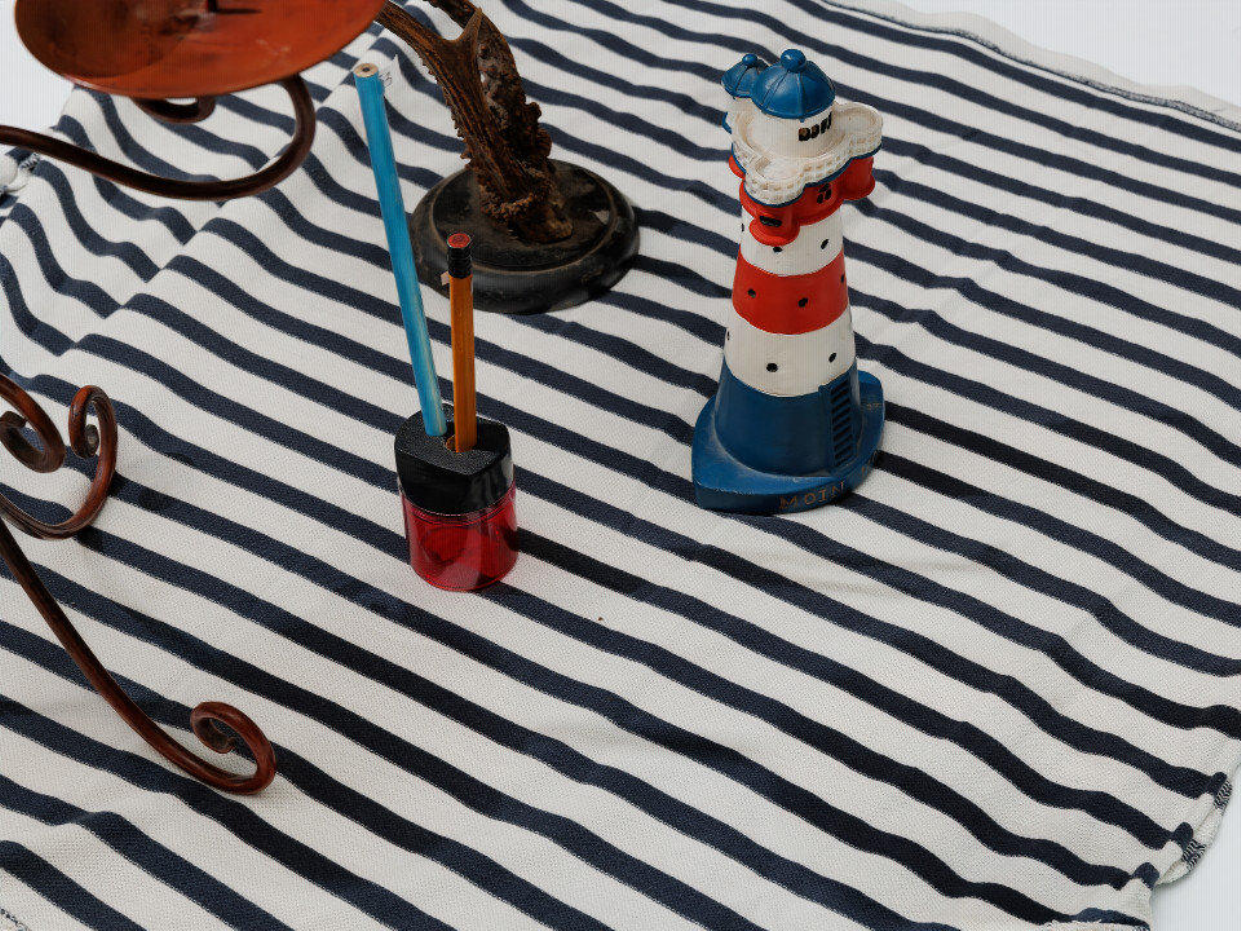} &
    \includegraphics[width=0.155\textwidth]{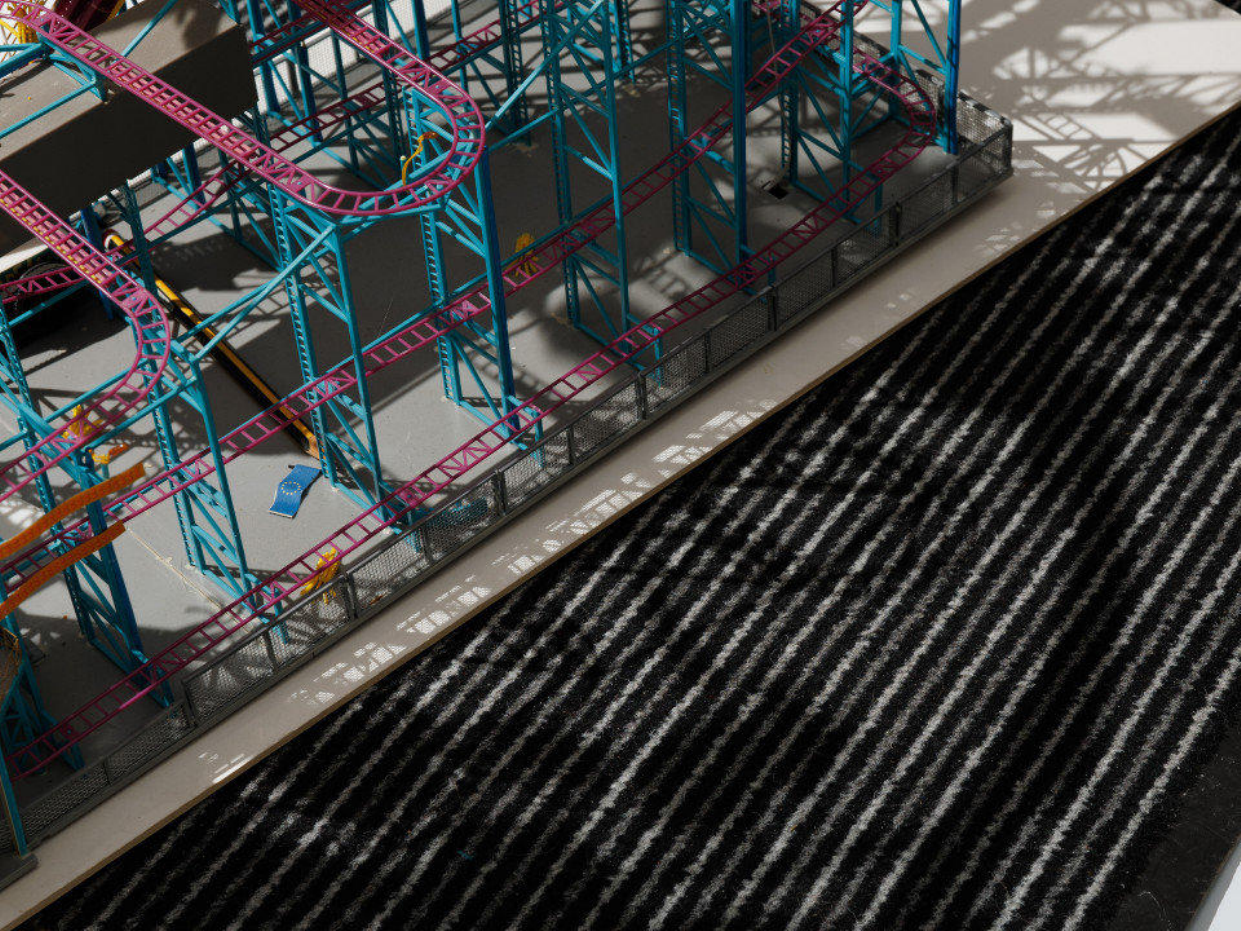} &
    \includegraphics[width=0.155\textwidth]{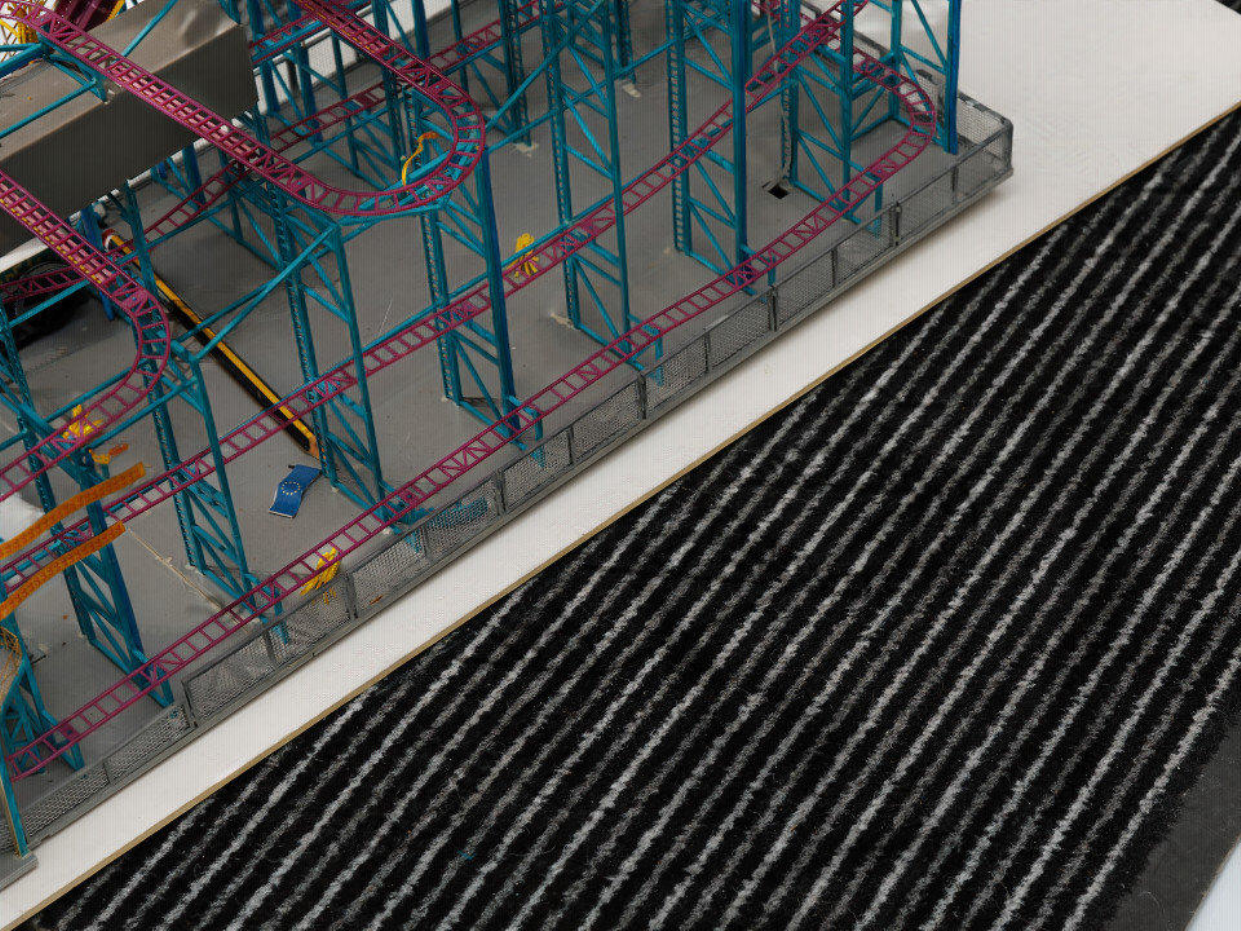} &
    \includegraphics[width=0.155\textwidth]{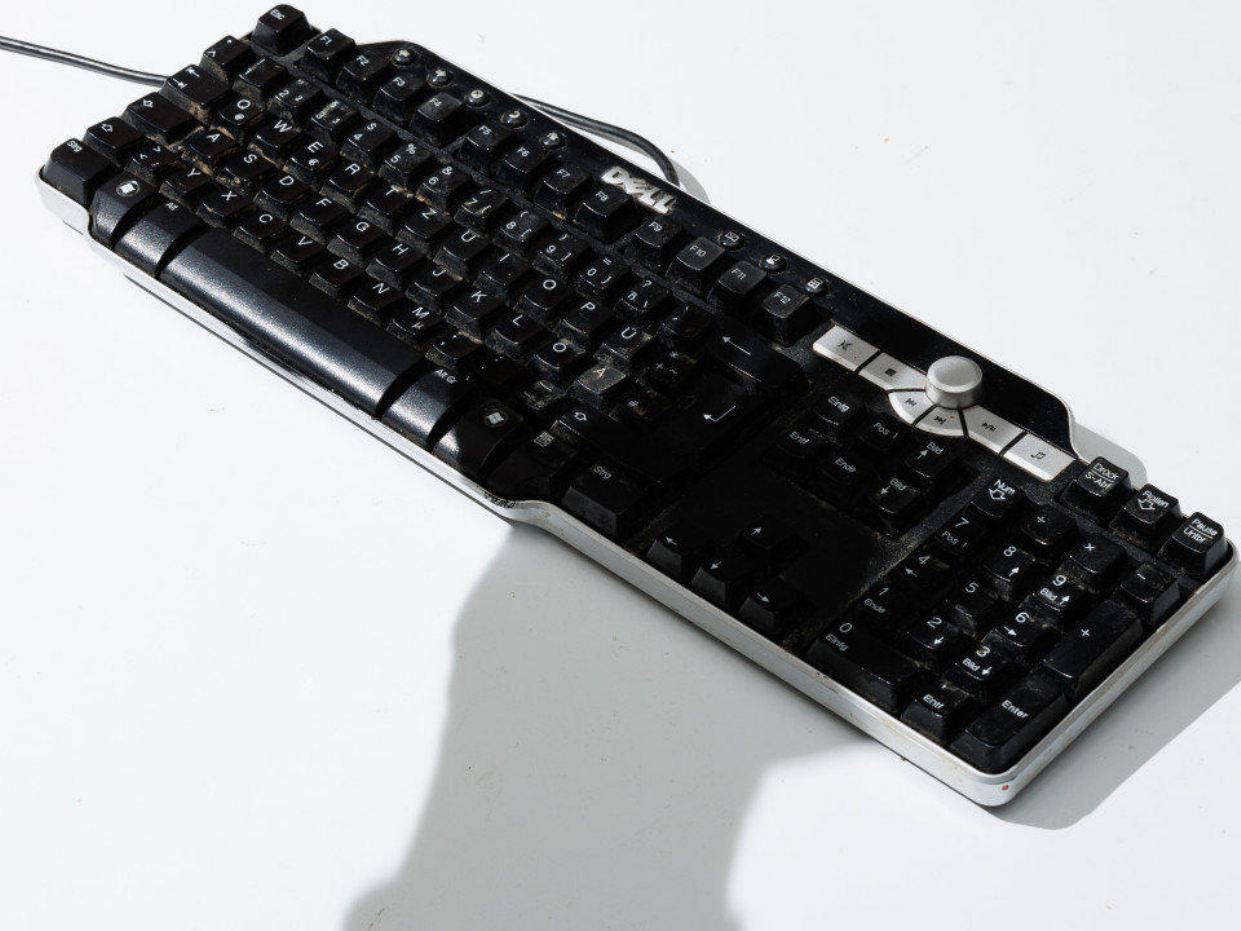} &
    \includegraphics[width=0.155\textwidth]{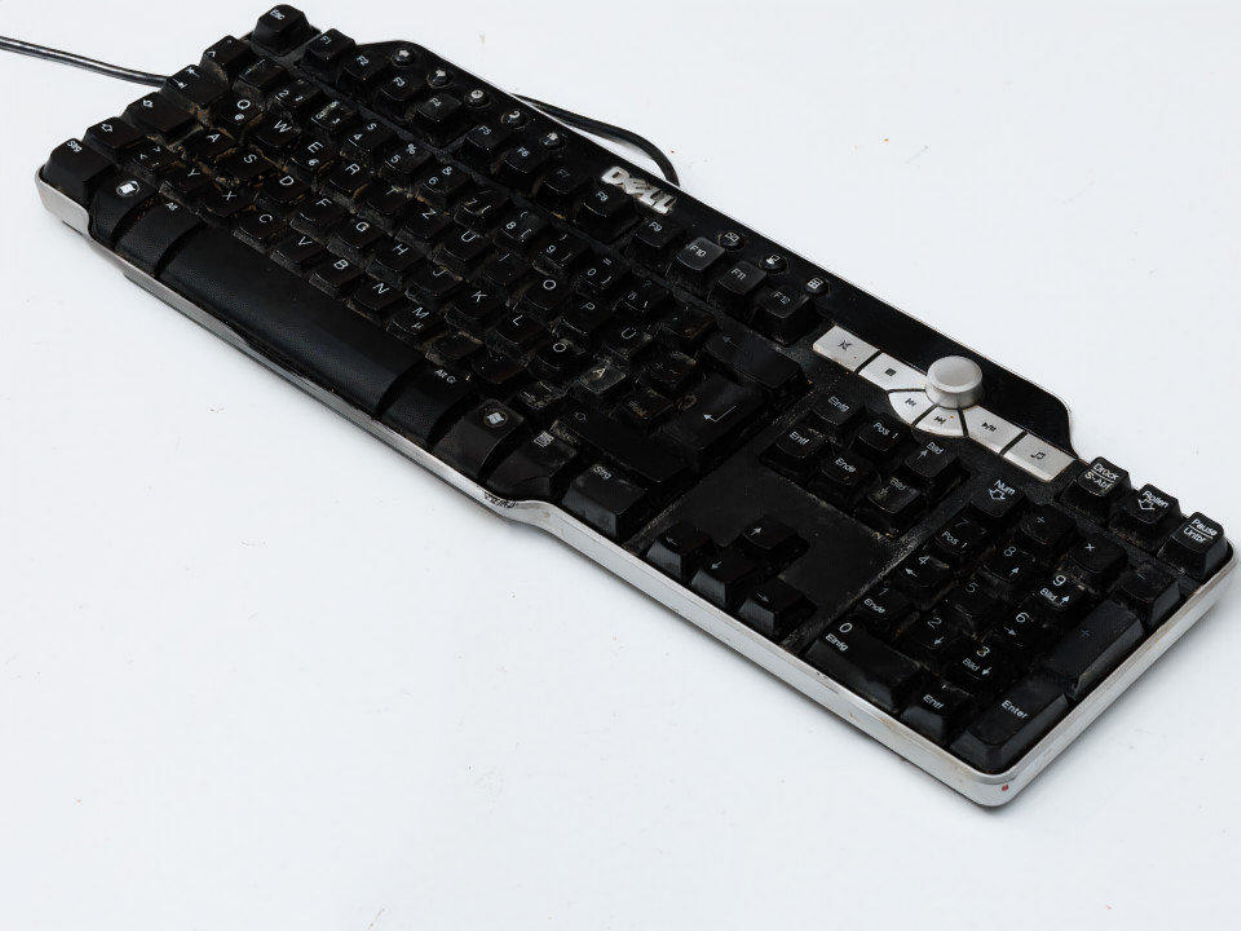} \\[1.5mm]
    
    \raisebox{0.4cm}{\rotatebox{90}{\scriptsize Dehazing}} &
    \includegraphics[width=0.155\textwidth]{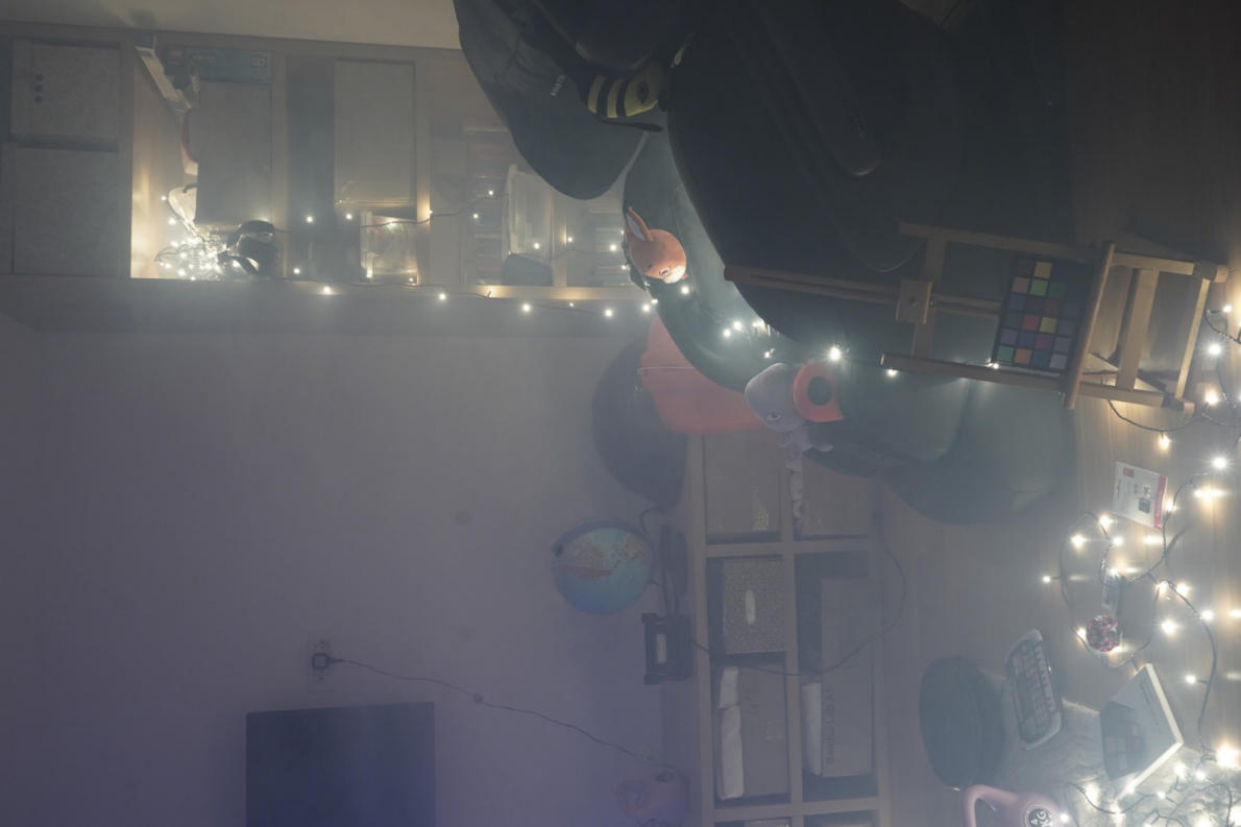} &
    \includegraphics[width=0.155\textwidth]{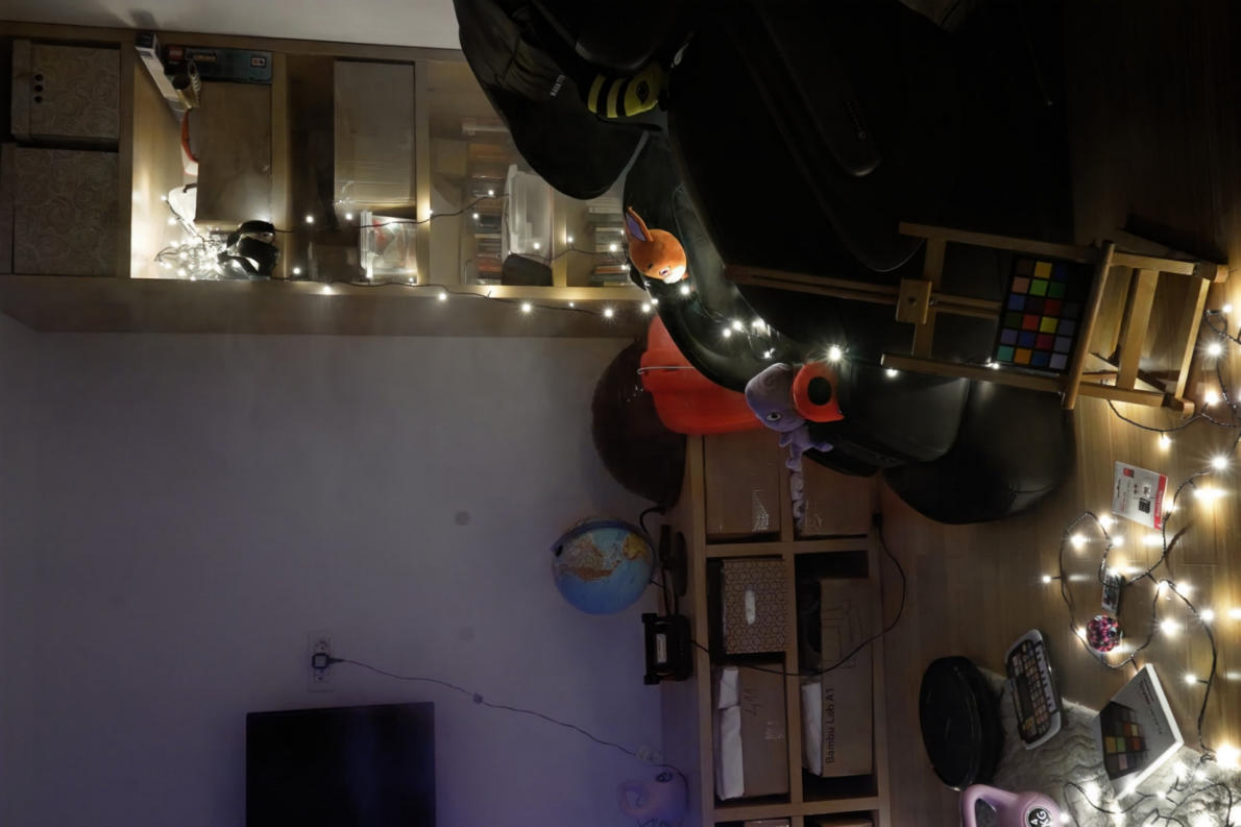} &
    \includegraphics[width=0.155\textwidth]{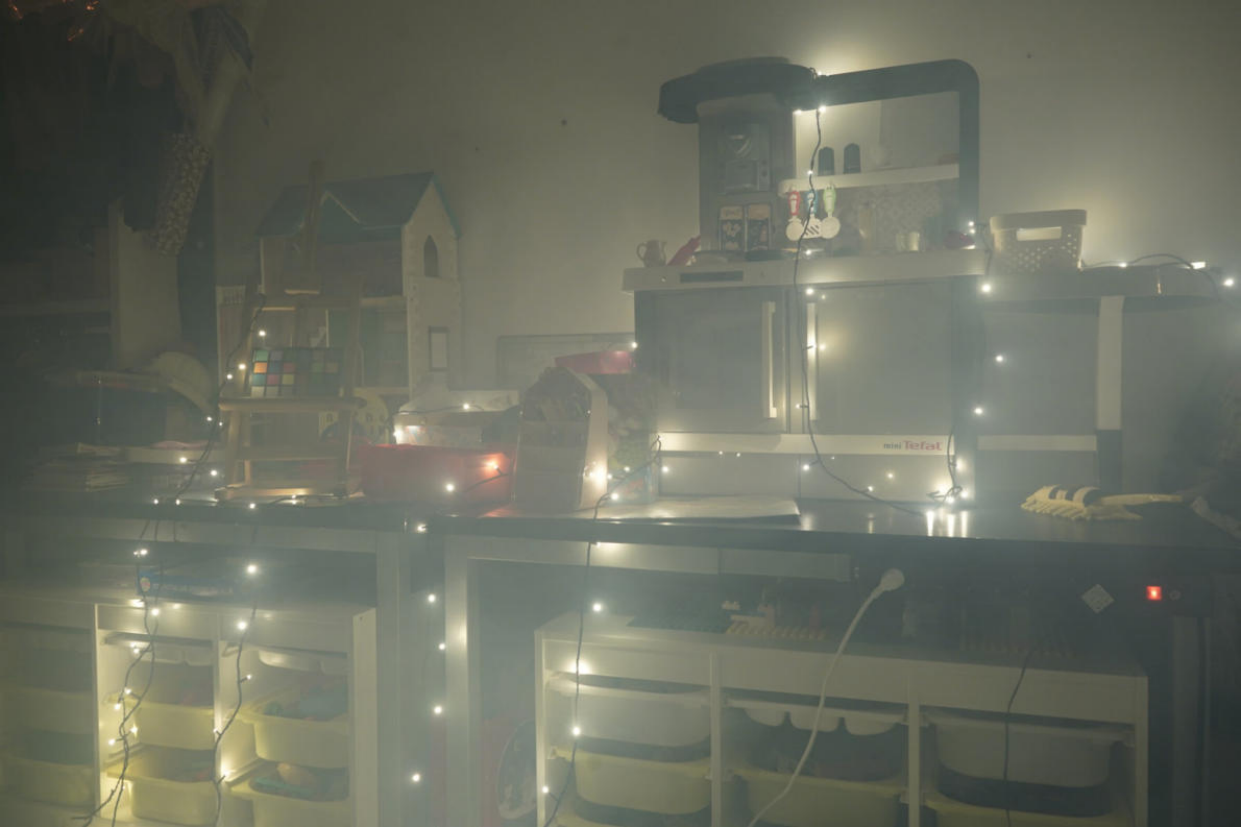} &
    \includegraphics[width=0.155\textwidth]{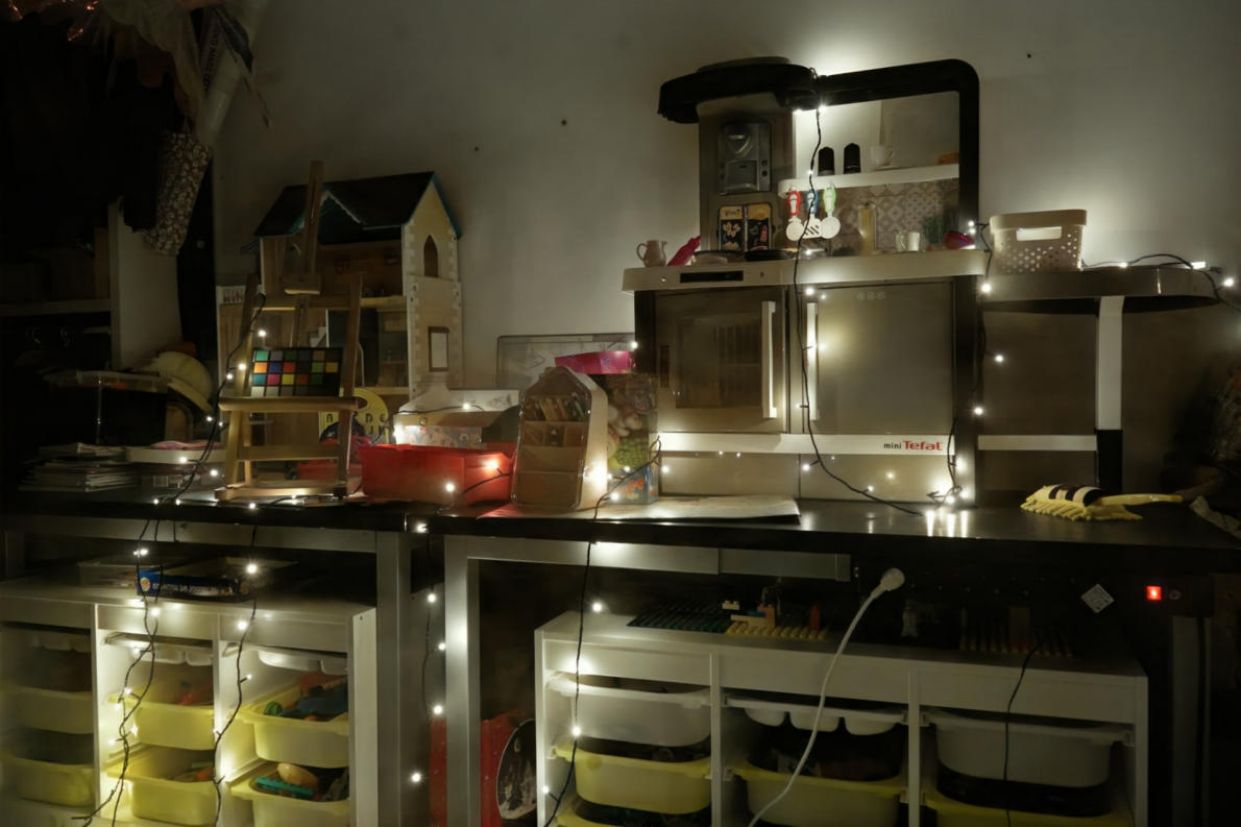} &
    \includegraphics[width=0.155\textwidth]{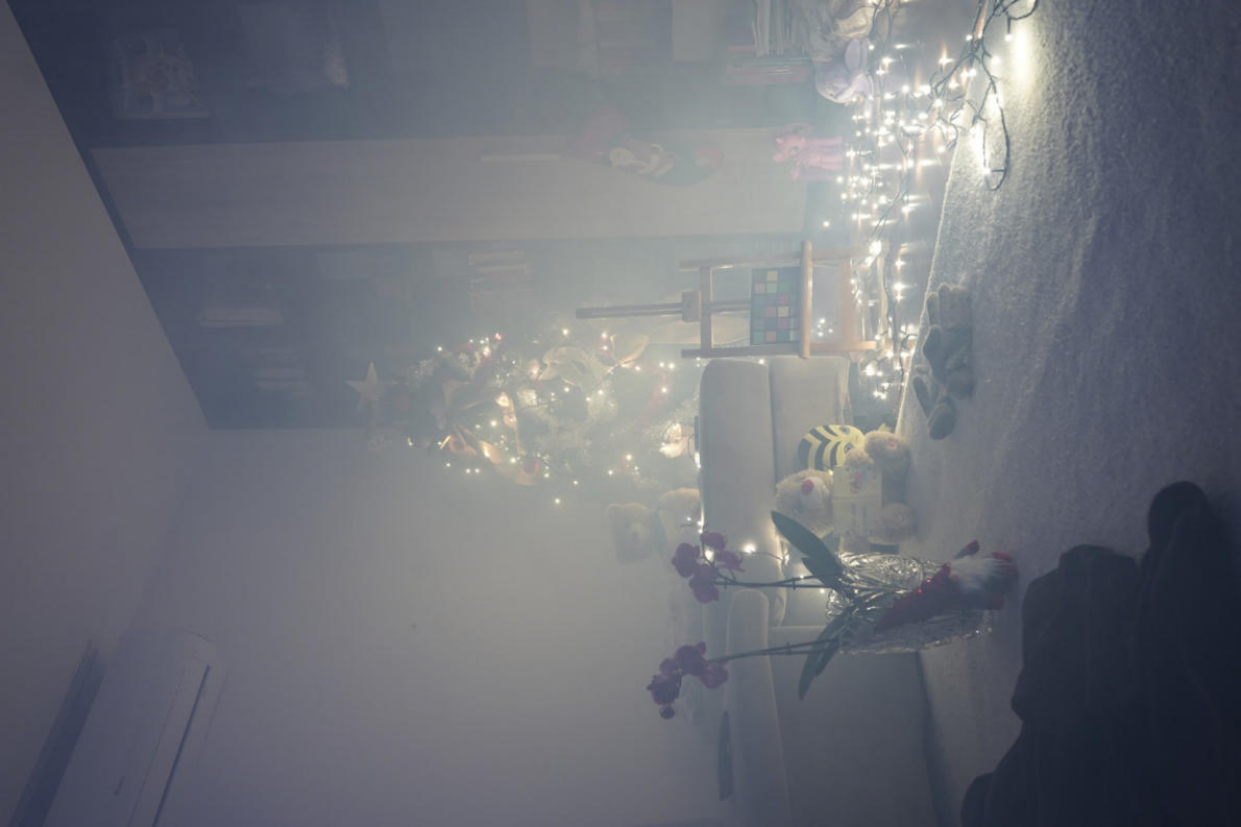} &
    \includegraphics[width=0.155\textwidth]{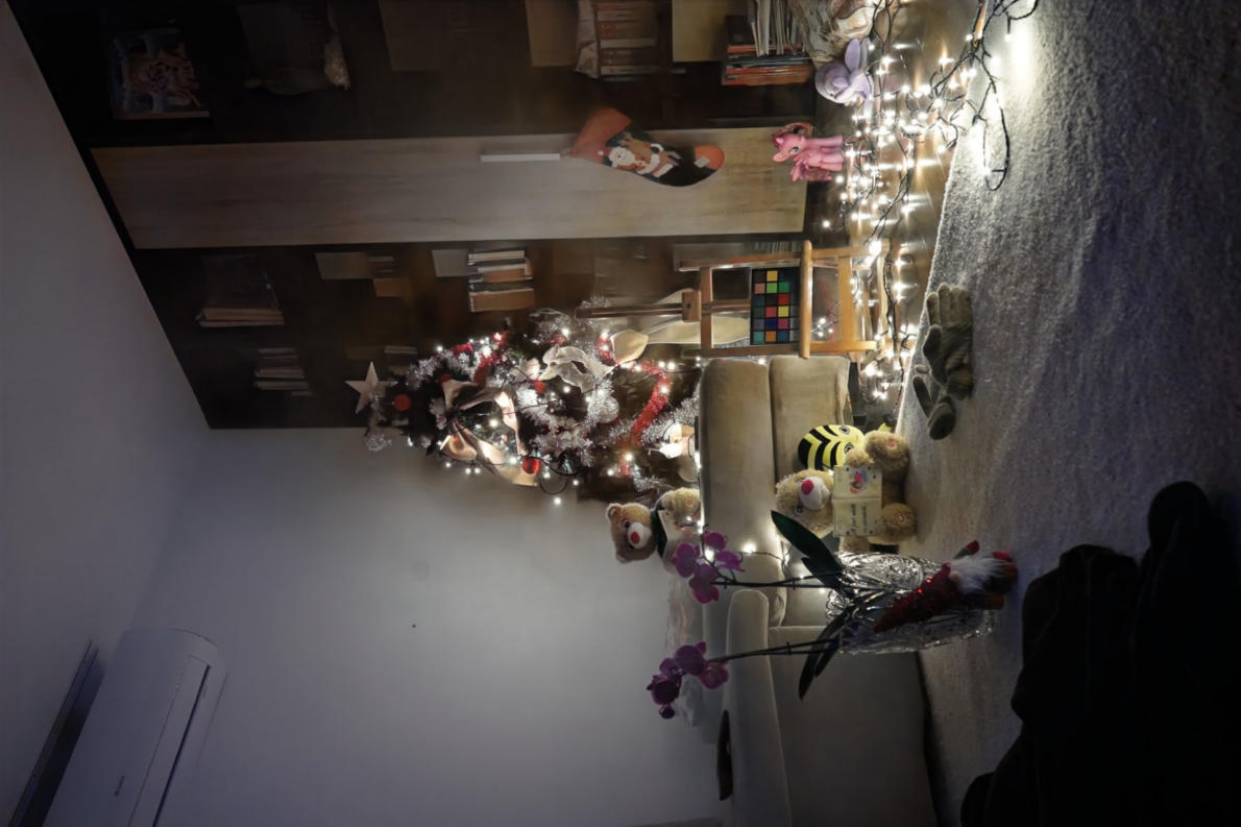} \\
  \end{tabular}

  \caption{Visual results of our proposed method on multiple image restoration tasks from NTIRE 2026 test datasets.}
  \label{fig:VisualComparisonNTIRE}
\end{figure*}

\textbf{Performance on Competitions.} Our unified RetinexDualV2 model demonstrates robust generalization across multiple NTIRE 2026 Challenge restoration tasks without task-specific modifications or compromising efficiency. It secured 4\textsuperscript{th} place in NTIRE 2026 The Second Challenge on Day and Night Raindrop Removal for Dual-Focused Images \cite{ntire26ugcvideo} (Table \ref{tab:raindrop-results}) with PSNR 27.24 dB, SSIM 0.806, and LPIPS 0.289, and 5\textsuperscript{th} place in Joint Noise Low-light Enhancement \cite{ntire26llie} (Table \ref{tab:joint-noise-results}) achieving PSNR 18.69 dB, SSIM 0.65, and LPIPS 0.50. It also performed competitively in Nighttime Dehazing \cite{ancuti2026ntire} (PSNR 24.87 dB, SSIM 0.868, LPIPS 0.134) and Shadow Removal \cite{ntire2026shadow} (PSNR 23.09 dB, SSIM 0.823, LPIPS 0.107). Qualitatively, Figure \ref{fig:VisualComparisonNTIRE} shows our method yields clearer details and reduced artifacts across these diverse datasets.

\textbf{Complexity vs. Performance Trade-off.} While RetinexDualV2 introduces a marginal parameter overhead ($\sim$0.12M, roughly a 2.5\% increase) compared to its predecessor RetinexDual, this addition encapsulates the dedicated TS-PGM and PC-MSA modules. This trade-off is intentionally designed: rather than chasing unbounded parameter scaling for pure metric inflation, this slight capacity increase effectively grounds the architecture in solid physical principles. It yields subtle but critical quantitative improvements (e.g., +0.06 dB on 4K-Rain13K, +0.05 dB on UHD-LL) while substantially improving the model's interpretability and stability across diverse, complex, and localized degradation scenarios, benefits that blindly scaling a data-driven model cannot reliably guarantee.

\begin{table}[t]
\centering
\caption{Ablation study analyzing the core physical grounding components' contributions to the network on UHD-LL\cite{Li2023ICLR}.}
\label{tab:ablation}
\vspace{-2mm}
{\fontsize{8}{10}\selectfont
\begin{tabular}{l c c | c c}
\hline
\textbf{Model Variant} & \textbf{TS-PGM} & \textbf{PC-MSA} & \textbf{PSNR$\uparrow$} & \textbf{SSIM$\uparrow$} \\
\hline
w/o Dual-Branch & $\times$ & $\times$ & 26.67 & 0.919 \\
w/o Physical Prior & $\times$ & $\times$ & 28.79 & 0.920 \\
w/o PC-MSA & \checkmark & $\times$ & 28.43 & 0.927 \\
Full Model (Ours) & \checkmark & \checkmark & \textbf{28.91} & \textbf{0.935} \\
\hline
\end{tabular}
}
\end{table}

\begin{table}[t]
\centering
\caption{Ablation study comparing different fusion strategies on the validation set of NTIRE 2026 The Second Challenge on Day and Night Raindrop Removal for Dual-Focused Images.}
\label{tab:fusion-ablation}
\vspace{-2mm}
{\fontsize{8}{10}\selectfont
\begin{tabular}{l | c c c}
\hline
\textbf{Fusion Strategy} & \textbf{PSNR$\uparrow$} & \textbf{SSIM$\uparrow$} & \textbf{LPIPS$\downarrow$} \\
\hline
W/o Fusion & 27.11 & 0.818 & 0.256 \\
MinMask & 26.90 & 0.809 & \textbf{0.230} \\
Median & 27.57 & 0.826 & 0.251 \\
Softblend (Ours) & \textbf{27.70} & \textbf{0.829} & 0.244 \\
\hline
\end{tabular}
}
\end{table}

\textbf{Ablation Study.} We conduct ablation studies to validate the physical grounding components and the rain removal fusion strategy in our pipeline.

\textbf{1) Physical Grounding Components.} We first illustrate the variations of our architectural components evaluated on the UHD-LL dataset \cite{Li2023ICLR} (Table \ref{tab:ablation}): (i) \textbf{w/o Dual-Branch}: a single-branch configuration (PG-SAMBA) without illumination-reflection decoupling. (ii) \textbf{w/o Physical Prior}: utilizes the dual-branch architecture but discards structural priors entirely. (iii) \textbf{w/o PC-MSA}: extracts priors via TS-PGM but relies on simple concatenation rather than selective conditioning. (iv) \textbf{Full Model}: explicitly steers representations via PC-MSA.
In terms of performance, the w/o Dual-Branch configuration yields only 26.67 dB PSNR and 0.919 SSIM. Incorporating the dual-branch structure (w/o Physical Prior) increases performance to 28.79 dB PSNR and 0.920 SSIM. Relying on simple concatenation (w/o PC-MSA) achieves 28.43 dB PSNR and 0.927 SSIM. Finally, our full RetinexDualV2 model reaches an optimal 28.91 dB PSNR and 0.935 SSIM, confirming the necessity of these targeted physical designs.

\textbf{2) Fusion Strategy for Rain Removal.} We evaluate four fusion strategies on the validation set of NTIRE 2026 The Second Challenge on Day and Night Raindrop Removal for Dual-Focused Images (Table \ref{tab:fusion-ablation}): (i) \textbf{W/o Fusion}: a baseline without any fusion. (ii) \textbf{MinMask}: per-pixel Minimum Mask fusion. (iii) \textbf{Median}: takes the median fusion across scene predictions. (iv) \textbf{Softblend (Ours)}: weights predictions inversely proportional to the generated mask.
Discussing the performance, the no-fusion baseline achieves 27.11 dB PSNR and 0.818 SSIM. While MinMask fusion yields the lowest LPIPS (0.230), it severely degrades PSNR (26.90 dB) and SSIM (0.809). Median fusion improves over the baseline at 27.57 dB PSNR. Ultimately, our Softblend approach attains the best results at 27.70 dB PSNR and 0.829 SSIM. Consequently, we employ Softblend to combine multi-variant scene predictions in our final pipeline.

\section{Conclusion}
\label{sec:conclusion}

We presented RetinexDualV2, a unified and parameter-efficient framework for diverse UHD image restoration tasks. By integrating a Task-Specific Physical Grounding Module (TS-PGM) and Physical-conditioned Multi-head Self-Attention (PC-MSA), our model explicitly leverages degradation-aware physical priors to guide a dual-branch Retinex decomposition network. This design gracefully overcomes the generalizability limits of purely data-driven black-box architectures, adapting seamlessly to various degradations without task-specific modifications. RetinexDualV2's robustness is evidenced by its competitive performance, notably securing 4\textsuperscript{th} place in the NTIRE 2026 Day and Night Raindrop Removal Challenge and 5\textsuperscript{th} place in the Joint Noise Low-light Enhancement Challenge. Our extensive evaluations underscore the value of integrating physical priors, highlighting meaningful progress toward interpretable foundation models for adverse weather image restoration.

{
    \small
    \bibliographystyle{ieeenat_fullname}
    \bibliography{main}
}


\end{document}